\documentclass{article}

\usepackage{PRIMEarxiv}

\usepackage[utf8]{inputenc} 
\usepackage[T1]{fontenc}    
\usepackage{hyperref}       
\usepackage{url}            
\usepackage{booktabs}       
\usepackage{amsfonts}       
\usepackage{nicefrac}       
\usepackage{microtype}      
\usepackage{lipsum}
\usepackage{fancyhdr}       
\usepackage{graphicx}       
\graphicspath{{media/}}     
\usepackage{amsmath}
\usepackage{bm}
\usepackage{tabularx}
\usepackage{color}
\usepackage{colortbl}
\usepackage{xcolor}
\usepackage{subcaption}
\usepackage{natbib}

\DeclareMathOperator*{\argmax}{arg\,max}
\DeclareMathOperator*{\argmin}{arg\,min}

\newif\ifdraft

\newif\ifcauchyerror

\newcommand{\condfootnote}[1]{\ifcauchyerror{\leavevmode\footnote{{#1}}}\else{\vspace{0ex}}\fi}

\ifcauchyerror
    \newcommand{\condfactor}{4\pi}
\else
    \newcommand{\condfactor}{2\pi}
\fi

\newcommand{\new}[1]{\ifdraft{\leavevmode\color{blue}{{#1}}}\else{{#1}}\fi}

\newcommand{\nnew}[1]{\ifdraft{\leavevmode\color{violet}{{#1}}}\else{{#1}}\fi}

\newcommand{\alex}[1]{\ifdraft{\leavevmode\color{violet}{[AM]:
{#1}}}\else{\vspace{0ex}}\fi}

\newcommand{\quotes}[1]{``#1''}

\setcitestyle{square}
\setcitestyle{citesep={,}}

\title{Kernel Density Estimation \\ for Multiclass Quantification
}

\author{
  Alejandro Moreo \\
  Istituto di Scienza e Tecnologie dell’Informazione, \\
  Consiglio Nationale delle Ricerche \\
  Pisa, Italy\\
  \texttt{alejandro.moreo@isti.cnr.it} \\
   \And
  Pablo González, Juan José del Coz \\
  Artificial Intelligence Center \\
  University of Oviedo \\
  Asturias, Spain\\
  \texttt{\{gonzalezgpablo,juanjo\}@uniovi.es} \\
}

\begin{document}
\maketitle

\begin{abstract}
Several disciplines, like the social sciences, epidemiology, sentiment analysis, or market research, are interested in knowing the distribution of the classes in a population rather than the individual labels of the members thereof. Quantification is the supervised machine learning task concerned with obtaining accurate predictors of class prevalence, and to do so particularly in the presence of label shift. The distribution-matching (DM) approaches represent one of the most important families among the quantification methods that have been proposed in the literature so far. Current DM approaches model the involved populations by means of histograms of posterior probabilities. In this paper, we argue that their application to the multiclass setting is suboptimal since the histograms become class-specific, thus missing the opportunity to model inter-class information that may exist in the data. We propose a new representation mechanism based on multivariate densities that we model via kernel density estimation (KDE). The experiments we have carried out show our method, dubbed KDEy, yields superior quantification performance with respect to previous DM approaches. We also investigate the KDE-based representation within the maximum likelihood framework and show KDEy often shows superior performance with respect to the expectation-maximization method for quantification, arguably the strongest contender in the quantification arena to date. 

\end{abstract}



\section{Introduction}
\label{sec:intro}



Quantification (variously called \emph{learning to quantify} or \emph{class prevalence estimation}) is the area of supervised machine learning concerned with estimating the percentages of instances 
\new{from a population (hereafter, \emph{a bag of examples}) belonging to each of the classes of interest}
\citep{Gonzalez:2017it,Book2023}.
Quantification finds applications in many disciplines, like the social sciences, 
epidemiology, 
or market research, in which the interest lies at the aggregate level, i.e., in which inferring characteristics of the single individual (e.g., via classification, or via regression) is of little concern since knowing group-level information is all we need.

Despite the fact that 
\new{binary quantification} (i.e., the setting in which the classes of interest are \emph{positive} vs. \emph{negative}) has been, by far, the most studied scenario in the quantification literature ~\citep{Card:2018pb,Forman:2008kx,Bella:2010kx,Esuli:2015gh,Hassan:2020kq,Moreo:2021sp}, the truth is that many of the applications of quantification naturally arise in the multiclass regime, i.e., in cases in which there are more than two mutually exclusive classes. 
\nnew{Examples of multiclass settings} 
are ubiquitous, and may include the allocation of human resources to different departments in a company  \citep{Forman:2005fk}, 
\nnew{the analysis of different phytoplankton species that could exist in a water sample \citep{gonzalez2019automatic}},
or the analysis of the various causes of death studied in verbal autopsies \citep{King:2008fk}, to name a few.
\nnew{A more concrete example could consist of providing answers to questions like: \emph{``What is the percentage of tweets conveying positive, neutral, and negative opinions concerning a specific hashtag?''} \citep{Gao:2016uq}.}

The multiclass extension of an originally conceived binary method is sometimes trivial. For example, the well-known \emph{adjusted classify \& count} (ACC) \citep{Forman:2005fk}, a method that corrects an initial estimate of the positive counts by considering the \emph{true positive rate} (tpr) and the \emph{false positive rate} (fpr) of the classifier, finds a natural extension to the multiclass regime by modelling the misclassification-rate matrix of the classifier \citep{Vucetic:2001fk}.
%
\new{However, not all methods readily lend themselves to such a seamless multiclass adaptation.} An example of this can be found in the distribution-matching (DM) methods, first proposed by  \citet{Forman:2005fk}, which nowadays represent a very important family of methods in the quantification literature \citep{Gonzalez-Castro:2010fk,Plessis:2014ff,Maletzke:2019qd,dussap2023}. 

In a nutshell, DM approaches aim at reconstructing the distribution of the test datapoints by seeking for the closest mixture of the class-conditional distributions of the training datapoints. The mixture parameter yielding the best such matching is an estimate of the sought class prevalence values.
While in principle this intuition seems directly applicable to the multiclass case, there are technical impediments that make it difficult in practice.
Devising better ways to overcome these impediments is the central topic of this paper.

\begin{figure}[htb!]
    \centering
    \includegraphics[width=.48\textwidth]{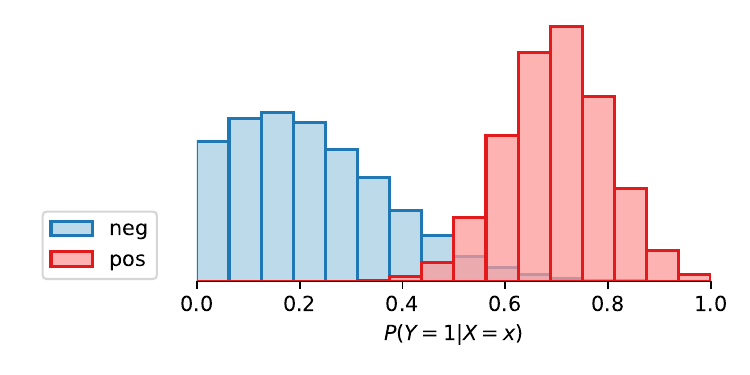}
    \includegraphics[width=.48\textwidth]{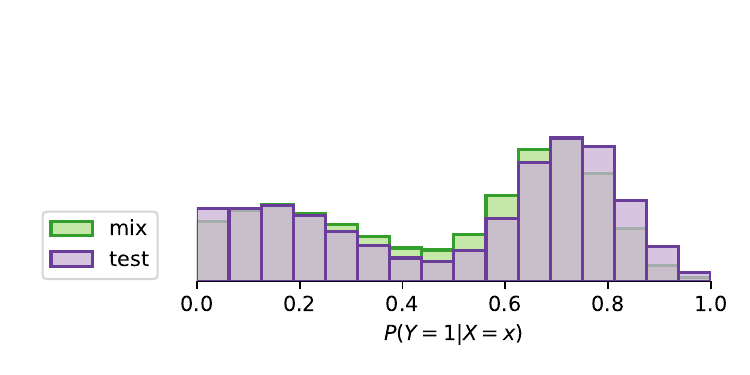}
    \caption{\nnew{DM in the binary case. Left: the distributions of the posterior probabilities of \quotes{being positive} are modelled for the positive (red) and negative (blue) examples in the training data. Right: the distribution of the posterior probabilities of \quotes{being positive} is modelled for the test examples (violet), and the parameter of a mixture of positives and negatives (green) yielding the best approximation of the test distribution is sought.}}
    \label{fig:mixture}
\end{figure}

Modelling the distribution of high-dimensional data is known to be extremely difficult. For this reason, most DM approaches first transform the datapoints into posterior probabilities, by means of a probabilistic classifier, and then try to model the distribution of posteriors in place of attempting to model the distribution of covariates (which seems hopeless specially in high dimensional spaces). 
Current DM methods model the distribution of posterior probabilities by means of \emph{histograms}. In the binary case \citep{Gonzalez-Castro:2013fk}, this comes down to generating one histogram for the distribution of posteriors from the positive training instances, and another histogram for the distribution of posteriors from the negative training instances; at inference time, another such histogram is created from the test instances, and a mixture parameter combining the training histograms is sought so that the resulting mixture of histograms best matches the test one.
\nnew{The intuition behind DM is sketched in Figure~\ref{fig:mixture}.}

Note that each of the representations (histograms) obtained for the binary case (i.e., for the positives, for the negatives, and for the test instances) need only account for the posterior distribution of ``being positive'', since the posterior probability of ``being negative'' is the complement of it and therefore it would add nothing useful to the optimization procedure if we explicitly model it. 
The multiclass case is a bit more complicated. 
In the multiclass case, one needs to generate one representation for each of the classes, plus one representation of the test instances. Furthermore, each of these representations needs to account for the posterior probabilities of all the $n$ classes, and not only of one (``being positive'') as in the binary case. Previous attempts in the literature \citep{Firat:2016uq,Bunse:2022zt} have proposed to define each such representation as a set of per-class histograms (see Section~\ref{sec:pitfalls}).
The main drawback of current DM approaches is that, when modelling a set of posterior probabilities by means of $n$ (independent) histograms, one per class, the possible dependencies among classes are irretrievably lost.
Such a representation thus precludes the method to learn from informative inter-class interactions that may exist in the data.

In this paper, we propose an alternative representation mechanism for modelling the distribution of posterior probabilities that preserves information about inter-class interactions.
In particular, we propose to replace the $n$ class-dependent histograms with a single kernel density estimation (KDE) model for the distribution of $n$-dimensional vectors (posterior probabilities) lying in the unit ($n-1$)-simplex.
We present compelling empirical evidence that our proposed KDE-based model, which we dub KDEy, does indeed improve the performance of previous (histogram-based) DM approaches. 
We also test the KDE-based representation within the maximum likelihood (ML) framework, and show our model often shows superior performance with respect to current state-of-the-art ML models for quantification as well.

The rest of this paper is structured as follows. 
In Section~\ref{sec:related} we review previous related work.
Section~\ref{sec:pitfalls} discusses the weaknesses and limitations of current DM approaches in multiclass problems.
In Section~\ref{sec:method} we propose our method while Section~\ref{sec:experiments}
reports the experimental results we have obtained.
Section~\ref{sec:conclusions} wraps up and points to some potential ideas for future work.  


\section{Background}
\label{sec:related}

\new{In this section we offer a comprehensive overview of previously proposed distribution matching approaches and how these have evolved over time. Before that, we fix our notation.}

\subsection{Notation and Problem Statement}
\label{sec:notation}

We use the following notation through this paper:
$\mathcal{X}=\mathbb{R}^d$ denotes the input space and $n$ denotes the number of classes. We write $\mathcal{Y} = \{1, 2, \ldots, n\}$ to denote the multiclass setting, and $\mathcal{Y}=\{0,1\}$ to denote the binary setting. 

We assume to have access to a labelled training set that we use for training our quantifiers, that we denote by $L=\{(\bm{x}_i,y_i)\}_{i=1}^N$, with $\bm{x}_i\in\mathcal{X}$ a datapoint with class label $y_i\in\mathcal{Y}$. The unlabelled test set on which we test our quantifiers is denoted by \nnew{$U=\{\bm{x}_i\}_{i=1}^{M}$,} with $\bm{x}_i\in\mathcal{X}$. 
We define 
\nnew{$L_i=\{\bm{x} : (\bm{x},y)\in L \wedge  y=i\}$}
as the subset of the training examples that are labelled with class $i$. Since the problem setting is single-label, note that $L=\bigcup_{i=1}^n L_i$ and $L_i \cap L_j = \emptyset$ if $i\neq j$.

The task of quantification consists of learning predictors of the probability distribution of the classes in \new{bags} of unlabelled instances, i.e., functions implementing a map $\mathbb{N}^\mathcal{X} \rightarrow \Delta^{n-1}$ that, given a multi-set \nnew{(bag)}, return a vector $\hat{\bm{\alpha}}$ (an estimate of the class distribution in the bag) lying on the unit $(n-1)$-simplex (also known as ``the probability simplex'') defined by $\Delta^{n-1}=\{(p_1,\ldots,p_n) : p_i\geq 0, \sum_{1\leq i\leq n}p_i=1\}$. 

The task of quantification is interesting precisely in situations in which the class prevalence is assumed non-stationary, i.e., in which future data \new{bags} are expected to be characterized by different class prevalence 
\new{with respect to the one observed in $L$.}
For this reason, quantification is typically studied under \emph{prior probability shift} (PPS) conditions \citep{Storkey:2009lp}  \citep[also known as \emph{label shift} or \emph{target shift} in][]{Lipton:2018fj,Alexandari:2020dn}, i.e., under the assumption that there are two different distributions, $\mathbb{P}$ and $\mathbb{Q}$, that govern the training and test observations, respectively, such that the prevalence of the classes (the \emph{priors}) are allowed to change
while the class-conditional distributions (the \emph{posteriors}) are expected not to change. 
\nnew{More formally,
%
   we say that $\mathbb{P}$ and $\mathbb{Q}$ are related to each other by means of prior probability shift if it holds that
   \begin{align}
   \label{eq:pps}
   \begin{split}
       \mathbb{P}(Y) &\neq \mathbb{Q}(Y) , \\
       \mathbb{P}(X|Y) &= \mathbb{Q}(X|Y) .
   \end{split}
   \end{align}   
}

One important implication of this problem setting is that one of the foundational assumptions underlying traditional machine learning methods is violated; i.e., the assumption that the training and the test data instances are sampled independently and identically (iid) from the same distribution.\footnote{The task of quantification is trivial in the iid setting, since a method that simply returns the training prevalence as an estimate of the test prevalence would perform almost perfectly (even though such a method does not even inspect the test set at all).}

Under PPS conditions, it makes sense to factorize our distributions as $\mathbb{P}=\sum_{i=1}^n \beta_i P_i$ and $\mathbb{Q}=\sum_{i=1}^n \alpha_i Q_i$, where $P_i=\mathbb{P}(X|Y=i)$ and $Q_i=\mathbb{Q}(X|Y=i)$ are the class-conditional densities, and where $\bm{\beta}=\{\beta_1,\ldots,\beta_n\}$ is the (observed) training class prevalence (i.e., $\beta_i=\mathbb{P}(Y=i)$), and $\bm{\alpha}=\{\alpha_1,\ldots,\alpha_n\}$ is the test class prevalence we want to estimate (i.e., $\alpha_i=\mathbb{Q}(Y=i)$). Note that the PPS assumptions imply $P_i=Q_i$ and $\bm{\beta}\neq\bm{\alpha}$.

In this paper, we analyze quantification methods that rely on
the predictions of a classifier. Quantifiers of this type are typically known as \emph{aggregative} quantifiers in the literature \citep{Book2023}. A (hard) classifier is a function $g : \mathcal{X} \rightarrow \mathcal{Y}$ issuing label predictions $\hat{y}=g(\bm{x})$.
\citet[Lemma 1]{Lipton:2018fj} showed that, since $\hat{y}$ depends on $y$ only via $\bm{x}$, the PPS assumptions \eqref{eq:pps} imply $\mathbb{P}(\hat{Y}|Y)=\mathbb{Q}(\hat{Y}|Y)$, where $\hat{Y}$ is the random variable of classifier predictions. 

This implication readily generalizes to other feature mappings $\phi : \mathcal{X} \rightarrow \widetilde{\mathcal{X}}$ provided that these mappings are measurable. In this context, the PPS assumption $\mathbb{P}(\widetilde{X}|Y)=\mathbb{Q}(\widetilde{X}|Y)$ in combination with the law of total probability, i.e., $\mathbb{Q}(\widetilde{X})=\sum_{i=1}^n \mathbb{Q}(\widetilde{X}|Y=i)\mathbb{Q}(Y=i)$ altogether settle down the bases that enable reframing a solution in terms of the factorization 
\begin{equation*}
    \label{eq:lawpps}
    \mathbb{Q}(\widetilde{X})=\sum_{i=1}^n \mathbb{P}(\widetilde{X}|Y=i)\mathbb{Q}(Y=i) ,
\end{equation*}
\noindent where $\widetilde{X}$ is the random variable of $\widetilde{\bm{x}}=\phi(\bm{x})\in\widetilde{\mathcal{X}}, \bm{x}\in\mathcal{X}$. This is the underlying principle at the heart of various frameworks for distribution matching adopted in the quantification \citep{Firat:2016uq,Bunse:2022zt} and label shift \citep{Garg:2020jt} literature.

In particular, we are interested in \emph{probabilistic aggregative} quantifiers, i.e., quantifiers that implement such mapping by means of a soft classifier  $s : \mathcal{X} \rightarrow \Delta^{n-1}$ issuing posterior probabilities $s(\bm{x})=(p_1,\ldots,p_n)$, where $p_i$ is the confidence score the classifier $s$ attributes to the belief that $\bm{x}$ belongs to class $i$. 

Through this article, we use $\widetilde{\cdot}$ to clearly indicate that we are modelling posterior probabilities. The following abbreviations will prove useful for the definitions to come:   
$\widetilde{\bm{x}}=s(\bm{x})$, $\widetilde{P}_i=\mathbb{P}(\widetilde{X}|Y=i)$, $\widetilde{Q}_i=\mathbb{Q}(\widetilde{X}|Y=i)$.



\alex{Quito la definición de histograma. Me convenció Pablo que es una cosa demasiado básica como para formalizarla; todo el mundo sabe lo que es un histograma y no aporta nada.}

\subsection{Previous Distribution Matching Methods for Multiclass Quantification}

Distribution matching methods represent one of the most genuine approaches for quantification. In this section, we overview the major advancements distribution matching methods have undergone in the related literature.

The first distribution matching method for quantification was proposed by \citet{Forman:2005fk} (see also \citep{Forman:2008kx}). The method, dubbed Mixture Models (MM), was devised for binary quantification and is based on the observation that
\begin{equation*}
    \mathbb{Q} = \alpha Q_{1} + (1-\alpha) Q_{0} ,
\end{equation*}
i.e., on the fact that the distribution of test instances ($\mathbb{Q}$) is a mixture of the class-conditional distribution of positives ($Q_{1}$) and the class-conditional distribution negatives ($Q_{0}$), where $\alpha$ is the mixture parameter, which corresponds to the proportion of positives.
Here, the distributions $\mathbb{Q}, Q_1, Q_0$ are represented by means of \new{discrete cumulative distribution functions (CDFs)} of classifier scores $\widetilde{Q}, \widetilde{Q}_1, \widetilde{Q}_0$. 
Of course, labels are not observed for the test set, and hence $\widetilde{Q}_1$ and $\widetilde{Q}_0$  are unknown. However, under PPS conditions the class-conditional probability distributions are assumed stationary, and therefore one can safely assume that $\widetilde{Q}_{1} \approx \widetilde{P}_{1}$ and $\widetilde{Q}_{0} \approx \widetilde{P}_{0}$, where $\widetilde{P}_{1}$ and $\widetilde{P}_{0}$ stand for the cumulative discrete distributions of classifier scores obtained, via cross-validation, for the positive and negative instances, respectively, in the training set. MM thus tries to reconstruct the test distribution as a linear combination of the class-conditional training distributions, with respect to any given specific criterion $\mathcal{L}$
\begin{equation*}
    \alpha^* = \argmin_{0\leq \alpha \leq 1} \mathcal{L}(\alpha \widetilde{P}_{1} + (1-\alpha) \widetilde{P}_{0}, \widetilde{Q}) .
\end{equation*}

Forman proposed two variants of this method, the Kolmogorov-Smirnov MM and the PP-Area MM, based on different choices of $\mathcal{L}$. The optimization procedure for $\alpha$ was simply formulated as a linear search in the interval $[0,1]$ in small steps. \nnew{It was later shown that minimizing the PP-Area is equivalent to minimizing the L1-norm between the two CDFs \citep{Firat:2016uq}, and it has been recently proved that this is equivalent to minimizing the Wasserstein distance (also known as the \emph{Earth Mover Distance}---EMD) between the corresponding PDFs \citep{castano2023equivalence}.}

Later on, \citet{Gonzalez-Castro:2013fk} proposed a variant of MM in which (i) the classifier scores are replaced by posterior probabilities, (ii) the discrete distributions ($\widetilde{P}_{1}, \widetilde{P}_{0},\widetilde{Q}$) are not cumulative, i.e., are \new{PDFs represented via} normalized \emph{histograms}, and (iii) the goodness of fit is computed in terms of the Hellinger Distance (HD). In the discrete case, the HD between two discrete distributions \nnew{$\widetilde{P}$ and $\widetilde{Q}$ with corresponding densities} $\bm{p}=(p_1,\ldots,p_b)$ and $\bm{q}=(q_1,\ldots,q_b)$ with $b$ bins is defined as
\begin{equation*}
    \operatorname{HD}(\bm{p},\bm{q}) =\frac{1}{\sqrt{2}} || \sqrt{\bm{p}}-\sqrt{\bm{q}}||_2 
              =\sqrt{1-\sum_{i=1}^b \sqrt{p_i q_i}} ,
\end{equation*}
so the mismatch function $\mathcal{L}$ to minimize becomes $\operatorname{HD}(\alpha \bm{p}_{1} + (1-\alpha) \bm{p}_{0}, \bm{q})$.

The method was named HDy (note the ``y'' postfix) as to make clear the fact that the discrete distributions model posterior probabilities. In the same article, HDx was also introduced, i.e., a variant that models the class-conditional histograms of the covariates (hence the ``x'' postfix). In HDx, each histogram regards a specific feature, and the HD between two bags is computed as the average across the per-feature HDs. This way, given two data samples represented as \nnew{$\bm{P}=(P_1,\ldots,P_d)\in\mathbb{R}^{b\times d}$ and $\bm{Q}=(Q_1,\ldots,Q_d)\in\mathbb{R}^{b\times d}$,} with $b$ the number of bins as before, and $d$ the number of features, the mismatch is defined as
\begin{equation}
\label{eq:hdy:mean}
    \mathcal{L}(\bm{P},\bm{Q}) = \frac{1}{d}\sum_{i=1}^{d} \operatorname{HD}(P_ i,Q_i) .
\end{equation}

In the same article, \citet{Gonzalez-Castro:2013fk} conducted an empirical evaluation that clearly demonstrated the superiority of HDy over HDx. 
These findings, along with the fact that non-aggregative quantifiers tend to be inefficient in high-dimensional spaces, settled a preference for subsequent MM-based methods leaning towards the aggregative type (Section~\ref{sec:notation}).



In a follow-up paper, \citet{Reis:2018fk} proposed replacing \nnew{HDy's} brute-force search of $\alpha$ with a ternary search in order to speed up the optimization problem. Nowadays, modern software packages for quantification, like \texttt{QuaPy} \citep{Moreo:2021bs}, \texttt{QFY} \citep{schumacher2023comparative}, or \texttt{qunfold} \citep{qunfold2023}, rely on different off-the-shelf optimization packages for this endeavor.

Subsequently, another work by \nnew{the same research team} was presented in which a generalization framework for binary distribution matching methods called DyS was proposed \citep{Maletzke:2019qd}. The framework considers the dissimilarity criterion as a hyperparameter of the model, of which HDy becomes one particular instance. The authors explored many other such dissimilarities, and found out that configurations using the Topsøe divergence~\citep{Cha:2007gf} often demonstrate superior quantification performance.

Yet another particularity of the original HDy concerns the way the number of bins was treated: the authors had proposed to explore this value in the range $[10,110]$ stepping by 10 and returning, as the final prevalence prediction, the median obtained across the 11 such configurations. 
\citet{Maletzke:2019qd} \nnew{also criticized this choice, arguing} that the best value for the number of bins typically falls below 20, and suggested treating this value simply as any other hyperparameter of the model which is liable to be optimized via model selection.

\alex{Juanjo dice que esto es poco relevante para el paper. Hablando con Pablo, vimos conveniente puntualizar que la finalidad de listar variantes de HDy era dejar claro que HDy había tenido repercusión en la literatura. Añado el siguiente comentario a ver si os convence:}
\nnew{HDy has then become highly influential in the field, as witnessed by the fact that much subsequent work in quantification has focused on describing new variants and new applications of it to different problems. Examples of this include the} 
ensembles models (EHDy) by \citet{Perez-Gallego:2019vl}, the variant of HDy for identifying the context of data samples by \citet{Reis:2018fk}, or the application of HDy to detect drifts in data streams proposed by \citet{maletzke2018need}, \nnew{to name a few}.
\nnew{It is noteworthy, though, that these, }
just like all the above-discussed methods, all cater for the binary setup only.

Nevertheless, a naive adaptation to the multiclass case always exists, which consists of (i) training $n$ independent binary quantifiers (one specific to each of the $n$ classes), (ii) generating prevalence estimates independently for each class, and (iii) reporting a normalized vector of class prevalence values (e.g., by means of L1-normalization or by applying a \emph{softmax} operator). This approach is typically referred to as One-vs-All (OvA). Despite its appealing simplicity, empirical studies have shown that OvA solutions fall short in terms of performance when compared with native multiclass solutions \citep{donyavi2023mc}. \nnew{This should come as no surprise since each binary problem models the negative class as an aggregation of all other classes. The negative distribution thus becomes a mixture of $(n-1)$ classes with potentially different priors in training and in test, something that directly clashes with the PPS assumptions by which the class conditional densities are expected not to change.}

The first attempt to convert HDy into a multiclass quantification method was by \citet{Firat:2016uq}. The extension was presented as part of a unification framework that expresses some of the previously known methods as a generalized constrained regression problem of the form $\bm{q}=\bm{M}\bm{\alpha}+\bm{\epsilon}$, where $\bm{\alpha}\in\Delta^{n-1}$ is the sought test prevalence vector, $\bm{M}\in\mathbb{R}^{l\times n}$ is a matrix whose columns are the class-wise $l$-dimensional representations (to be defined by each specific quantification method) of the training instances,  $\bm{q}\in\mathbb{R}^l$ is the vector representing the test instances, and $\bm{\epsilon}$ is a vector of irreducible noise. The original (binary) HDy thus emerges from the framework as the particular case in which the loss to minimize is the HD (actually, Firat used the squared Hellinger distance---hereafter $\operatorname{HD}^2$), and in which $l=b$ since $\bm{q}$ is the histogram of the test posterior probabilities, and $\bm{M}\in\mathbb{R}^{b \times 2}$ is a matrix containing, as its two columns, the histogram of posteriors generated from the positive and the histogram of posteriors generated from the negative training instances. 

Under Firat's framework, the method was said to be directly applicable to the multiclass case in a natural way. The intuition simply came down to adding one column (i.e., one class-specific histogram) for each of the $n$ classes. However, the extension proposed hides one important implication: while for $n=2$ the histograms can simply account for the \emph{a posteriori} distribution $P(Y=1|X)$ (i.e., of ``being positive''), and ignore the complementary $P(Y=0|X)$, in the multiclass case one needs to model the \emph{a posteriori} distribution of \emph{all} classes (and not, say, of $n-1$ classes), for reasons that are discussed in more detail in Section~\ref{sec:pitfalls}. 
Firat proposed to simply \emph{concatenate} the class-wise histograms (i.e., to set $l=n \cdot b$) so that $\bm{q}\in\mathbb{R}^{n\cdot b}$, and $\bm{M}\in\mathbb{R}^{(n\cdot b) \times n}$.
A side (seemingly unwanted) effect of this representation is that $\operatorname{HD}^2$ is no longer comparing \new{proper probability distributions} 
since the concatenated arrays \new{sum up to $n$.} 
Normalizing them (say, multiplying by scalar $1/n$) does not solve the issue, since this causes every value be artificially constrained to lie in the range $[0,\frac{1}{n}]$.
\new{Altogether, this mechanism produces representations that are no longer ``natural distributions''; this seems rather undesirable since the original DM framework was conceived to operate with proper PDFs.}

An alternative way for porting HDy to the multiclass regime would come down to defining the loss function as the average across $n$ independent evaluations of the HD, each operating on a specific class-dependent histogram. Note this alternative would more naturally align with the original formulation of the counterpart HDx, which already computed the average HD across the feature-dependent histograms, as in \eqref{eq:hdy:mean}. This is the formulation adopted in the more recent framework proposed by \citet{Bunse:2022zt}, and the variant implemented in software libraries like \texttt{qunfold} \citep{qunfold2023} or \texttt{QuaPy} \citep{Moreo:2021bs}. 

The differences between Firat's formulation and Bunse's formulation are subtle in practice, though. Indeed, it turns out that, when $\operatorname{HD}^2$ is adopted as the divergence to minimize, both criteria are equivalent \nnew{under the lens of} the minimization procedure, since
\begin{align*}
    \mathcal{L}(\bm{p}',\bm{q}')_{\operatorname{Firat}} & = 1 - \sum_{i=1}^{b\cdot n} \sqrt{p'_i q'_i} \\
    \mathcal{L}(\bm{p},\bm{q})_{\operatorname{Bunse}} & = \frac{1}{n}\sum_{i=1}^{n} \left(1 - \sum_{j=1}^b \sqrt{p^{(i)}_j q^{(i)}_j} \right) = 1 - \frac{1}{n}\sum_{i=1}^{n} \sum_{j=1}^b \sqrt{p^{(i)}_j q^{(i)}_j}
\end{align*}
and since there is a one-to-one correspondence between both representations given by $p^{(i)}_j=p'_{i\cdot b + j}$, i.e., given that the concatenation $\bm{p}'$ can be rewritten by reordering the elements of $\bm{p}$ as $\bm{p}'=(p^{(1)}_1,\ldots,p^{(1)}_b,\ldots,p^{(n)}_1,\ldots,p^{(n)}_b)$; analogously for $\bm{q},\bm{q}'$.

In any case, even while both mechanisms enable an optimization routine search for a good mixture parameter $\hat{\bm{\alpha}}$, the truth is that, for $n>2$, \emph{none of these procedures compute a proper divergence between two \new{PDFs}}, but rather some proxy of it. The remainder of this article is devoted to dealing with proper ways for modelling mixtures of distributions in the multiclass setup.

As a final note, 
even while aggregative quantifiers have typically led to higher performance \citep[see, e.g.,][]{schumacher2023comparative} a recent study by \citet{dussap2023} suggests that more sophisticated non-aggregative methods based on matching feature distributions can yield competitive performance. 
In this paper, we focus our attention on aggregative quantifiers only and leave the exploration of the non-aggregative type for further research.








\section{Pitfalls of Histograms on Multiclass Data}
\label{sec:pitfalls}

Histograms are useful tools for modelling univariate distributions and lie at the heart of many binary quantification techniques that have achieved remarkable success in the past \citep{Forman:2005fk,Forman:2008kx,Gonzalez-Castro:2013fk,Reis:2018fk,Maletzke:2019qd,Perez-Gallego:2019vl,maletzke2018need}.
However, in the multiclass setup, we are required to model multivariate distributions. In Section~\ref{sec:related}, we have discussed previous attempts that opt for breaking down the multivariate distributions as a series of univariate distributions that can then be modelled using standard histograms.
In this section, we argue that such an approach is inadequate for tackling multiclass problems.

We start by noting that a proper multivariate extension of histograms is possible in theory, but infeasible in practice.
The problem is that the number of bins one has to handle in multiple dimensions is doomed to grow 
\new{combinatorially}
with the number of dimensions. 
\new{More precisely, and assuming $b$ equally sized bins, every vector of posterior probabilities is mapped onto one out of the 
${b+n-1\choose n-1}$
valid combinations (i.e., combinations summing up to 1). For example, }
in the experiments we report in Section~\ref{sec:experiments} we consider a dataset with no less than $n=28$ categories; even for a moderately small number of bins (say, \new{$b=8$}) we would easily end up facing an intractable model (e.g., with \new{23,535,820} 
bins). 
A good implementation would make it possible anyway (e.g., by only allocating memory for the non-empty bins), but it is very likely that the vast majority of the bins will either remain empty or almost empty.

\nnew{A second limitation histograms exhibit has to do with the fact that the density mass centers are entirely agnostic to the data distribution. In other words, the centers of the bins are predetermined independently of the data. As a consequence, different choices of binning may lead to abrupt changes in the model. The reasons why this poses a limitation to the model are analogous to the reasons one may prefer a soft classifier over a hard classifier; the former provides more information to the model than the latter. By binning our distribution of posterior probabilities (soft assignments), we embrace a model of hard assignments (with bins acting as fine-grained classes) in which some information is necessarily lost.}

Before turning back to the class-wise histogram approach, let us note that, in the binary case, modelling only the posterior probabilities for one of the classes (say, for the \emph{positives}) suffices, since the other class (say, the \emph{negatives}) is constrained. This means that the histogram for one of the classes can be uniquely recovered from the other, and thus explicitly representing them both adds no useful information for the model. 

Following a similar reasoning, one may thus be tempted to think that, for $n>2$ classes, representing histograms for $(n-1)$ classes might suffice, since the remaining one should be similarly constrained. However, this is not true for the multiclass case. To see why, imagine we have $n=3$ classes and two \new{bags}, $A=\{a_1,a_2,a_3\}$ and $B=\{b_1,b_2,b_3\}$, whose elements have already been mapped into posterior probabilities (i.e.,  $a_i,b_i\in\Delta^{n-1}$) using some soft classifier $s$ returning 
$s(z)=(\textcolor{blue}{P(Y=1|X=z)}, \textcolor{olive}{P(Y=2|X=z)}, \textcolor{violet}{P(Y=3|X=z)})$ 
(note the color coding). Let $A$ and $B$ be defined as follows
\begin{align*}
\begin{matrix}
A=
\begin{Bmatrix}
a_1&=&(\textcolor{blue}{0.1}, & \textcolor{olive}{0.2}, & \textcolor{violet}{0.7})\\ 
a_2&=&(\textcolor{blue}{0.1}, & \textcolor{olive}{0.1}, & \textcolor{violet}{0.8})\\ 
a_3&=&(\textcolor{blue}{0.2}, & \textcolor{olive}{0.3}, & \textcolor{violet}{0.5})\\
\end{Bmatrix} , & 
B=
\begin{Bmatrix}
b_1&=&(\textcolor{blue}{0.1}, & \textcolor{olive}{0.3}, & \textcolor{violet}{0.6})\\ 
b_2&=&(\textcolor{blue}{0.1}, & \textcolor{olive}{0.2}, & \textcolor{violet}{0.7})\\ 
b_3&=&(\textcolor{blue}{0.2}, & \textcolor{olive}{0.1}, & \textcolor{violet}{0.7}) \\
\end{Bmatrix} .
\end{matrix} 
\end{align*}
\noindent Let us generate all $3$ class-wise histograms for each \new{bag}. Let $H_1,H_2,H_3$ be the histograms representing the distribution of posteriors $\textcolor{blue}{P(Y=1|X)},\textcolor{olive}{P(Y=2|X)},\textcolor{violet}{P(Y=3|X)}$, respectively, for $A$ and let $H'_1,H'_2,H'_3$ be the analogous histograms for $B$. The histograms would be defined by 
\begin{equation*}
\begin{matrix}
A':=
\begin{Bmatrix}
H_1=\mathrm{hist(\{\textcolor{blue}{0.1, 0.1, 0.2}\})}\\ 
H_2=\mathrm{hist(\{\textcolor{olive}{0.2, 0.1, 0.3}\})}\\ 
H_3=\mathrm{hist(\{\textcolor{violet}{0.7, 0.8, 0.5}\})}
\end{Bmatrix} , & 
B':=
\begin{Bmatrix}
H'_1=\mathrm{hist(\{\textcolor{blue}{0.1, 0.1, 0.2}\})}\\
H'_2=\mathrm{hist(\{\textcolor{olive}{0.3, 0.2, 0.1}\})}\\
H'_3=\mathrm{hist(\{\textcolor{violet}{0.6, 0.7, 0.7}\})}
\end{Bmatrix} .
\end{matrix}
\end{equation*}
\noindent Assume the number of bins is sufficiently fine grained (e.g., $b=10$).
Now, note the following facts:
\begin{enumerate}
    \item $H_1=H'_1$,
    \item $H_2=H'_2$ since histograms are permutation-invariant functions,
    \item $H_3\neq H'_3$.
\end{enumerate}
This seems to indicate that a bag of posterior probabilities cannot be meaningfully represented by means of $(n-1)$ histograms only, \new{because $H_3$ can be described either as a function of $H_1$ or as a function of $H_2$} (e.g., \new{the bag} $A$ is not well represented by $\{H_1,H_2\}$) since such a representation would fail in distinguishing between $A$ and other \new{bags} (e.g., $B$). 

This, however, does not pose a serious problem for the model; after all, one can simply represent a bag by means of $n$ histograms, instead of by means of $(n-1)$ histograms. \nnew{Nevertheless, this seemingly harmless asymmetry in the method's behaviour is actually a symptom of a more serious illness, i.e., that \emph{the model has become unable to retain class-class information}.}

As a consequence, note that even using $n$ histograms, different bags may become indistinguishable for the model. This issue has nothing to do with the resolution of the histogram, i.e., with the number of bins (should this be the case, we could simply increase $b$ at will), but \nnew{is rather a direct effect of} the fact that the relations between the classes get lost in the class-wise histogram representation. 
To see why, imagine the following case, with $n=4$
\begin{align*}
\begin{matrix}
A=
\begin{Bmatrix}
a_1&=&(\textcolor{blue}{0.1}, & \textcolor{olive}{0.2}, & \textcolor{violet}{0.3}, & \textcolor{teal}{0.4})\\ 
a_2&=&(\textcolor{blue}{0.2}, & \textcolor{olive}{0.3}, & \textcolor{violet}{0.4}, & \textcolor{teal}{0.1})\\ 
a_3&=&(\textcolor{blue}{0.3}, & \textcolor{olive}{0.4}, & \textcolor{violet}{0.1}, & \textcolor{teal}{0.2})\\
\end{Bmatrix} , & 
B=
\begin{Bmatrix}
b_1&=&(\textcolor{blue}{0.1}, & \textcolor{olive}{0.3}, & \textcolor{violet}{0.4}, & \textcolor{teal}{0.2})\\ 
b_2&=&(\textcolor{blue}{0.3}, & \textcolor{olive}{0.2}, & \textcolor{violet}{0.1}, & \textcolor{teal}{0.4})\\ 
b_3&=&(\textcolor{blue}{0.2}, & \textcolor{olive}{0.4}, & \textcolor{violet}{0.3}, & \textcolor{teal}{0.1})\\
\end{Bmatrix} ,
\end{matrix}
\end{align*}
and note that the corresponding histogram-based representation would be
\begin{equation*}
\begin{matrix}
A':=
\begin{Bmatrix}
H_1=\mathrm{hist(\{\textcolor{blue}{0.1, 0.2, 0.3}\})}\\ 
H_2=\mathrm{hist(\{\textcolor{olive}{0.2, 0.3, 0.4}\})}\\ 
H_3=\mathrm{hist(\{\textcolor{violet}{0.3, 0.4, 0.1}\})}\\
H_4=\mathrm{hist(\{\textcolor{teal}{0.4, 0.1, 0.2}\})}\\
\end{Bmatrix} , & 
B':=
\begin{Bmatrix}
H'_1=\mathrm{hist(\{\textcolor{blue}{0.1, 0.3, 0.2}\})}\\
H'_2=\mathrm{hist(\{\textcolor{olive}{0.3, 0.2, 0.4}\})}\\
H'_3=\mathrm{hist(\{\textcolor{violet}{0.4, 0.1, 0.3}\})}\\
H'_4=\mathrm{hist(\{\textcolor{teal}{0.2, 0.4, 0.1}\})}\\
\end{Bmatrix} ,
\end{matrix}
\end{equation*}
i.e., that $A'\equiv B'$. As a way of example, notice that there is pattern in $A$, i.e., ``the posterior for class $y=1$ (blue) is always slightly smaller than for class $y=2$ (pale green)'', which is not captured in the representation $A'$ (indeed, note that this representation is identical to $B'$, even though the pattern is not characteristic of $B$).

\nnew{Summing up, a histogram-based representation for quantification presents the following limitations in the multiclass case: 
\begin{enumerate}
    \item A native extension incurs a computational cost that scales combinatorially with the number of classes and bins.
    \item The hard assignment to data-agnostic bin centers sacrifices information.
    \item A set of class-specific histograms fails to capture inter-class information.
\end{enumerate}
}



In Section~\ref{sec:method} we propose a representation mechanism for multivariate distributions that \nnew{overcomes all these limitations.} 
In Section~\ref{sec:experiments} we present empirical evidence demonstrating that this mechanism effectively leads to improved quantification performance.





\section{Method}
\label{sec:method}

\begin{figure}[h]
    \centering
    
\newlength\q
\setlength\q{\dimexpr .25\textwidth -2\tabcolsep}
\noindent\begin{tabular}{p{\q}p{\q}p{\q}p{\q}}
\centering
(A) Representation for $L_1$ & \centering (B) Representation for $L_2$ & \centering (C) Representation for $L_3$ & \centering (D) Representation for $U$ \\
\end{tabular}

    \includegraphics[width=\textwidth]{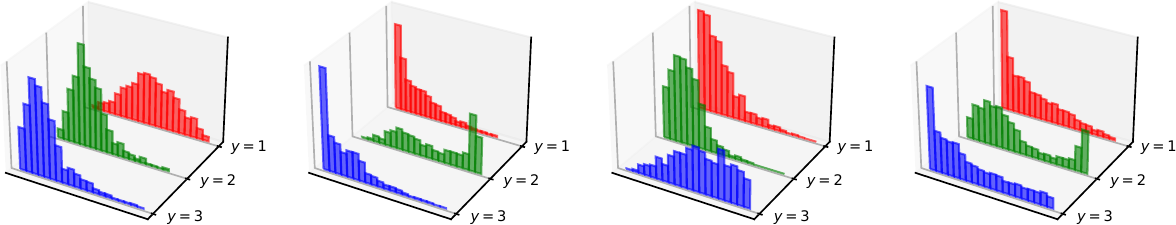} 
    \\
    \vspace{0.3cm}
    \includegraphics[width=\textwidth]{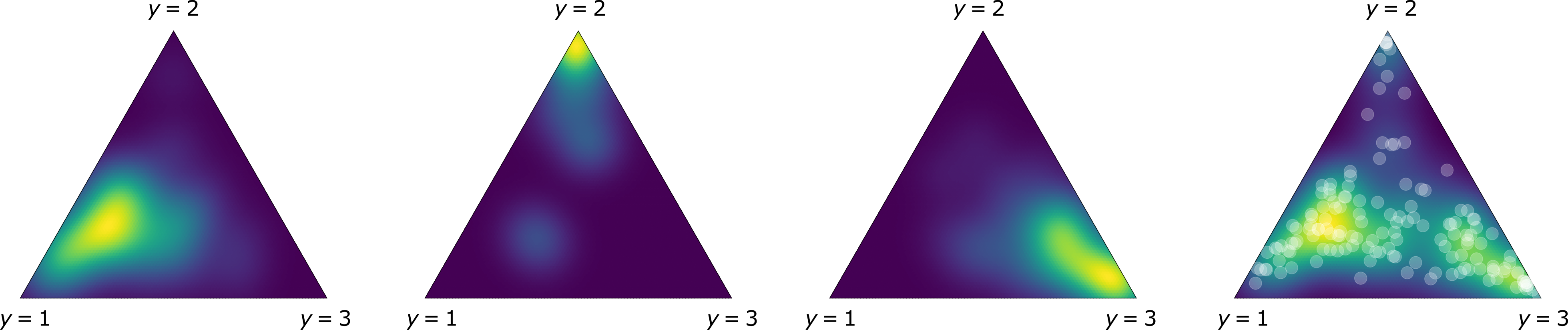}
    \caption{Problem representations obtained with traditional class-wise histograms (first row) and with our proposed mechanism based on GMMs (second row) on a 3-class sentiment problem (the dataset is called ``wb'' and belongs to the ``Tweets'' group described later in Section \ref{sec:datasets}). The first three columns (A,B,C) show the model representations for the training sets $L_1$, $L_2$, and $L_3$, while the last column (D) shows the representation for the test set $U$. The quantification problem is framed as the task of reconstructing (D) as a convex linear combination of (A,B,C).}
    \label{fig:representation}
\end{figure}

In this section, we turn to describe our KDE-based solution to the multiclass quantification problem. 
In a nutshell, the main idea of our method consists of switching the problem representation from  discrete, univariate PDFs (histograms) of the posterior probabilities to continuous, multivariate PDFs on the unit $(n-1)$-simplex. Our PDFs will be represented by means of Gaussian Mixture Models (GMMs) obtained via KDE (Figure~\ref{fig:representation}).

In Section~\ref{sec:repr}, we describe how to represent bags as GMMs using KDE. Then, we show two different ways of framing the optimization problem: one based on distribution matching (Section~\ref{sec:dm-solution}) and another based on maximum likelihood (Section~\ref{sec:ml-solution}).

\subsection{KDE-based representation}
\label{sec:repr}

Let $p_\theta$ be the kernel density estimator defined by 
\begin{align}
\label{eq:kde}
p_\theta(\bm{x}) = \frac{1}{|X|} \sum_{\bm{x}_i\in X} K\left(\frac{\bm{x}-\bm{x}_i}{h}\right) ,
\end{align}
where $\theta=\{K,h,X\}$ are parameters of the model, with $K$ the kernel function controlled by the bandwidth $h$, and $X$ the set of reference points. 
The different density functions we will deal with all share the components $K$ and $h$ (that we treat as model hyperparameters) and differ only on the set of reference points. We will thus alleviate notation by simply writing $p_X$, instead of $p_\theta$ or $p_{\{K,h,X\}}$.

We will concentrate our attention on Gaussian kernels, i.e., the case in which the kernel $K\left(\frac{\bm{x}-\bm{x}_i}{h}\right)=K(\bm{x}; \bm{x}_i, h)$ corresponds to the multivariate normal distribution $\mathcal{N}(\bm{x}|\bm{\mu},\Lambda^{-1})$ with mean $\bm{\mu}=\bm{x}_i$ and covariance matrix $\Lambda^{-1}=h^2I_D$, with $I_D$ the identity matrix of order $D$, that is
\begin{equation*}
    K(\bm{x}; \bm{x}_i, h) = \frac{1}{h^{D}(2\pi)^{\frac{D}{2}}}\exp{\left(-\frac{||\bm{x}-\bm{x}_i||^2_2}{2h^2}\right)} .
\end{equation*}

Note also that $D$ is the dimensionality of the input space, i.e., that $\bm{x}\in\mathbb{R}^D$. We here wrote $D$, and not $d$ as in the notation of Section~\ref{sec:notation}, on purpose, since we are not here assuming to be modelling the input space $\mathcal{X}=\mathbb{R}^d$ in which our covariates live, but rather the probability simplex $\widetilde{\mathcal{X}}=\Delta^{n-1}$. 
For this reason, we first map every datapoint into a posterior probability by means of a soft classifier $s$.
Note, thus, that in our case this translates into $D=n$ since our posterior probabilities have $n$ dimensions (although only $n-1$ degrees of freedom). To ease notation, we simply overline the symbol denoting a set of examples to indicate that its members have been converted into posterior probabilities; e.g., by $\widetilde{L}_i$ we indicate the set $\{s(\bm{x}):\bm{x}\in L_i\}$ where $L_i=\{\bm{x} : (\bm{x}, y)\in L \wedge y=i\}$ as before. This mapping is carried out via $k$-fold cross-validation in the training set. 

At training time, the quantifier needs to model a dedicated density function for each of the class-conditional distributions of posterior probabilities. Given a set of reference points for each class $(\widetilde{L}_1, \widetilde{L}_2, \ldots, \widetilde{L}_n)$ and a mixture vector $\bm{\alpha}=(\alpha_1,\alpha_2,\ldots,\alpha_n)\in\Delta^{n-1}$, we thus define $\bm{p}_{\bm{\alpha}}:\Delta^{n-1}\rightarrow\mathbb{R}_{\geq 0}$ as 
\begin{align}
\label{eq:kdemixture}
\bm{p}_{\bm{\alpha}}(\widetilde{\bm{x}}) = \sum_{i=1}^n \alpha_i p_{\widetilde{L}_i}(\widetilde{\bm{x}}) ,
\end{align}
where 
$\widetilde{\bm{x}}$ 
is a vector of posterior probabilities.
For a fixed $\bm{\alpha}$ and given a soft classifier $s$, the density of a datapoint $\bm{x}\in\mathcal{X}$ \nnew{(i.e., a vector of covariates)} is estimated by the KDE mixture  as $\bm{p}_{\bm{\alpha}}(s(\bm{x}))$. Note that, since we chose our kernel to be Gaussian, $\bm{p}_{\bm{\alpha}}$ is a also GMM.

\nnew{GMMs are less sensitive to the choice of the bandwidth than histograms are to the choice of binning. The latter follows from the fact that the density-mass ``blocks'' become smooth when using Gaussian kernels, and from the fact that the centers of these are not data-agnostic, as in histograms. In other words, a GMM in which the Gaussians are centered at the datapoints is akin to a model of soft assignments (the location of a datapoint influences the density estimation in any other location of the space) conversely to the hard assignments of histograms (the location of a datapoint affects one and only one bin in the model).}

Finally, note that replacing histograms with GMMs brings about the following \nnew{ advantages with respect to the problems discussed in Section~\ref{sec:pitfalls}:
\begin{enumerate}
    \item GMMs scale smoothly with the number of classes. 
    \item GMMs rely on soft assignments, which effectively retain more information.
    \item Inter-class correlations are preserved.
\end{enumerate}
}


We will use ``KDEy'' as a prefix for the variants we propose in the following sections.



\subsection{Distribution Matching framework}
\label{sec:dm-solution}

The distribution matching approach addresses the following optimization problem
\begin{align}
\label{eq:dm}
\hat{\bm{\alpha}} &= \argmin_{\bm{\alpha}\in\Delta^{n-1}} \mathcal{D}(\bm{p}_{\bm{\alpha}}||q_{\widetilde{U}}) ,
\end{align}
where $q_{\widetilde{U}}$ is the KDE function (hence, a GMM) modelled on the reference set $\widetilde{U}=\{s(\bm{x}): \bm{x} \in U \}$ over the unlabelled test set $U$, and where $\mathcal{D}$ is any divergence function measuring the degree of discrepancy between the mixture distribution and the test distribution. 

Note that we are dealing with \emph{continuous} density functions, and not discrete probability functions as in previously proposed DM approaches (Section~\ref{sec:related}). 
This means that evaluating typical divergences (such as HD, Topsøe,  KLD, L2) requires handling an integral over the simplex $\Delta^{n-1}$; this process can be computationally expensive and may make it impractical to use within any standard optimization routine to find $\hat{\bm{\alpha}}$.
In order to overcome this limitation, we propose two alternative solutions: one consisting of Monte Carlo approximation (Section~\ref{sec:sec:monte}) and another based on a closed-form solution (Section~\ref{sec:sec:closed}).

\subsubsection{Monte Carlo approximation}
\label{sec:sec:monte}

Popular divergences previously used in the quantification literature (such as HD, Topsøe, KLD, L2), do not admit a closed-form solution when the densities functions are GMMs \citep{HersheyO07,nielsen2016guaranteed, nielsen2012closedformGMM,krishnamurthy2015estimating}.
An alternative solution comes down to numerical approximation via Monte Carlo sampling \citep{nielsen2020non}. \alex{Esa cita no parece una publicación.}

Most of these divergences belong to the family of the so-called $f$-divergences, i.e.,  \new{a subfamily of} divergences $\mathcal{D}$ that can be \new{computed} as
\begin{equation}
    \label{eq:fdiv2}
    \mathcal{D}_f(p||q)=\int_{\Omega} f\left(\frac{p(x)}{q(x)}\right) q(x) d\mu(x) ,
\end{equation}
where $p$ and $q$ are the density functions of the two probability distributions, 
and $f$ is a special type of convex function called the ``generator function''. 
In our case, $\mu$ is the standard Lebesgue measure and $\Omega$ is the probability simplex $\Delta^{n-1}$.
Different choices for $f$ give rise to well-known divergences like the (reverse) KLD by choosing $f(u)=u\log u$, the $\operatorname{HD}^2$ by choosing $f(u)=(\sqrt{u}-1)^2$, or the Jensen-Shannon divergence (JS) by choosing $f(u)=-(u+1)\log \frac{u+1}{2}+u\log u$.
Note that \eqref{eq:fdiv2} corresponds to
\begin{align}
\mathcal{D}_f(p||q)= \mathbb{E}_q\left[f\left(\frac{p(x)}{q(x)}\right)\right] ,   
\end{align}
and so the Monte Carlo approximation is given by
\begin{equation}
\label{eq:montecarlo}
    \hat{D}_f(p||q)= \frac{1}{t} \sum_{i=1}^t f\left(\frac{p(x_i)}{q(x_i)}\right) ,
\end{equation}
where $t$ is the number of \emph{trials} and $x_1,\ldots,x_t\sim_{\mathrm{iid}} q$. 
The right-most distribution from which iid samples are to be drawn is called the \emph{reference distribution}. As long as $q$ is a GMM, we could resort to well-established procedures for drawing iid samples directly from it. However, there are practical reasons why we might prefer taking, as our reference distribution, a different one, and resort to \emph{importance sampling} instead. This leads us to compute
\begin{equation}
\label{eq:montecarlo2}
    \hat{D}_f(p||q)= \frac{1}{t} \sum_{i=1}^t f\left(\frac{p(x_i)}{q(x_i)}\right) \frac{q(x_i)}{r(x_i)} ,
\end{equation}
in which $x_1,\ldots,x_t\sim_{\mathrm{iid}} r$ with $r$ our reference distribution. 

The rationale behind opting for this course of action is that we are required to compute $\hat{D}_f(\bm{p}_{\bm{\alpha}}||q_{\widetilde{U}})$, but since  $q_{\widetilde{U}}$ can be modelled exclusively at test time, this would force us to postpone the entire computational burden of every inference we carry out to the test phase. In quantification evaluation, it is nowadays a standard practice to confront every quantification system against \emph{many} test bags (this is explained in more detail in Section~\ref{sec:eval}), which would render this inference procedure impractical.
Given that divergence measures are not necessarily symmetric one may be tempted to compute $\hat{D}_f(q_{\widetilde{U}}||\bm{p}_{\bm{\alpha}})$ instead (in principle, there are no grounds to favor one over another) so as to take our reference distribution be $\bm{p}_{\bm{\alpha}}$. This, however, is not a way out of this problem, because at training time we only know the mixture components $p_{\widetilde{L}_1},\ldots,p_{\widetilde{L}_n}$, but not the mixture parameter $\bm{\alpha}$ which needs to be explored as part of the optimization procedure at test time.

\new{We thus propose to approximate the divergence via stochastic Monte Carlo sampling using importance sampling and considering, as our reference distribution, the function $\bm{p}_{\bm{u}}$ in which $\bm{u}=\left(\frac{1}{n},\frac{1}{n},\ldots,\frac{1}{n}\right)$ is the uniform prevalence vector, and applying \eqref{eq:kdemixture}. Note that, acting this way, we are able to anticipate most of the computational cost to the training phase.
Not only the datapoints $\widetilde{\bm{x}}_1,\ldots,\widetilde{\bm{x}}_t$ can be pre-sampled\footnote{\nnew{In this case, we write $\widetilde{\bm{x}}_i$ (instead of $\bm{x}_i$) to indicate that the datapoints lie in $\Delta^{n-1}$. However, note these datapoints are directly sampled from the GMM, and are not the output of a soft classifier.}} at training time once and for all, but also the densities of each mixture component can be pre-computed. During training, we thus allocate in memory a matrix $\bm{M}\in\mathbb{R}^{t\times n}$ with the class-conditional densities $\bm{M}[i,j]=p_{j}(\widetilde{\bm{x}}_i)$ and pre-compute $r(\widetilde{\bm{x}}_i)\equiv \bm{p}_{\bm{u}}(\widetilde{\bm{x}}_i)$ simply by taking the mean of the rows in $\bm{M}$. From here, all that remains ahead to compute \eqref{eq:montecarlo2} is modelling $q_{\widetilde{U}}$ via KDE, computing $q_{\widetilde{U}}(\widetilde{\bm{x}}_i)$ for the previously sampled datapoints $\widetilde{\bm{x}}_1,\ldots,\widetilde{\bm{x}}_t$, and computing the densities $\bm{p}_{\bm{\alpha}}(\widetilde{\bm{x}}_i)$ for $\widetilde{\bm{x}}_1,\ldots,\widetilde{\bm{x}}_t$ for every $\bm{\alpha}$ explored. Although the last computation needs to be performed at every iteration of the optimization search, it is straightforward since it now reduces to a matrix multiplication ($\bm{M}\bm{\alpha}$), an operation for which current hardware is highly optimized.}

For the Monte Carlo implementation, we chose the $\operatorname{HD}^2$ as our divergence measure since it has proven good quantification performance in the past \citep{Gonzalez-Castro:2013fk}. We do not consider the Topsøe distance, which has recently been shown to perform slightly better \citep{Maletzke:2019qd}, for the following reason. The Topsøe distance is defined as twice the Jensen-Shannon divergence, which in turn is defined as
\begin{equation*}
    \mathcal{D}_{\mathrm{JS}}(p||q)=\frac{1}{2}\left( \mathcal{D}_{\mathrm{KL}}(p||r)+\mathcal{D}_{\mathrm{KL}}(q||r) \right) ,
\end{equation*}
where $r$ is the reference distribution defined as the mixture $r=\frac{1}{2}(p+q)$. This implies that, in order to generate Monte Carlo samples, we need to sample from $r$, and this forces us to defer the entire process to the test phase, since only then would the mixutre $r$ be determined.

We denote this variant KDEy-HD.






\subsubsection{Closed-Form solutions}
\label{sec:sec:closed}

One of the reasons why a closed-form solution does not exist for most commonly used divergences is the presence of \emph{log-sum terms}, that cannot be factored out from the integral \citep{michalowicz2008calculation}. However, other divergence measures exist that do not involve log-sum terms. One such example is the Cauchy-Schwarz divergence ($\mathcal{D}_{\mathrm{CS}}$), defined as
\begin{equation*}
    \label{eq:cauchyschwarzdiv}
    \mathcal{D}_{\mathrm{CS}}(p||q)=-\log\left(\frac{\int p(\bm{x})q(\bm{x})dx}{\sqrt{\int p(\bm{x})^2dx \int q(\bm{x})^2dx}}\right) .
\end{equation*}

\citet{kampa2011closed} show that, when $p$ and $q$ are both defined as the GMMs given by
\begin{align*}
    p(\bm{x})&=\sum_{m=1}^N \pi_m  \mathcal{N}(\bm{x}|\bm{\mu}_m, \Lambda_m^{-1}) , \\
    q(\bm{x})&=\sum_{k=1}^M \tau_k \mathcal{N}(\bm{x}|\bm{\nu}_k, \Omega_k^{-1}) ,
\end{align*}
where $\pi_m,\tau_k$ are the mixture parameters, $\bm{\mu}_m,\bm{\nu}_k$ are the means at which the Gaussians distributions are centered, and $\Lambda_m^{-1},\Omega_k^{-1}$ are the covariance matrices of the Gaussian distributions, then a closed-form solution for $\mathcal{D}_{\mathrm{CS}}$ exists, which is given by\condfootnote{The eagle-eyed reader might have spot a difference between \eqref{eq:Dcs} and the one originally reported in \citet[Equation 3]{kampa2011closed}. Specifically, the terms $(4\pi)^{\frac{D}{2}}$ in our formulation replace the terms $(2\pi)^{\frac{D}{2}}$ in the original paper. The reason why is explained in more detail in Appendix \ref{app:formula}.}
\begin{align}
\begin{split}
\label{eq:Dcs}
    \mathcal{D}_{\mathrm{CS}}(p||q) 
        =& -\log\left( \sum_{m=1}^N \sum_{k=1}^M \pi_m \tau_k z_{mk} \right) \\
        & +\frac{1}{2}\log\left( \sum_{m=1}^N \frac{\pi_m^2 |\Lambda_m|^{\frac{1}{2}}}{(\condfactor)^{\frac{D}{2}}} + 2\sum_{m=1}^N \sum_{m'<m} \pi_m \pi_{m'}z_{mm'} \right) \\
        & +\frac{1}{2}\log\left( \sum_{k=1}^M \frac{\tau_k^2 |\Omega_k|^{\frac{1}{2}}}{(\condfactor)^{\frac{D}{2}}} + 2\sum_{k=1}^M \sum_{k'<k} \tau_k \tau_{k'}z_{kk'} \right) ,
\end{split}
\end{align}
where
\begin{align}
\begin{split}
    z_{mk} &= \mathcal{N}(\bm{\mu}_m | \bm{\nu}_k, (\Lambda_m^{-1}+\Omega_k^{-1})) , \\
    \label{eq:zmm}
    z_{mm'} &= \mathcal{N}(\bm{\mu}_m | \bm{\mu}_{m'}, (\Lambda_m^{-1}+\Lambda_{m'}^{-1})) , \\
    z_{kk'} &= \mathcal{N}(\bm{\nu}_k | \bm{\nu}_{k'}, (\Omega_k^{-1}+\Omega_{k'}^{-1})) .
\end{split}
\end{align}

Their derivation covers the general case in which each Gaussian component can be characterized by a dedicated covariance matrix ($\Lambda_m^{-1}$ and $\Omega_{k}^{-1}$) and a dedicated mixture parameter ($\pi_m$ and $\tau_k$).
When we plug in our training and test density functions $\bm{p}_{\bm{\alpha}}$ and $q_{\widetilde{U}}$, the equation simplifies a lot.
Before we show why, first note that the second and third factors of \eqref{eq:Dcs} can be re-written in a more compact (less computationally efficient) form
\begin{align}
\begin{split}
\label{eq:Dcs2}
    \mathcal{D}_{\mathrm{CS}}(p||q) 
        =& -\log\left( \sum_{m=1}^N \sum_{k=1}^M \pi_m \tau_k z_{mk} \right) \\
        & +\frac{1}{2}\log\left( \sum_{m=1}^N \sum_{m'=1}^N \pi_m \pi_{m'}z_{mm'} \right) \\
        & +\frac{1}{2}\log\left( \sum_{k=1}^M \sum_{k'=1}^M \tau_k \tau_{k'}z_{kk'} \right) ,
\end{split}
\end{align}
in which the trivial components from the diagonal terms in the summations ($m=m'$ and $k=k'$) are not factored out, and in which the duplicated symmetric computations ($m'\geq m$ and $k'\geq k$) are not avoided.
Now note that the GMMs generated via KDE all share a common covariance matrix, i.e., $\Lambda_m^{-1}=\Omega_{k}^{-1}=h^2I_D$, for all $m,k$ and that, in our case \nnew{$N=|L|=\sum_{i=1}^n |L_i|$ is the number of training instances and $M=|U|$ is the number of test instances (as by the definitions of Section~\ref{sec:notation}) while $D=n$ is the dimensionality of the vectors of posterior probabilities that we model (see Section~\ref{sec:method}).} Furthermore, for our test distribution \new{$q_{\widetilde{U}}$} we have $\tau_k=\frac{1}{|U|}$ for all $k$, while for the training mixture components $p_{\widetilde{L}_i}$ within $\bm{p}_{\bm{\alpha}}$, these weights are fixed to $\frac{\alpha_i}{|L_i|}$.
We can thus rewrite \eqref{eq:Dcs2} as 
\begin{align}
\begin{split}
\label{eq:Dcs3}
    \mathcal{D}_{\mathrm{CS}}(\bm{p}_{\bm{\alpha}}||q_{\widetilde{U}}) 
        =& -\log\left( \sum_{i=1}^n \sum_{j=1}^{|L_i|}  \sum_{k=1}^{|U|}  \frac{\alpha_i}{|L_i|}\frac{1}{|U|} K(\widetilde{\bm{x}}_j^{(i)} ; \widetilde{\bm{x}}_k, 2h) \right) \\
        & +\frac{1}{2}\log\left( \sum_{i=1}^n \sum_{j=1}^{|L_i|} \sum_{i'=1}^n \sum_{j'=1}^{|L_{i'}|} \frac{\alpha_i}{|L_i|} \frac{\alpha_{i'}}{|L_{i'}|} K(\widetilde{\bm{x}}_j^{(i)} ; \widetilde{\bm{x}}_{j'}^{(i')}, 2h) \right) \\
        & +\frac{1}{2}\log\left( \sum_{k=1}^{|U|} \sum_{k'=1}^{|U|} \frac{1}{|U|^2} K(\widetilde{\bm{x}}_k ; \widetilde{\bm{x}}_{k'}, 2h) \right) ,
\end{split}
\end{align}
where $\widetilde{\bm{x}}_j^{(i)}$ refers to the $j$th training example in the reference set \new{$\widetilde{L}_i$}, and $\widetilde{\bm{x}}_k$ refers to the $k$th test example in the reference set \new{$\widetilde{U}$}. 
We can now express \eqref{eq:Dcs3} more conveniently in matrix form as
\begin{align}
\label{eq:Dcs4}
    \mathcal{D}_{\mathrm{CS}}(\bm{p}_{\bm{\alpha}}||q_{\widetilde{U}}) 
        = -\log\left( \sum_{i=1}^n  \bm{r}_i^\top A_{i} \bm{t} \right) 
         +\frac{1}{2}\log\left( \sum_{i=1}^n \sum_{i'=1}^n \bm{r}_i^\top B_{ii'} \bm{r}_{i'} \right) 
        +\frac{1}{2}\log\left( \bm{t}^\top C \bm{t}\right) ,
\end{align}
where $\bm{r}_i$ and $\bm{t}$ are column vectors that are obtained by multiplying the vector of all ones (with appropriate dimensions) by scalars $\frac{\alpha_i}{|L_i|}$ and $\frac{1}{|U|}$, respectively, and where $A_i, B_{ii'}, C$ are the (train-test, train-train, and test-test) Gram matrices of kernel evaluations such that
\begin{align*}
A_i[j,k] & =K(\widetilde{\bm{x}}_j^{(i)}; \widetilde{\bm{x}}_k, 2h) , \\
B_{ii'}[j,j'] & = K(\widetilde{\bm{x}}_j^{(i)} ; \widetilde{\bm{x}}_{j'}^{(i')}, 2h) , \\
C[k,k'] & = K(\widetilde{\bm{x}}_k ; \widetilde{\bm{x}}_{k'}, 2h) . 
\end{align*}

The closed-form variant of KDE, that we dub KDEy-CS, is highly efficient, since (i) the tensor containing all train-train matrices $B_{ii'}$ (expected to be the most computationally expensive term in the formula) can be entirely pre-computed at training time; and (ii) tensor $B$ and matrix $C$ are symmetric, meaning that only the upper triangular elements must be computed. 

The method is also efficient in terms of memory footprint since there is no need to retain the whole Gram matrices in memory. Notably, \eqref{eq:Dcs4} can be computed in its entirety by simply retaining the summation of the elements within each Gram matrix; for example, the term $\bm{r}_i^\top A_{i} \bm{t}$ can be computed as $\left(\frac{\alpha_i}{|L_i|}\right) \times \left(\sum_{j=1}^{|L_i|}\sum_{k=1}^{|U|}A_i[j,k]\right)\times\left(\frac{1}{|U|}\right)$. 
We can therefore further simplify \eqref{eq:Dcs4} as
\begin{equation}
    \label{eq:matrixform_cs}
    \mathcal{D}_{\mathrm{CS}}(\bm{p}_{\bm{\alpha}}||q_{\widetilde{U}}) 
        = -\log\left( \bar{\bm{r}}^\top \bar{\bm{a}} \cdot t \right) 
         +\frac{1}{2}\log\left( \bar{\bm{r}}^\top \bar{\bm{B}} \bar{\bm{r}} \right) 
        +\frac{1}{2}\log\left( t^2 \bar{c} \right) ,
\end{equation}
where 
$\bar{\bm{a}}\in\mathbb{R}^n$ are the sums of the train-test Gram matrices 
$\bar{\bm{a}}[i] = \sum_{j=1}^{|L_i|}\sum_{k=1}^{|U|} A_i[j,k]$; 
and $\bar{\bm{B}}\in\mathbb{R}^{n\times n}$ are the sums of train-train Gram matrices 
$\bar{\bm{B}}[i,i'] = \sum_{j=1}^{|L_i|} \sum_{j'=1}^{|L_{i'}|} B_{ii'}[j,j']$; 
and $\bar{\bm{r}}\in\mathbb{R}^n$ are ratios $\bar{\bm{r}}[i] = \frac{\alpha_i}{|L_i|}$;
and $\bar{c}\in\mathbb{R}$ is the sum of the test-test Gram matrix 
$\bar{c} = \sum_{k=1}^{|U|}\sum_{k'=1}^{|U|} C[k,k']$; 
and $t\in\mathbb{R}$ is the constant $t =\frac{1}{|U|}$.

It is worth noting that the search 
for the optimal $\hat{\bm{\alpha}}$ in \eqref{eq:dm} becomes very efficient when the CS divergence is adopted, as it only involves calculations with two vectors $\bar{\bm{a}},\bar{\bm{r}}\in\mathbb{R}^n$, one matrix $\bar{\bm{B}}\in\mathbb{R}^{n\times n}$, and two scalars $\bar{c}$ and $t$. \new{Finally, note that the last term of \eqref{eq:matrixform_cs} plays no role in the distribution matching optimization procedure of \eqref{eq:dm}, since it does not depend on $\bm{\alpha}$ (only terms involving $\overline{\bm{r}}$ do) and thus becomes a constant that can be ignored (i.e., there is no need to compute $C$). The final optimization procedure is given by}
\begin{align}
\hat{\bm{\alpha}} &= \argmin_{\bm{\alpha}\in\Delta^{n-1}} \left\{ -\log\left( \bar{\bm{r}}^\top \bar{\bm{a}} \cdot t \right) 
         +\frac{1}{2}\log\left( \bar{\bm{r}}^\top \bar{\bm{B}} \bar{\bm{r}} \right) \right\} .
\end{align}

In this section, we have presented KDEy-CS, based on the derivation for the Cauchy-Schwarz divergence of \citet{kampa2011closed}. 
Other closed-form derivations 
are discussed by \citet{wang2009closed,nielsen2012closedformGMM}, that could be applied within our framework as well.

\subsection{Maximum Likelihood framework}
\label{sec:ml-solution}

In this section, we investigate the implications of adopting, as our divergence function, the Kullback-Leibler divergence (the quintessential divergence in statistics). This derivation is similar to the one by \citet{Garg:2020jt}, but we present it here adapted to our problem setting.

For reasons that will become clear later, we will analyze the case for $\mathcal{D}_{\mathrm{KL}}(q_{\widetilde{U}}||\bm{p}_{\bm{\alpha}})$, i.e., we will take $\bm{p}_{\bm{\alpha}}$ to be our reference distribution instead of $q_{\widetilde{U}}$. Since the KLD is an $f$-divergence, we can plug in the corresponding generator function $f(u)=u\log u$ in \eqref{eq:fdiv2} as follows
\begin{align}
\label{eq:dklinv}
\begin{split}
    \mathcal{D}_{\mathrm{KL}}(q_{\widetilde{U}}||\bm{p}_{\bm{\alpha}}) &= \int \log\frac{q_{\widetilde{U}}(\widetilde{\bm{x}})}{\bm{p}_{\bm{\alpha}}(\widetilde{\bm{x}})} q_{\widetilde{U}}(\widetilde{\bm{x}}) dx \\
    &= \mathbb{E}_{q_{\widetilde{U}}} \left[ \log\frac{q_{\widetilde{U}}(\widetilde{\bm{x}})}{\bm{p}_{\bm{\alpha}}(\widetilde{\bm{x}})} \right] .
\end{split}
\end{align}

Minimizing \eqref{eq:dklinv} as part of the distribution matching framework of \eqref{eq:dm} thus amounts to the following optimization problem
\begin{align*}
\begin{split}
    \hat{\bm{\alpha}} &= \argmin_{\bm{\alpha}\in\Delta^{n-1}} \mathbb{E}_{q_{\widetilde{U}}} \left[ \log q_{\widetilde{U}}(\widetilde{\bm{x}}) \right] - \mathbb{E}_{q_{\widetilde{U}}} \left[ \log \bm{p}_{\bm{\alpha}}(\widetilde{\bm{x}}) \right] \\
    &= \argmin_{\bm{\alpha}\in\Delta^{n-1}} - \mathbb{E}_{q_{\widetilde{U}}} \left[ \log \bm{p}_{\bm{\alpha}}(\widetilde{\bm{x}}) \right] .
\end{split}
\end{align*}

The last simplification follows from the fact that the first expectation does not depend on $\bm{\alpha}$. In order to estimate the remaining expectation we can resort to a Monte Carlo approach (see Section~\ref{sec:sec:monte}), this time by drawing samples $\widetilde{\bm{x}}_1,\ldots,\widetilde{\bm{x}}_t$ iid from $q_{\widetilde{U}}$. However, note that in this case, we are not requested to model $q_{\widetilde{U}}$ at all (the optimization procedure does not depend on it by any means) and therefore we can directly use $\widetilde{U}$ as an iid sample of the test distribution (which is the reason why we were interested in computing $\mathcal{D}_{\mathrm{KL}}(q_{\widetilde{U}}||\bm{p}_{\bm{\alpha}})$ from the beginning).\footnote{\nnew{If we had needed to compute evaluations $q_{\widetilde{U}}(\widetilde{\bm{x}})$, then it would not be appropriate to use $\widetilde{U}$ as our sampling (trials). The reason is that $q_{\widetilde{U}}$ is modelled using this exact data sample, and therefore the density evaluations would be strongly biased.}} This leads to the following optimization problem
\begin{align}
\begin{split}
\label{eq:kdeml}
    \hat{\bm{\alpha}} &\approx \argmin_{\bm{\alpha}\in\Delta^{n-1}} -\frac{1}{|U|} \sum_{\widetilde{\bm{x}}\in\widetilde{U}} \log \bm{p}_{\bm{\alpha}} (\widetilde{\bm{x}}) \\
    &= \argmin_{\bm{\alpha}\in\Delta^{n-1}} -\sum_{\widetilde{\bm{x}}\in\widetilde{U}} \log \sum_{i=1}^n \alpha_i p_{\widetilde{L}_i} (\widetilde{\bm{x}}) . \\
\end{split}
\end{align}

We solve \eqref{eq:kdeml} using standard optimization routines. This optimization is efficient since the densities $p_{\widetilde{L}_i} (\widetilde{\bm{x}})$ can be computed once and for all before the optimization procedure starts. Finally, note that this minimization problem is equivalent to maximizing
%
%
%
\begin{align*}
\begin{split}
    \hat{\bm{\alpha}} 
    &=  \argmax_{\bm{\alpha}\in\Delta^{n-1}}   \prod_{\widetilde{\bm{x}}\in\widetilde{U}} \sum_{i=1}^n \alpha_i p_{\widetilde{L}_i} (\widetilde{\bm{x}}) \\
    &\approx  \argmax_{\bm{\alpha}\in\Delta^{n-1}}   \prod_{\widetilde{\bm{x}}\in\widetilde{U}} \sum_{i=1}^n \alpha_i \widetilde{P}_i (\widetilde{X}=\widetilde{\bm{x}}) \\ &=  \argmax_{\bm{\alpha}\in\Delta^{n-1}}   \prod_{\widetilde{\bm{x}}\in\widetilde{U}} \sum_{i=1}^n \alpha_i \widetilde{Q}_i (\widetilde{X}=\widetilde{\bm{x}}) \\
    &=  \argmax_{\bm{\alpha}\in\Delta^{n-1}}   \prod_{\widetilde{\bm{x}}\in\widetilde{U}}  \mathbb{Q} (\widetilde{X}=\widetilde{\bm{x}}) ,\\
\end{split}
\end{align*}
where the last steps directly follow from the PPS assumptions. That is, minimizing the distribution matching in terms of the KLD is equivalent to maximizing the likelihood of the test data \citep{Garg:2020jt}.

This formulation points to the well-known method EMQ of \citet{Saerens:2002uq} as the most direct competitor, since EMQ is also a derivation of the maximum likelihood framework. The main difference between both methods resides in the way the optimization problem is approached: we resort to standard optimization routines to directly maximize the likelihood of a mixture of KDEs in the simplex, while EMQ reframes the problem in terms of the expectation maximization algorithm, by iteratively updating the posterior probabilities generated by a soft classifier (the E-step) and the prior estimates (the M-step) in a mutually recursive way, until convergence. In the experiments of Section~\ref{sec:experiments} we compare our method against EMQ and against the variant proposed by \citet{Alexandari:2020dn} that relies on a recalibration preprocessing.


We denote this variant KDEy-ML.

\section{Experiments}
\label{sec:experiments}

In this section, we turn to describe the experiments we have carried out for assessing the quantification performance of the KDEy variants we have proposed.

\subsection{Evaluation}
\label{sec:eval}

As the evaluation measures we adopt Absolute Error (AE) and Relative Absolute Error (RAE), two well-established evaluation measures in the quantification literature \citep[see][]{Sebastiani:2020qf}. AE and RAE are defined as
\begin{align*}
    \mathrm{AE}(\bm{\alpha},\hat{\bm{\alpha}}) &= \frac{1}{n}\sum_{i=1}^n |\alpha_i-\hat{\alpha}_i| , \\
    \mathrm{RAE}(\bm{\alpha},\hat{\bm{\alpha}}) &= \frac{1}{n}\sum_{i=1}^n\frac{|\alpha_i-\hat{\alpha}_i|}{\alpha_i+\epsilon} , 
\end{align*}
where $\bm{\alpha}$ is the true distribution and \new{$\hat{\bm{\alpha}}$} is an estimated distribution, and in which $\epsilon$ is a smoothing factor that we set, following \citet{Forman:2008kx}, to $\epsilon=0.5 z^{-1}$ with $z$ the test bag size.\footnote{Smoothing is required since, otherwise, RAE is undefined when the true prevalence of any class is zero. In the related literature RAE is more commonly presented as $\mathrm{RAE}(\bm{\alpha},\hat{\bm{\alpha}})=\frac{1}{n}\sum_{i=1}^n \frac{\bm{\nu}(\alpha_i)-\bm{\nu}(\hat{\alpha}_i)}{\bm{\nu}(\alpha_i)}$ with $\bm{\nu}(\alpha_i)=\frac{\epsilon+\alpha_i}{n\epsilon+1}$ the smoothing function. Both formulae are just equivalent.} 

Quantification methods are requested to be robust to prior probability shift. For this reason, quantifiers are typically evaluated in their ability to provide accurate class prevalence predictions in situations in which the class prevalence of the test \new{bags} is allowed to vary widely. In order to generate different such test \new{bags}, a \emph{sampling generation protocol} is employed. 

The most widely used such protocol, and the one we adopt here, is the so-called \emph{artificial-prevalence protocol} (APP). The goal of APP is to provide a uniform coverage of the space of plausible class distributions, i.e., of the $\Delta^{n-1}$ simplex. Early versions of this protocol due to \citet{Forman:2005fk} addressed the binary case only, in which the prevalence of the positive class was varied along a predefined grid of values (e.g., $[0, 0.01, 0.02, \ldots, 0.99, 1]$). For each prevalence value $g$ in the grid, a fixed number of realisations were drawn from a pool of test instances so that $100g\%$ of these are positive. Exploring such a grid in the multiclass case is cumbersome, and easily becomes infeasible even for relatively small number of classes. More recent applications of APP rather generate bags by iteratively sampling from $\Delta^{n-1}$ uniformly at random (we use the Kraemer sampling algorithm following \citet{Esuli:2022wf}), and then drawing examples from the pool according to the sampled distribution. 

We also use APP for generating validation bags for model selection, following \citet{Moreo:2021sp}. That is, in order to optimize the hyperparameters of a model, we explore a grid of combinations, that we test against validation bags characterized by different class prevalence values.

We use the implementation provided in \texttt{QuaPy} that guarantees the replicability of the \new{bags} generated (meaning that all methods are confronted with the exact same test \new{bags} during evaluation, and same validation \new{bags} during model selection).
The bag size and the number of bags we generate for each benchmark collection are reported in Table~\ref{tab:datasets}.

For each pair method-dataset, we report the mean AE and the mean RAE (MAE and MRAE, respectively), across all generated test bags. The statistical significance of the differences in performance is assessed using the Wilcoxon signed-rank statistical test at various confidence levels (0.05 and 0.005).

\subsection{Datasets}
\label{sec:datasets}

We use a total of 17 multiclass datasets in our experiments. These datasets can be naturally arranged in the following three groups whose characteristics are summarized in Table~\ref{tab:datasets}:
\begin{table}
    \centering
    \resizebox{\textwidth}{!}{%
    \begin{tabular}{ccccccccc}
        \toprule
         Group & \#Datasets & Training size & Validation pool size & Test pool size & \#Classes & \#Val. bags & \#Test bags & Bag size\\
         \midrule
         Tweets & 11& [797, 9684]& [263, 2000] (held-out) & [787, 3813] (held-out) & 3& 250& 1000& 100\\
         UCI multi & 5& [3096, 14000]& 40\% of training& [1328, 6000]& [3, 26]&   250& 1000& 500\\
         LeQua-T1B & 1& 20000& bags provided& bags provided& 28&  1000& 5000& 1000\\    
         \bottomrule
    \end{tabular}
    }%
    \caption{Characteristics of the datasets used in this work.}
    \label{tab:datasets}
\end{table}
\begin{itemize}
    \item Tweets: the 11 datasets used by \citet{Gao:2016uq} consisting of tweets labelled by sentiment in a three-class scheme (negative, neutral, positive). Each dataset comes with a training set (with sizes ranging in $[797, 9684]$) that we use for training our quantifiers, and a validation set (with sizes ranging in $[263,2000]$) that we use for model selection (more on this later). Model selection is carried out independently for each dataset by generating 250 bags (of 100 instances each) from the validation set using APP. Once the best hyperparameters have been chosen, we join the training and the validation sets and retrain the model anew. We then evaluate our optimized models on 1,000 bags (of 100 instances each) generated using APP on the test pool. 
    The 11 datasets are publicly available via Zenodo.\footnote{\url{https://zenodo.org/records/4255764}} These datasets are described in more detail in \citet{Gao:2016uq,Moreo:2022bf}.
    
    \item UCI multi: a collection of multiclass datasets from the UCI machine learning repository \citep{ucimlrepo}. This collection contains 5 datasets, which correspond to all the datasets that can be retrieved from the platform using the following filters: (i) datasets for classification, (ii) with more than 2 classes, (iii) containing at least 1,000 instances, and (iv) that can be imported using the Python API. In these datasets there are no predefined validation partitions, so we extract 40\% out of the training set, uniformly at random and with stratification, for model selection. We generate 250 validation bags of 500 instances each for model selection, after which we join the training set and \new{validation} set and retrain the selected model anew. For testing, we generate 1,000 bags of 500 instances each from the test pool using APP. The details of these datasets can be consulted in Table~\ref{tab:dataset:ucimulti}.

    \item LeQua-T1B: The multiclass problem proposed at the LeQua 2022 competition on quantification \citep{Esuli:2022wf}. This dataset comprises  Amazon product reviews labelled by merchandise category, for a total of 28 such categories. All the instances are already provided in numerical form as 300-dimensional dense vectors. The organizers made available 20,000 training instances (with individual class labels) plus 1,000 validation and 5,000 test bags (with per-bag prevalence labels); in all cases, the bags contain 1,000 instances each. The dataset is available via Zenodo.\footnote{\url{https://zenodo.org/records/6546188}} Check also the official web site of the competition\footnote{\url{https://lequa2022.github.io/}} and the overview paper of the competition \citep{Esuli:2022wf} for further details.
    
\end{itemize}
%


\begin{table}[h!]
    \centering
    \begin{tabular}{lrrr}
    \toprule
         \multicolumn{1}{l}{Dataset name} &  \multicolumn{1}{c}{\#training}&  \multicolumn{1}{c}{\#test} & \multicolumn{1}{c}{\#classes}\\
         \midrule
         dry-bean&  9527&  4084& 7\\
         wine-quality&  3428&  1470& 7\\
         academic-success&  3096&  1328& 3\\
         digits&  3933&  1687& 10\\
         letter&  14000&  6000& 26\\
         \bottomrule
    \end{tabular}    
    \caption{UCI machine learning repository multiclass datasets for classification used in this paper.}
    \label{tab:dataset:ucimulti}
\end{table}

\subsection{Baselines}
\label{sec:baselines}

We chose the following baseline systems in order to compare the performance of our KDE-based variants, that we organize in three categories: the adjustment methods, the distribution matching methods, and the maximum likelihood methods.


The \emph{adjustment methods} comprise methods that correct a preliminary estimate obtained by a classify-and-count method. These methods rely on the law of total probability, according to which
\begin{equation}
    \mathbb{Q}(\hat{Y}=i)=\sum_{i=1}^n \mathbb{Q}(\hat{Y}=i|Y=j) \mathbb{Q}(Y=j) ,
\end{equation}
where $\hat{Y}$ is the random variable describing the outcomes of a classifier. Note that $\mathbb{Q}(\hat{Y})$ is observed, \new{while the misclassification rates $\mathbb{Q}(\hat{Y}|Y)$ are unknown. However, under PPS conditions, the misclassification rates can be estimated in training via cross-validation} (see Section \ref{sec:notation}). The problem thus comes down to solving a system of $n$ linear equations with $n$ unknowns. We consider the following variants:

\begin{itemize}
    \item ACC: \emph{Adjusted Classify \& Count} \citep{Forman:2008kx}, also known as the \emph{Confusion Matrix Method} by \citet{Saerens:2002uq}, \nnew{and \emph{Black Box Shift Estimation hard}} \new{(BBSE-hard) by \citet{Lipton:2018fj}}. ACC uses a crisp classifier to compute $\mathbb{Q}(\hat{Y}=i)$ simply as the fraction of test instances labelled as belonging to class $i$, and estimates the misclassification rates $\mathbb{Q}(\hat{Y}=i|Y=j)$ of the crisp classifier via cross-validation in training. 

    \item PACC: \emph{Probabilistic Adjusted Classify \& Count} \citep{Bella:2010kx},  \nnew{also known as \emph{Black Box Shift Estimation soft} (BBSE-soft) by \citet{Lipton:2018fj}}, is the probabilistic counterpart of ACC, in which both $\mathbb{Q}(\hat{Y}=i)$ and $\mathbb{Q}(\hat{Y}=i|Y=j)$ are estimated using the expected counts returned by a soft classifier, instead of by a crisp classifier.
    
\end{itemize}

The \emph{distribution matching methods} account for some of the methods already discussed in Section~\ref{sec:related}. Specifically, we consider the following variants:

\begin{itemize}
    \item HDy-OvA: the method proposed by \citet{Gonzalez-Castro:2013fk}. Since this method is binary only, we apply the one-vs-all strategy followed by L1 normalization as described in Section~\ref{sec:related}.

    \item DM-T: a multiclass extension of the DM framework  \citep[as proposed by][]{Firat:2016uq,Bunse:2022zt} equipped with the Topsøe divergence as the dissimilarity measure. This method is added since the Topsøe divergence was found to be the best performance divergence in \citet{Maletzke:2019qd}.

    \item DM-HD: a multiclass extension of the DM framework \citep[as proposed by][]{Firat:2016uq,Bunse:2022zt} equipped with the $\operatorname{HD}^2$ divergence as the dissimilarity measure. This method is added to serve as a direct comparison against our KDEy-HD variant of Section~\ref{sec:sec:monte}.

    \item DM-CS: a multiclass extension of the DM framework \citep[as proposed by][]{Firat:2016uq,Bunse:2022zt} equipped with the CS divergence as the dissimilarity measure. This method is added to serve as a direct comparison against our KDEy-CS variant of Section~\ref{sec:sec:closed}.
    
\end{itemize}

The \emph{maximum likelihood methods} we use as baselines include:

\begin{itemize}
    \item EMQ: the expectation and maximization method proposed by \citet{Saerens:2002uq}, sometimes also called SLD (see, e.g., \citet{Book2023}) after the name of the inventors, \nnew{or Maximum Likelihood Label Shift (MLLS) in \citet{Lipton:2018fj}.} This method consists of an iterative, mutually recursive re-computation of the priors (the ``expectation step'') using the posteriors, and of the posteriors (the ``maximization step'') using the priors, until convergence.

    \item \nnew{EMQ-BCTS: a refinement of EMQ proposed by \citet{Alexandari:2020dn} in which the posterior probabilities are recalibrated by means of the Bias-Corrected Temperature Scaling (BCTS), the best-performing variant among the ones investigated by \citet{Alexandari:2020dn}, and in which the true training prevalence is replaced with an estimate of it. Our implementation of EMQ-BCTS relies on the implementation of the calibration methods made available by the authors.\footnote{\url{https://github.com/kundajelab/abstention}}
    }

    \item DIR: we propose a new maximum likelihood method that differs from our KDEy-ML of Section~\ref{sec:ml-solution} simply in the way the densities of the mixture components are modelled. In particular, we propose to model the probability density function $p_{\widetilde{L}_i}$ of each class $i$ by fitting a Dirichlet distribution on the simplex $\Delta^{n-1}$, using the set of (cross-validated) posterior probabilities of the training instances belonging to this class and then applying \eqref{eq:kdeml}. The only difference with respect to KDEy-ML thus lies in the fact that DIR fits (via maximum likelihood) a Dirichlet distribution, while we use KDE instead.\footnote{Note that KDE does not undergo a proper ``fit'' phase, since modelling a density function via KDE simply accounts for ``memorizing'' the set of reference points.}
    To fit the Dirichlet distribution, we rely on the software package \texttt{dirichlet}.\footnote{\url{https://github.com/ericsuh/dirichlet/tree/master}}
    
\end{itemize}

All the baseline systems, as well as the three methods we propose, rely on the outputs of a classifier. We adopt the L2-regularised logistic regression (LR), as implemented in \texttt{scikit-learn}\footnote{\url{https://scikit-learn.org/stable/modules/generated/sklearn.linear_model.LogisticRegression.html}} \citep{Pedregosa:2011yo}, in all cases since 
it is a probabilistic classifier that is known to deliver fairly well-calibrated posterior probabilities (unlike other classifiers like, e.g., SVMs), and since 
LR is the most widely used classifier in the quantification literature \citep{schumacher2023comparative,Moreo:2021bs,Moreo:2022bf}.

\textbf{Model Selection:}
We tune the following hyperparameters. 
For LR we optimize the hyperparameter $C$ in the range $\{10^{-3}, 10^{-2},\ldots, 10^{3}\}$, and the hyperparameter \textsc{class-weight} in ``balanced'' (reweights the importance of the instances so that the overall contribution of all classes is equated) or ``None'' (all instances have the same weight, so that more populated classes are likely to have a deeper impact in the decision function). Note that $C$ and \textsc{class-weight} are regarded as hyperparameters of the quantifier (to be optimized along with the rest of hyperparameters of the quantififer, if any), and not has hyperparameters of the classifier. This means that model selection is carried out according to a quantification-oriented loss (MAE, or MRAE) as explained in Section~\ref{sec:eval}, and not according to a classification-oriented loss. These are all the hyperparameters we optimize for methods ACC, PACC, and HDy-OvA,\footnote{Recall that HDy internally explores the number of bins from 10 to 110 at steps of 10 and returns the median of the predictions \citep{Gonzalez-Castro:2013fk}} EMQ, and DIR, since these quantifiers do only depend on the classifier's hyperparameters.

Concerning DM-T, DM-HD, and DM-CS, we additionally explore the number of bins in the range $b\in[2,3,\ldots,10,12,\ldots,32,64]$, while for our methods KDEy-HD, KDEy-CS, and KDEy-ML, we explore the bandwidth in the range $h\in[0.01, 0.02, \ldots, 0.2]$. Note that both the DM-based and the KDEy-based approaches are allowed to explore a fine-grained grid with \new{approximately the same number of points each (21 for DM, 20 for KDEy)}, so that the effort spent on optimizing all methods is comparable. Finally, note that we adopt the Gaussian kernel and do not further explore this choice as a hyperparameter of our KDE-based solutions, for two reasons: (i) it has been argued that the Gaussian is, in practice, a good default choice and that most kernels tend to behave fairly similarly \citep{chen2017tutorialKDE}, (ii) our KDEy-CS variant is tailored to GMMs, which requires KDE to operate with Gaussian kernels.

\textbf{Implementation details:} All methods here confronted are implemented as part of the \texttt{QuaPy} package \citep{Moreo:2021bs}. Some of them (specifically: ACC, PACC, HDy-OvA, DM-T, DM-HD, and EMQ) were already available in the package, while the rest (specifically: DM-CS, DIR, KDEy-HD, KDEy-CS, and KDEy-ML) have been implemented by us. Our implementation of the KDEy variants rests upon the \texttt{scikit-learn} implementation of KDE\new{ which in turn relies on KDTree to speed up the computation}.\footnote{\url{https://scikit-learn.org/stable/modules/generated/sklearn.neighbors.KernelDensity.html}} 
\new{For the Monte Carlo approximation (method KDEy-HD) we set the number of trials to $t=10,000$.}
The implementation of our methods, as well as all the necessary scripts to replicate all our experiments, is made available via the repository.\footnote{\url{https://github.com/HLT-ISTI/QuaPy/tree/kdey}}



\subsection{Results}
\label{sec:results}

Table~\ref{tab:multi_mae} reports the MAE results we have obtained for the baseline quantification methods of Section~\ref{sec:baselines} and for our proposed methods KDEy-HD, KDEy-CS, and KDEy-ML, with hyperparameters optimized for MAE, on the datasets of Section~\ref{sec:datasets}. Table~\ref{tab:multi_mrae} reports analogous results when the methods are optimized for, and evaluated in terms of MRAE. 
\new{In both tables, and for the sake of a clearer comparison, we omit the results for the methods ACC, HDy-OvA, DIR,} \nnew{and EMQ-BCTS;} \new{these methods performed comparably worse than the rest of the methods and some of them (HDy-OvA and DIR) did so by a large margin (the colour-coding we use ended up assigning an intense green color to any other method). The complete tables that include these omitted methods can be consulted in Appendix~\ref{app:multiclassfull}.}

\begin{table}[t!]
 \centering
\resizebox{\textwidth}{!}{%

            \begin{tabular}{|c|c|c|c|c|c|c|c|c|} \cline{2-9}           
            \multicolumn{1}{c}{} & 
            \multicolumn{1}{|c}{Adjustment} & 
            \multicolumn{5}{|c|}{Distribution Matching} & 
            \multicolumn{2}{c|}{Maximum Likelihood} \\
            \hline               
            \textbf{Tweets} & PACC & DM-T & DM-HD & KDEy-HD & DM-CS & KDEy-CS & EMQ & KDEy-ML \\\hline
gasp & .04270$^{\phantom{\ddag}}$ \cellcolor{red!6} & .04029$^{\phantom{\ddag}}$ \cellcolor{green!11} & .03976$^{\phantom{\ddag}}$ \cellcolor{green!15} & .03824$^{\phantom{\ddag}}$ \cellcolor{green!27} & .04639$^{\phantom{\ddag}}$ \cellcolor{red!35} & .04198$^{\phantom{\ddag}}$ \cellcolor{red!1} & .04474$^{\phantom{\ddag}}$ \cellcolor{red!22} & \textbf{.03721}$^{\phantom{\ddag}}$ \cellcolor{green!35} \\\cline{2-9}
hcr & .07136$^{\phantom{\ddag}}$ \cellcolor{green!25} & .08687$^{\phantom{\ddag}}$ \cellcolor{red!35} & .08313$^{\phantom{\ddag}}$ \cellcolor{red!20} & .07979$^{\phantom{\ddag}}$ \cellcolor{red!7} & .07600$^{\phantom{\ddag}}$ \cellcolor{green!7} & .07984$^{\phantom{\ddag}}$ \cellcolor{red!7} & .07871$^{\phantom{\ddag}}$ \cellcolor{red!3} & \textbf{.06878}$^{\phantom{\ddag}}$ \cellcolor{green!35} \\\cline{2-9}
omd & .06163$^{\ddag}$ \cellcolor{green!32} & .07749$^{\phantom{\ddag}}$ \cellcolor{red!24} & .06163$^{\ddag}$ \cellcolor{green!32} & .08041$^{\phantom{\ddag}}$ \cellcolor{red!35} & .06797$^{\phantom{\ddag}}$ \cellcolor{green!9} & .06646$^{\phantom{\ddag}}$ \cellcolor{green!14} & \textbf{.06085}$^{\phantom{\ddag}}$ \cellcolor{green!35} & .06099$^{\ddag}$ \cellcolor{green!34} \\\cline{2-9}
Sanders & .05104$^{\phantom{\ddag}}$ \cellcolor{red!35} & .04935$^{\phantom{\ddag}}$ \cellcolor{red!24} & .04955$^{\phantom{\ddag}}$ \cellcolor{red!25} & \textbf{.03945}$^{\phantom{\ddag}}$ \cellcolor{green!35} & .04417$^{\phantom{\ddag}}$ \cellcolor{green!6} & .04598$^{\phantom{\ddag}}$ \cellcolor{red!4} & .04476$^{\phantom{\ddag}}$ \cellcolor{green!2} & .03989$^{\ddag}$ \cellcolor{green!32} \\\cline{2-9}
SemEval13 & .07274$^{\phantom{\ddag}}$ \cellcolor{green!22} & .07065$^{\phantom{\ddag}}$ \cellcolor{green!27} & .07081$^{\phantom{\ddag}}$ \cellcolor{green!27} & .07161$^{\phantom{\ddag}}$ \cellcolor{green!25} & .07641$^{\phantom{\ddag}}$ \cellcolor{green!12} & .06804$^{\ddag}$ \cellcolor{green!34} & .09491$^{\phantom{\ddag}}$ \cellcolor{red!35} & \textbf{.06792}$^{\phantom{\ddag}}$ \cellcolor{green!35} \\\cline{2-9}
SemEval14 & .05813$^{\phantom{\ddag}}$ \cellcolor{green!23} & \textbf{.05486}$^{\phantom{\ddag}}$ \cellcolor{green!35} & .05792$^{\phantom{\ddag}}$ \cellcolor{green!24} & .05657$^{\phantom{\ddag}}$ \cellcolor{green!29} & .06078$^{\phantom{\ddag}}$ \cellcolor{green!14} & .06225$^{\phantom{\ddag}}$ \cellcolor{green!9} & .07518$^{\phantom{\ddag}}$ \cellcolor{red!35} & .06012$^{\phantom{\ddag}}$ \cellcolor{green!16} \\\cline{2-9}
SemEval15 & .09170$^{\phantom{\ddag}}$ \cellcolor{green!0} & .08988$^{\phantom{\ddag}}$ \cellcolor{green!10} & .09112$^{\phantom{\ddag}}$ \cellcolor{green!3} & .08803$^{\phantom{\ddag}}$ \cellcolor{green!19} & .09141$^{\phantom{\ddag}}$ \cellcolor{green!2} & .08616$^{\ddag}$ \cellcolor{green!29} & .09847$^{\phantom{\ddag}}$ \cellcolor{red!35} & \textbf{.08518}$^{\phantom{\ddag}}$ \cellcolor{green!35} \\\cline{2-9}
SemEval16 & .11263$^{\phantom{\ddag}}$ \cellcolor{green!18} & .13654$^{\phantom{\ddag}}$ \cellcolor{red!16} & .13610$^{\phantom{\ddag}}$ \cellcolor{red!15} & .12691$^{\phantom{\ddag}}$ \cellcolor{red!2} & .14938$^{\phantom{\ddag}}$ \cellcolor{red!35} & .13643$^{\phantom{\ddag}}$ \cellcolor{red!16} & \textbf{.10104}$^{\phantom{\ddag}}$ \cellcolor{green!35} & .12048$^{\phantom{\ddag}}$ \cellcolor{green!6} \\\cline{2-9}
sst & .05867$^{\phantom{\ddag}}$ \cellcolor{red!20} & .05626$^{\phantom{\ddag}}$ \cellcolor{green!6} & .05650$^{\phantom{\ddag}}$ \cellcolor{green!4} & .05533$^{\phantom{\ddag}}$ \cellcolor{green!17} & .05997$^{\phantom{\ddag}}$ \cellcolor{red!35} & .05945$^{\phantom{\ddag}}$ \cellcolor{red!29} & .05565$^{\dag}$ \cellcolor{green!13} & \textbf{.05376}$^{\phantom{\ddag}}$ \cellcolor{green!35} \\\cline{2-9}
wa & .04540$^{\phantom{\ddag}}$ \cellcolor{red!35} & .03754$^{\phantom{\ddag}}$ \cellcolor{green!27} & .03771$^{\phantom{\ddag}}$ \cellcolor{green!26} & .03734$^{\phantom{\ddag}}$ \cellcolor{green!29} & .04248$^{\phantom{\ddag}}$ \cellcolor{red!11} & .03877$^{\phantom{\ddag}}$ \cellcolor{green!17} & .03906$^{\phantom{\ddag}}$ \cellcolor{green!15} & \textbf{.03661}$^{\phantom{\ddag}}$ \cellcolor{green!35} \\\cline{2-9}
wb & .04612$^{\phantom{\ddag}}$ \cellcolor{red!35} & .03550$^{\phantom{\ddag}}$ \cellcolor{green!25} & .03547$^{\phantom{\ddag}}$ \cellcolor{green!25} & .03638$^{\phantom{\ddag}}$ \cellcolor{green!20} & .03875$^{\phantom{\ddag}}$ \cellcolor{green!6} & .03963$^{\phantom{\ddag}}$ \cellcolor{green!1} & \textbf{.03375}$^{\phantom{\ddag}}$ \cellcolor{green!35} & .03507$^{\phantom{\ddag}}$ \cellcolor{green!27} \\\cline{2-9}
\hline 
 \textit{Average} & .06474$^{\phantom{\ddag}}$ \cellcolor{red!1} & .06684$^{\phantom{\ddag}}$ \cellcolor{red!20} & .06543$^{\phantom{\ddag}}$ \cellcolor{red!7} & .06455$^{\phantom{\ddag}}$ \cellcolor{red!0} & .06852$^{\phantom{\ddag}}$ \cellcolor{red!35} & .06591$^{\phantom{\ddag}}$ \cellcolor{red!12} & .06610$^{\phantom{\ddag}}$ \cellcolor{red!13} & \textbf{.06055}$^{\phantom{\ddag}}$ \cellcolor{green!35} \\\cline{2-9}
\textit{Rank} &  5.5 \cellcolor{red!24} &  4.6 \cellcolor{red!10} &  4.5 \cellcolor{red!7} &  3.5 \cellcolor{green!7} &  6.2 \cellcolor{red!35} &  5.2 \cellcolor{red!18} &  4.6 \cellcolor{red!10} & \textbf{1.8} \cellcolor{green!35} \\\hline
\hline
\textbf{UCI-multi} & PACC & DM-T & DM-HD & KDEy-HD & DM-CS & KDEy-CS & EMQ & KDEy-ML \\\hline
dry-bean & .00496$^{\phantom{\ddag}}$ \cellcolor{red!2} & .00483$^{\phantom{\ddag}}$ \cellcolor{green!4} & .00494$^{\phantom{\ddag}}$ \cellcolor{red!1} & .00437$^{\phantom{\ddag}}$ \cellcolor{green!29} & .00500$^{\phantom{\ddag}}$ \cellcolor{red!4} & .00493$^{\phantom{\ddag}}$ \cellcolor{red!0} & .00556$^{\phantom{\ddag}}$ \cellcolor{red!35} & \textbf{.00427}$^{\phantom{\ddag}}$ \cellcolor{green!35} \\\cline{2-9}
wine-quality & .16971$^{\phantom{\ddag}}$ \cellcolor{red!35} & .14494$^{\phantom{\ddag}}$ \cellcolor{red!13} & .10197$^{\phantom{\ddag}}$ \cellcolor{green!24} & .11481$^{\phantom{\ddag}}$ \cellcolor{green!12} & \textbf{.08939}$^{\phantom{\ddag}}$ \cellcolor{green!35} & .15445$^{\phantom{\ddag}}$ \cellcolor{red!21} & .10982$^{\phantom{\ddag}}$ \cellcolor{green!17} & .09836$^{\phantom{\ddag}}$ \cellcolor{green!27} \\\cline{2-9}
academic-success & .02864$^{\phantom{\ddag}}$ \cellcolor{green!11} & .02295$^{\phantom{\ddag}}$ \cellcolor{green!30} & \textbf{.02177}$^{\phantom{\ddag}}$ \cellcolor{green!35} & .02513$^{\phantom{\ddag}}$ \cellcolor{green!23} & .02822$^{\phantom{\ddag}}$ \cellcolor{green!12} & .03545$^{\phantom{\ddag}}$ \cellcolor{red!11} & .04225$^{\phantom{\ddag}}$ \cellcolor{red!35} & .02402$^{\phantom{\ddag}}$ \cellcolor{green!27} \\\cline{2-9}
digits & .00372$^{\phantom{\ddag}}$ \cellcolor{red!35} & .00296$^{\phantom{\ddag}}$ \cellcolor{green!5} & .00351$^{\phantom{\ddag}}$ \cellcolor{red!24} & .00264$^{\phantom{\ddag}}$ \cellcolor{green!22} & .00315$^{\phantom{\ddag}}$ \cellcolor{red!4} & .00297$^{\phantom{\ddag}}$ \cellcolor{green!5} & .00249$^{\phantom{\ddag}}$ \cellcolor{green!30} & \textbf{.00241}$^{\phantom{\ddag}}$ \cellcolor{green!35} \\\cline{2-9}
letter & .00520$^{\phantom{\ddag}}$ \cellcolor{red!6} & .00508$^{\phantom{\ddag}}$ \cellcolor{red!4} & .00543$^{\phantom{\ddag}}$ \cellcolor{red!9} & .00447$^{\phantom{\ddag}}$ \cellcolor{green!4} & .00714$^{\phantom{\ddag}}$ \cellcolor{red!35} & .00481$^{\phantom{\ddag}}$ \cellcolor{red!0} & .00521$^{\phantom{\ddag}}$ \cellcolor{red!6} & \textbf{.00243}$^{\phantom{\ddag}}$ \cellcolor{green!35} \\\cline{2-9}
\hline 
 \textit{Average} & .04245$^{\phantom{\ddag}}$ \cellcolor{red!35} & .03615$^{\phantom{\ddag}}$ \cellcolor{red!7} & .02753$^{\phantom{\ddag}}$ \cellcolor{green!29} & .03029$^{\phantom{\ddag}}$ \cellcolor{green!17} & .02658$^{\phantom{\ddag}}$ \cellcolor{green!33} & .04052$^{\phantom{\ddag}}$ \cellcolor{red!26} & .03307$^{\phantom{\ddag}}$ \cellcolor{green!5} & \textbf{.02630}$^{\phantom{\ddag}}$ \cellcolor{green!35} \\\cline{2-9}
\textit{Rank} &  6.6 \cellcolor{red!35} &  3.8 \cellcolor{green!4} &  4.6 \cellcolor{red!6} &  3.2 \cellcolor{green!12} &  5.4 \cellcolor{red!18} &  5.2 \cellcolor{red!15} &  5.6 \cellcolor{red!20} & \textbf{1.6} \cellcolor{green!35} \\\hline
\hline
\textbf{LeQua} & PACC & DM-T & DM-HD & KDEy-HD & DM-CS & KDEy-CS & EMQ & KDEy-ML \\\hline
T1B & .01277$^{\phantom{\ddag}}$ \cellcolor{red!23} & .01070$^{\phantom{\ddag}}$ \cellcolor{green!18} & .01112$^{\phantom{\ddag}}$ \cellcolor{green!10} & \textbf{.00990}$^{\phantom{\ddag}}$ \cellcolor{green!35} & .01336$^{\phantom{\ddag}}$ \cellcolor{red!35} & .00996$^{\phantom{\ddag}}$ \cellcolor{green!33} & .01177$^{\phantom{\ddag}}$ \cellcolor{red!2} & .01153$^{\phantom{\ddag}}$ \cellcolor{green!2} \\\cline{2-9}

\textit{Rank} &  7 \cellcolor{red!24} &  3 \cellcolor{green!15} &  4 \cellcolor{green!4} & \textbf{1} \cellcolor{green!35} &  8 \cellcolor{red!35} &  2 \cellcolor{green!25} &  6 \cellcolor{red!15} &  5 \cellcolor{red!4} \\\hline

\end{tabular}
}%
 \caption{Values of MAE obtained in our experiments for different multiclass quantification methods optimized for MAE on different groups of datasets. \textbf{Boldface} indicates the best method for the given dataset. Superscripts $\dag$ and $\ddag$ denote the methods (if any) whose scores are not statistically significantly different from the best one according to a Wilcoxon signed-rank test at different confidence levels: symbol $\dag$ indicates $0.005 < p\mathrm{-value}<0.05$ while symbol $\ddag$ indicates $0.05\leq p\mathrm{-value}$. Cells are colour-coded so as to facilitate readability and allow for quick comparisons across the results in a row, in which we highlight the best result in intense green, the worst result in intense red, and where all other in-between scores are linearly interpolated between these tones. For each group of datasets, we report the total average and the average rank position.}
 \label{tab:multi_mae}
\end{table}

\begin{table}[t!]
 \centering
\resizebox{\textwidth}{!}{%

            \begin{tabular}{|c|c|c|c|c|c|c|c|c|} \cline{2-9}           
            \multicolumn{1}{c}{} & 
            \multicolumn{1}{|c}{Adjustment} & 
            \multicolumn{5}{|c|}{Distribution Matching} & 
            \multicolumn{2}{c|}{Maximum Likelihood} \\
            \hline               
            \textbf{Tweets} & PACC & DM-T & DM-HD & KDEy-HD & DM-CS & KDEy-CS & EMQ & KDEy-ML \\\hline
gasp &  0.29326$^{\phantom{\ddag}}$ \cellcolor{green!9} &  0.32874$^{\phantom{\ddag}}$ \cellcolor{red!14} &  0.28929$^{\phantom{\ddag}}$ \cellcolor{green!11} &  0.26124$^{\phantom{\ddag}}$ \cellcolor{green!30} &  0.35572$^{\phantom{\ddag}}$ \cellcolor{red!32} &  0.27852$^{\ddag}$ \cellcolor{green!19} &  0.35914$^{\phantom{\ddag}}$ \cellcolor{red!35} & \textbf{0.25496}$^{\phantom{\ddag}}$ \cellcolor{green!35} \\\cline{2-9}
hcr &  0.79477$^{\phantom{\ddag}}$ \cellcolor{red!35} &  0.59522$^{\phantom{\ddag}}$ \cellcolor{green!4} &  0.63863$^{\phantom{\ddag}}$ \cellcolor{red!3} &  0.60654$^{\phantom{\ddag}}$ \cellcolor{green!2} &  0.56566$^{\phantom{\ddag}}$ \cellcolor{green!10} &  0.55952$^{\phantom{\ddag}}$ \cellcolor{green!12} & \textbf{0.44521}$^{\phantom{\ddag}}$ \cellcolor{green!35} &  0.46854$^{\ddag}$ \cellcolor{green!30} \\\cline{2-9}
omd &  0.42726$^{\phantom{\ddag}}$ \cellcolor{green!10} &  0.61895$^{\phantom{\ddag}}$ \cellcolor{red!35} &  0.37808$^{\phantom{\ddag}}$ \cellcolor{green!22} &  0.45936$^{\phantom{\ddag}}$ \cellcolor{green!2} &  0.48768$^{\phantom{\ddag}}$ \cellcolor{red!3} &  0.41857$^{\phantom{\ddag}}$ \cellcolor{green!12} &  0.39500$^{\phantom{\ddag}}$ \cellcolor{green!18} & \textbf{0.32427}$^{\phantom{\ddag}}$ \cellcolor{green!35} \\\cline{2-9}
Sanders &  0.36327$^{\phantom{\ddag}}$ \cellcolor{red!35} &  0.30559$^{\phantom{\ddag}}$ \cellcolor{red!4} &  0.26275$^{\phantom{\ddag}}$ \cellcolor{green!18} & \textbf{0.23231}$^{\phantom{\ddag}}$ \cellcolor{green!35} &  0.26648$^{\phantom{\ddag}}$ \cellcolor{green!16} &  0.26750$^{\phantom{\ddag}}$ \cellcolor{green!16} &  0.28101$^{\phantom{\ddag}}$ \cellcolor{green!8} &  0.26037$^{\phantom{\ddag}}$ \cellcolor{green!19} \\\cline{2-9}
SemEval13 &  0.50860$^{\phantom{\ddag}}$ \cellcolor{green!7} &  0.43602$^{\phantom{\ddag}}$ \cellcolor{green!28} &  0.44950$^{\phantom{\ddag}}$ \cellcolor{green!24} &  0.44898$^{\phantom{\ddag}}$ \cellcolor{green!24} &  0.49009$^{\phantom{\ddag}}$ \cellcolor{green!13} &  0.42761$^{\ddag}$ \cellcolor{green!30} &  0.66054$^{\phantom{\ddag}}$ \cellcolor{red!35} & \textbf{0.41223}$^{\phantom{\ddag}}$ \cellcolor{green!35} \\\cline{2-9}
SemEval14 &  0.45413$^{\dag}$ \cellcolor{green!11} &  0.41251$^{\phantom{\ddag}}$ \cellcolor{green!28} &  0.40196$^{\phantom{\ddag}}$ \cellcolor{green!32} &  0.40232$^{\phantom{\ddag}}$ \cellcolor{green!32} &  0.47810$^{\phantom{\ddag}}$ \cellcolor{green!2} &  0.40633$^{\phantom{\ddag}}$ \cellcolor{green!30} &  0.57333$^{\phantom{\ddag}}$ \cellcolor{red!35} & \textbf{0.39554}$^{\phantom{\ddag}}$ \cellcolor{green!35} \\\cline{2-9}
SemEval15 &  0.66874$^{\phantom{\ddag}}$ \cellcolor{green!0} &  0.60081$^{\phantom{\ddag}}$ \cellcolor{green!27} &  0.60188$^{\phantom{\ddag}}$ \cellcolor{green!27} & \textbf{0.58260}$^{\phantom{\ddag}}$ \cellcolor{green!35} &  0.63606$^{\phantom{\ddag}}$ \cellcolor{green!13} &  0.60223$^{\ddag}$ \cellcolor{green!27} &  0.75774$^{\phantom{\ddag}}$ \cellcolor{red!35} &  0.58579$^{\phantom{\ddag}}$ \cellcolor{green!33} \\\cline{2-9}
SemEval16 &  0.96384$^{\ddag}$ \cellcolor{red!31} &  0.83537$^{\phantom{\ddag}}$ \cellcolor{red!1} &  0.81140$^{\phantom{\ddag}}$ \cellcolor{green!3} & \textbf{0.67679}$^{\phantom{\ddag}}$ \cellcolor{green!35} &  0.84354$^{\phantom{\ddag}}$ \cellcolor{red!3} &  0.97161$^{\phantom{\ddag}}$ \cellcolor{red!33} &  0.98026$^{\ddag}$ \cellcolor{red!35} &  0.76681$^{\phantom{\ddag}}$ \cellcolor{green!14} \\\cline{2-9}
sst &  0.47170$^{\phantom{\ddag}}$ \cellcolor{red!35} &  0.37498$^{\phantom{\ddag}}$ \cellcolor{green!17} &  0.35942$^{\phantom{\ddag}}$ \cellcolor{green!25} & \textbf{0.34271}$^{\phantom{\ddag}}$ \cellcolor{green!35} &  0.42177$^{\phantom{\ddag}}$ \cellcolor{red!7} &  0.46344$^{\phantom{\ddag}}$ \cellcolor{red!30} &  0.44465$^{\ddag}$ \cellcolor{red!20} &  0.35142$^{\dag}$ \cellcolor{green!30} \\\cline{2-9}
wa &  0.29951$^{\phantom{\ddag}}$ \cellcolor{red!33} & \textbf{0.21424}$^{\phantom{\ddag}}$ \cellcolor{green!35} &  0.22128$^{\phantom{\ddag}}$ \cellcolor{green!29} &  0.21789$^{\ddag}$ \cellcolor{green!32} &  0.30173$^{\phantom{\ddag}}$ \cellcolor{red!35} &  0.23331$^{\ddag}$ \cellcolor{green!19} &  0.22379$^{\ddag}$ \cellcolor{green!27} &  0.23197$^{\ddag}$ \cellcolor{green!20} \\\cline{2-9}
wb &  0.31116$^{\phantom{\ddag}}$ \cellcolor{red!35} &  0.24166$^{\phantom{\ddag}}$ \cellcolor{green!13} &  0.23300$^{\phantom{\ddag}}$ \cellcolor{green!19} &  0.22224$^{\phantom{\ddag}}$ \cellcolor{green!26} &  0.27188$^{\phantom{\ddag}}$ \cellcolor{red!7} &  0.25751$^{\phantom{\ddag}}$ \cellcolor{green!2} & \textbf{0.20992}$^{\phantom{\ddag}}$ \cellcolor{green!35} &  0.22016$^{\phantom{\ddag}}$ \cellcolor{green!27} \\\cline{2-9}
\hline 
 \textit{Average} &  0.50511$^{\phantom{\ddag}}$ \cellcolor{red!35} &  0.45128$^{\phantom{\ddag}}$ \cellcolor{red!2} &  0.42247$^{\phantom{\ddag}}$ \cellcolor{green!14} &  0.40482$^{\phantom{\ddag}}$ \cellcolor{green!25} &  0.46534$^{\phantom{\ddag}}$ \cellcolor{red!11} &  0.44420$^{\phantom{\ddag}}$ \cellcolor{green!1} &  0.48460$^{\phantom{\ddag}}$ \cellcolor{red!22} & \textbf{0.38837}$^{\phantom{\ddag}}$ \cellcolor{green!35} \\\cline{2-9}
\textit{Rank} &  6.8 \cellcolor{red!35} &  4.6 \cellcolor{red!3} &  3.6 \cellcolor{green!10} &  2.7 \cellcolor{green!23} &  6.0 \cellcolor{red!23} &  4.7 \cellcolor{red!5} &  5.5 \cellcolor{red!16} & \textbf{1.9} \cellcolor{green!35} \\\hline
\hline
\textbf{UCI-multi} & PACC & DM-T & DM-HD & KDEy-HD & DM-CS & KDEy-CS & EMQ & KDEy-ML \\\hline
dry-bean &  0.09932$^{\phantom{\ddag}}$ \cellcolor{red!22} &  0.08905$^{\phantom{\ddag}}$ \cellcolor{green!0} &  0.08681$^{\phantom{\ddag}}$ \cellcolor{green!4} &  0.07687$^{\ddag}$ \cellcolor{green!26} &  0.10495$^{\phantom{\ddag}}$ \cellcolor{red!35} &  0.08853$^{\phantom{\ddag}}$ \cellcolor{green!1} &  0.08582$^{\phantom{\ddag}}$ \cellcolor{green!7} & \textbf{0.07321}$^{\phantom{\ddag}}$ \cellcolor{green!35} \\\cline{2-9}
wine-quality &  4.11163$^{\phantom{\ddag}}$ \cellcolor{red!35} &  2.34737$^{\ddag}$ \cellcolor{green!27} &  2.15553$^{\ddag}$ \cellcolor{green!34} &  2.23004$^{\dag}$ \cellcolor{green!32} &  2.99908$^{\phantom{\ddag}}$ \cellcolor{green!4} &  2.52138$^{\phantom{\ddag}}$ \cellcolor{green!21} &  2.56550$^{\ddag}$ \cellcolor{green!20} & \textbf{2.14771}$^{\phantom{\ddag}}$ \cellcolor{green!35} \\\cline{2-9}
academic-success &  0.31747$^{\phantom{\ddag}}$ \cellcolor{red!35} & \textbf{0.16107}$^{\phantom{\ddag}}$ \cellcolor{green!35} &  0.17045$^{\ddag}$ \cellcolor{green!30} &  0.21692$^{\phantom{\ddag}}$ \cellcolor{green!10} &  0.22369$^{\phantom{\ddag}}$ \cellcolor{green!6} &  0.28252$^{\phantom{\ddag}}$ \cellcolor{red!19} &  0.27068$^{\phantom{\ddag}}$ \cellcolor{red!14} &  0.20673$^{\phantom{\ddag}}$ \cellcolor{green!14} \\\cline{2-9}
digits &  0.08512$^{\phantom{\ddag}}$ \cellcolor{red!11} &  0.05744$^{\phantom{\ddag}}$ \cellcolor{green!25} &  0.08200$^{\phantom{\ddag}}$ \cellcolor{red!7} &  0.05270$^{\phantom{\ddag}}$ \cellcolor{green!31} &  0.10291$^{\phantom{\ddag}}$ \cellcolor{red!35} &  0.05563$^{\phantom{\ddag}}$ \cellcolor{green!27} &  0.05121$^{\ddag}$ \cellcolor{green!33} & \textbf{0.04997}$^{\phantom{\ddag}}$ \cellcolor{green!35} \\\cline{2-9}
letter &  0.31580$^{\phantom{\ddag}}$ \cellcolor{green!6} &  0.25673$^{\phantom{\ddag}}$ \cellcolor{green!16} &  0.27577$^{\phantom{\ddag}}$ \cellcolor{green!13} &  0.21494$^{\phantom{\ddag}}$ \cellcolor{green!22} &  0.58296$^{\phantom{\ddag}}$ \cellcolor{red!35} &  0.24267$^{\phantom{\ddag}}$ \cellcolor{green!18} &  0.26120$^{\phantom{\ddag}}$ \cellcolor{green!15} & \textbf{0.13769}$^{\phantom{\ddag}}$ \cellcolor{green!35} \\\cline{2-9}
\hline 
 \textit{Average} &  0.98587$^{\phantom{\ddag}}$ \cellcolor{red!35} &  0.58233$^{\phantom{\ddag}}$ \cellcolor{green!26} &  0.55411$^{\phantom{\ddag}}$ \cellcolor{green!30} &  0.55829$^{\phantom{\ddag}}$ \cellcolor{green!29} &  0.80272$^{\phantom{\ddag}}$ \cellcolor{red!7} &  0.63815$^{\phantom{\ddag}}$ \cellcolor{green!17} &  0.64688$^{\phantom{\ddag}}$ \cellcolor{green!16} & \textbf{0.52306}$^{\phantom{\ddag}}$ \cellcolor{green!35} \\\cline{2-9}
\textit{Rank} &  7.4 \cellcolor{red!35} &  4.0 \cellcolor{green!4} &  4.0 \cellcolor{green!4} &  2.8 \cellcolor{green!18} &  7.2 \cellcolor{red!32} &  4.8 \cellcolor{red!4} &  4.4 \cellcolor{red!0} & \textbf{1.4} \cellcolor{green!35} \\\hline
\hline
\textbf{LeQua} & PACC & DM-T & DM-HD & KDEy-HD & DM-CS & KDEy-CS & EMQ & KDEy-ML \\\hline
T1B &  1.37638$^{\phantom{\ddag}}$ \cellcolor{red!8} &  0.92422$^{\phantom{\ddag}}$ \cellcolor{green!24} &  0.91116$^{\phantom{\ddag}}$ \cellcolor{green!25} & \textbf{0.78339}$^{\phantom{\ddag}}$ \cellcolor{green!35} &  1.73593$^{\phantom{\ddag}}$ \cellcolor{red!35} &  0.84367$^{\phantom{\ddag}}$ \cellcolor{green!30} &  0.87802$^{\phantom{\ddag}}$ \cellcolor{green!28} &  0.82795$^{\phantom{\ddag}}$ \cellcolor{green!31} \\\cline{2-9}

\textit{Rank} &  7 \cellcolor{red!24} &  6 \cellcolor{red!15} &  5 \cellcolor{red!4} & \textbf{1} \cellcolor{green!35} &  8 \cellcolor{red!35} &  3 \cellcolor{green!15} &  4 \cellcolor{green!4} &  2 \cellcolor{green!25} \\\hline

\end{tabular}
}%
\caption{Values of MRAE obtained in our experiments for different multiclass quantification methods optimized for MRAE on different groups of datasets. Notational conventions are as in Table \ref{tab:multi_mae}.}
 \label{tab:multi_mrae}
\end{table}

Overall, our results show KDEy-ML and KDEy-HD are the strongest quantification methods of the lot. In particular, KDEy-ML obtains the best results in the ``Tweets'' group, both in terms of MAE (6 best results out of 11, \new{plus 2 cases in which the result is not statistically significantly different from the best one with high confidence}) and MRAE (4 best results out of 11, \new{plus 3 cases in which the result is not statistically significantly different from the best one with high confidence}). Also concerning the ``UCI-multi'' group \new{KDEy-ML clearly stands out in terms of MAE (3 best results out of 5) as well as in terms of MRAE (4 best results out of 5)}. 
Finally, concerning the LeQua-T1B dataset, KDEy-HD obtained the best results, both in terms of MAE \new{(closely followed by KDEy-CS)} as well as in terms of MRAE \new{(followed by KDEy-ML)}. 

The LeQua-T1B deserves further comments since, as recalled from Section~\ref{sec:datasets}, this dataset was proposed as part of a public competition. For this dataset, we observed that our variants surpassed the best results obtained by all other participant teams in the competition (0.01173 in MAE, and 0.087987 in MRAE---the official evaluation measure for the competition). We analyzed the hyperparameters chosen via model selection, and found some of them where obtained for values of the bandwidth in the boundaries of the grid explored (i.e., for $h=0.2$); see also the following Section~\ref{sec:stability}. In these particular cases, we further explored the bandwidth up to value 0.3; the results reported in the tables already account for this extended configuration \new{(when we stick to values of $h\in[0.01,0.20]$, the variant KDEy-HD already obtained results that were superior to all other systems---in particular, MAE=0.00993 and MRAE=0.80552)}. We did similar checks for the hyperparameter $b$ \new{(the number of bins)} of the DM baselines but, surprisingly enough, in all cases the value found optimal was either $b=2$ or $b=3$.
It is also worth mentioning that the baselines DM-T and DM-HD obtained better MAE scores than the best scores obtained in the competition by any of the participants; however, this might be explained by the fact that MRAE was the official evaluation measure of the competition, and therefore the methods submitted were optimized for MRAE, and not for MAE \new{(recall that the results displayed in Table~\ref{tab:multi_mae} are from methods optimized for MAE, while the results displayed in Table~\ref{tab:multi_mrae} are from methods optimized for MRAE)}. Finally, it is also noteworthy the fact that EMQ obtained much better results in our experiments than those reported in the competition (the method EMQ was named SLD in the overview paper of the competition \citep{Esuli:2022wf}). The likely reason is that our implementation does not attempt to re-calibrate the posterior probabilities of the LR classifier \new{as suggested by \citet{Alexandari:2020dn}}, while the variant used in the competition instead did. \new{In particular, the organizers of LeQua used Platt's scaling for calibrating the outputs of LR. 
The variant EMQ-BCTS that we tested did not show any improvement with respect to ``vanilla'' EMQ (no calibration).}
Apparently, the re-calibration phase harmed the quality of the posterior probabilities. \new{A likely reason is that LR is already assumed to be fairly well-calibrated.}

Note that, in the above-mentioned Tables~\ref{tab:multi_mae}~and~\ref{tab:multi_mrae} we have placed each of our variants side to side with its most direct competitor, so as to allow for a direct comparison of the two. Something that jumps to the eye, is that, in the great majority of cases, KDEy-HD outperforms DM-HD, while KDEy-CS outperforms DM-CS. This is particularly relevant since these two pairs of variants rely on the same dissimilarity measure (HD and CS) and differ only in the way the bags are represented in the model. \new{This analysis thus effectively isolates the contribution of the representation mechanism;}
the results  
seem to suggest that our KDE-based representation is indeed better suited for multiclass quantification problems than the concatenation of histograms used by the DM variants. This observation seems to confirm our initial hypothesis, according to which bringing to bear the inter-class interactions into the model should be beneficial for multiclass quantification. \new{Similarly, the variant KDEy-ML proves superior to EMQ in most of the cases. This is important, since EMQ has long been considered a ``hard to beat'' method in the quantification literature \citep{Moreo:2022bf,schumacher2023comparative} and the label shift literature \citep{Alexandari:2020dn}. We have verified that the differences in averaged performance across all tests carried out, are statistically significant with $p$-value $\ll 0.005$ for the pairs DM-HD vs. KDEy-HD (both in terms of MAE and MRAE), DM-CS vs. KDEy-CS (only for MRAE), and EMQ vs. KDEy-ML (both in terms of MAE and MRAE).}

A secondary observation that emerges from our experiments is that the One-vs-All approach gives rise to very weak multiclass quantification systems \new{(the results were indeed omitted from Tables~\ref{tab:multi_mae} and \ref{tab:multi_mrae} since they were not comparable, but can be consulted in Appendix~\ref{app:multiclassfull})}. This \new{observation is} actually in line with the observations echoed by \citet{donyavi2023mc}. 

Something that might come as a surprise, though, is the fact that our DM-HD variant obtained better results than the DM-T, since the Topsøe has been considered superior to the HD after the work \citet{Maletzke:2019qd}. One possible explanation for this misalignment may have to do with the fact that the experiments of \citet{Maletzke:2019qd} involved only the binary case. 

Lastly, the DIR baseline we propose does not seem to be competitive in terms of performance with respect to the stronger baselines \new{(the results can be consulted in Appendix~\ref{app:multiclassfull})}. The only difference between DIR and our KDEy-ML variant resides in the way the probability density functions are modelled, which indirectly speaks in favour of modelling class-conditional densities as GMMs obtained via KDE.






\subsection{Stability Analysis}
\label{sec:stability}

In this section, we turn to analyze the sensitivity of KDEy towards the bandwidth hyperparameter. In particular, we aim to investigate the method's sensitivity to this hyperparameter and assess its stability, i.e., if small variations of the bandwidth can lead to abrupt variations in performance. For this analysis, we chose KDEy-ML, the best among the KDE variants, and do not optimize the hyperparameters of the classifier LR (which we simple left at their defaults values).

\begin{figure}
    \centering
    \includegraphics[width=\textwidth]{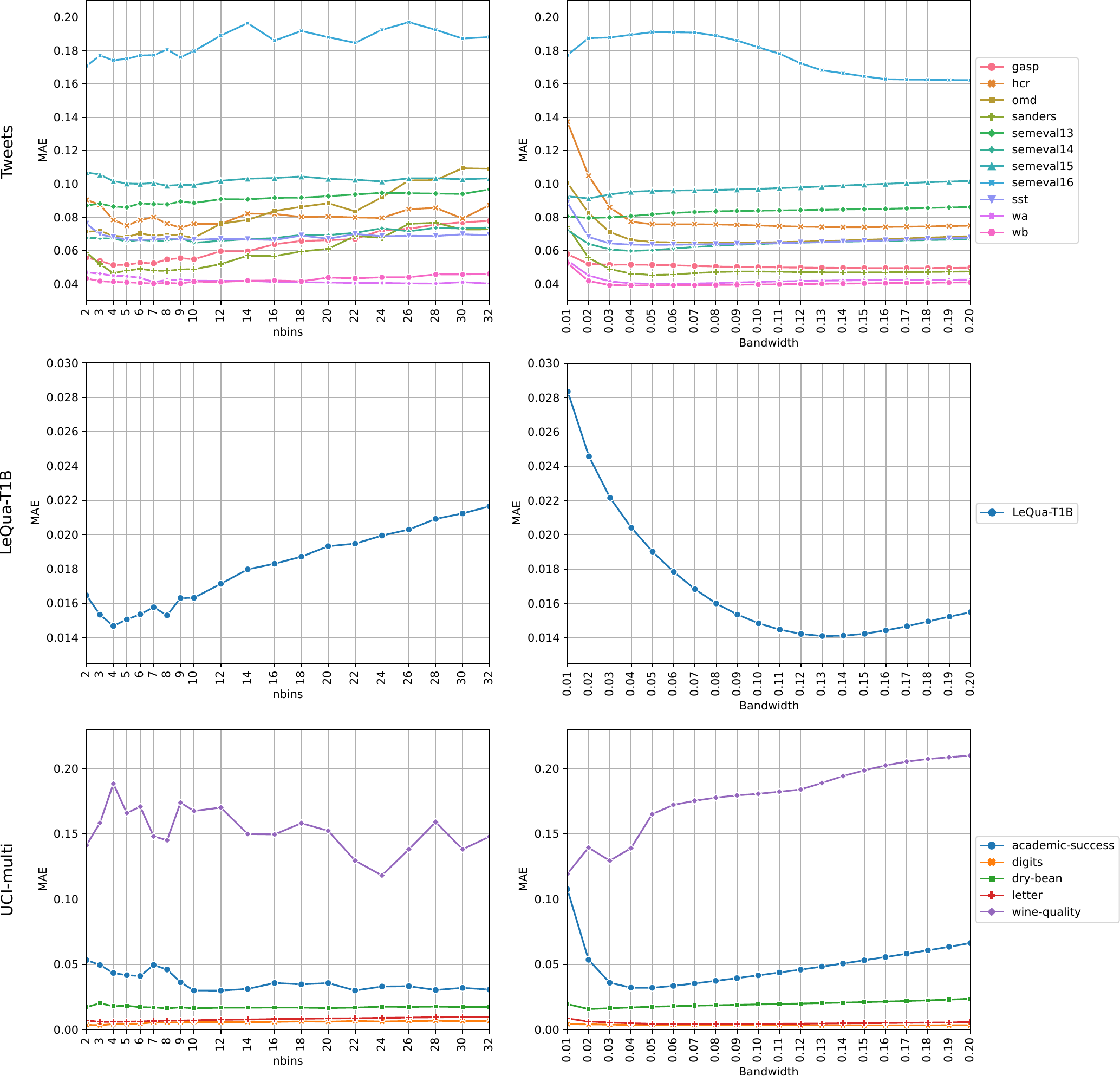}    
    \caption{Sensitivity of DM-HD (left) and KDEy-ML (right) to the hyperparameters, number of bins  (``nbins'') and bandwidth, respectively.}
    \label{fig:sensitivity}
\end{figure}

Figure~\ref{fig:sensitivity} displays the sensitivity of KDEy-ML in terms of MAE due to variations of the bandwidth. This figure also reports results for DM-HD, the best among the DM-based variants, at variations in ``nbins'' (i.e., $b$, the equivalent counter-part of the bandwidth).

These results seem to confirm that KDEy-ML is stable with respect to the bandwidth (i.e., small variations in the bandwidth result in small variations in performance). This is a desirable property of the model which may encourage more sophisticated ways for optimizing this hyperparameter, beyond simple grid-search exploration. We leave these considerations for future work.

Conversely, DM-HD seems to behave more unstably with respect to the number of bins. This is witnessed by the fact that small increments in the number of bins sometimes result in abrupt fluctuations in performance; see, e.g., the fluctuations experienced in datasets ``semeval16'', ``hcr'', and ``omd'' (group ``Tweets'') or in datasets ``wine-quality'' and ``academic-success'' (group ``UCI-multi''). 

As a final note, some heuristics have been explored in the literature to decide for the bandwidth of KDE; examples borrowed from \citet{chen2017tutorialKDE} include the ``rule of thumb'' of Silverman, the cross-validation method by Scott, or the plug-in method of Woodroofe. We have investigated these heuristics during our preliminary experiments with no success. For this reason, we preferred simply treating the bandwidth as any other hyperparameter of the quantifier, and submit it to a  quantification-oriented model selection procedure.

\subsection{KDEy for Binary Quantification}
\label{sec:binary}

Even though KDEy was conceived with multiclass problems in mind, nothing prevents its application to binary problems. In this section, we try to answer the question: How does KDEy compare against other quantification methods in the binary setup?

To this aim, we use 29 UCI binary datasets\footnote{We decided to discard ``balance.2'' because we have observed that all methods (with no exception) produce errors which are orders of magnitude higher than in the rest of the datasets, and this has disproportionate effect in the average. We noted these results are actually aligned with the results reported by \citet{Perez-Gallego:2017wt,Moreo:2021bs}, so we suspect there is something wrong with that dataset. This phenomenon did not harm our methods more than the rest of baselines.} used in previous quantification papers \citep{Perez-Gallego:2017wt,Moreo:2021bs} \new{plus LeQua-T1A, the binary task presented at the LeQua competition (see \citet{Esuli:2022wf} for further details).
Tables~\ref{tab:bin_mae} and \ref{tab:bin_mrae} report the results we have obtained in terms of MAE and MRAE, for methods optimized, respectively, for MAE or MRAE. For the sake of conciseness, only the averaged values are presented for the ``UCI-binary'' group. The complete results for the 29 datasets can be consulted in the Appendix \ref{app:binresults}.
}

\begin{table}[t]
 \centering
\resizebox{\textwidth}{!}{%

            \begin{tabular}{|c|c|c|c|c|c|c|c|c|c|c|c|c|} \cline{2-13}           
            \multicolumn{1}{c}{} & 
            \multicolumn{2}{|c}{Adjustment} & 
            \multicolumn{6}{|c|}{Distribution Matching} & 
            \multicolumn{4}{c|}{Maximum Likelihood} \\
            \hline               
            \textbf{UCI-binary} & ACC & PACC & HDy & DM-T & DM-HD & KDEy-HD & DM-CS & KDEy-CS & DIR & EMQ & EMQ-BCTS & KDEy-ML \\\hline
\hline 
 \textit{Average} & .06963$^{\phantom{\ddag}}$ \cellcolor{red!11} & .05679$^{\phantom{\ddag}}$ \cellcolor{green!10} & .08336$^{\phantom{\ddag}}$ \cellcolor{red!35} & .05434$^{\phantom{\ddag}}$ \cellcolor{green!14} & .05647$^{\phantom{\ddag}}$ \cellcolor{green!11} & \textbf{.04249}$^{\phantom{\ddag}}$ \cellcolor{green!35} & .04906$^{\phantom{\ddag}}$ \cellcolor{green!23} & .04956$^{\phantom{\ddag}}$ \cellcolor{green!22} & .06192$^{\phantom{\ddag}}$ \cellcolor{green!1} & .05122$^{\phantom{\ddag}}$ \cellcolor{green!20} & .07339$^{\phantom{\ddag}}$ \cellcolor{red!17} & .04529$^{\phantom{\ddag}}$ \cellcolor{green!30} \\\cline{2-13}

\textit{Rank} &  9.4 \cellcolor{red!35} &  7.6 \cellcolor{red!9} &  8.9 \cellcolor{red!28} &  5.7 \cellcolor{green!17} &  5.7 \cellcolor{green!17} & \textbf{4.4} \cellcolor{green!35} &  6.5 \cellcolor{green!5} &  6.2 \cellcolor{green!9} &  6.8 \cellcolor{green!1} &  4.8 \cellcolor{green!29} &  7.3 \cellcolor{red!5} &  4.9 \cellcolor{green!27} \\\hline
\hline
\textbf{LeQua} & ACC & PACC & HDy & DM-T & DM-HD & KDEy-HD & DM-CS & KDEy-CS & DIR & EMQ & EMQ-BCTS & KDEy-ML \\\hline
T1A & .03133$^{\phantom{\ddag}}$ \cellcolor{red!35} & .02642$^{\phantom{\ddag}}$ \cellcolor{green!7} & .02660$^{\phantom{\ddag}}$ \cellcolor{green!6} & .02364$^{\dag}$ \cellcolor{green!31} & .02371$^{\dag}$ \cellcolor{green!31} & .02350$^{\ddag}$ \cellcolor{green!33} & .02502$^{\phantom{\ddag}}$ \cellcolor{green!19} & .02473$^{\phantom{\ddag}}$ \cellcolor{green!22} & .02463$^{\phantom{\ddag}}$ \cellcolor{green!23} & .02359$^{\dag}$ \cellcolor{green!32} & \textbf{.02327}$^{\phantom{\ddag}}$ \cellcolor{green!35} & .02417$^{\phantom{\ddag}}$ \cellcolor{green!27} \\\cline{2-13}

\textit{Rank} &  12 \cellcolor{red!35} &  10 \cellcolor{red!22} &  11 \cellcolor{red!28} &  4 \cellcolor{green!15} &  5 \cellcolor{green!9} &  2 \cellcolor{green!28} &  9 \cellcolor{red!15} &  8 \cellcolor{red!9} &  7 \cellcolor{red!3} &  3 \cellcolor{green!22} & \textbf{1} \cellcolor{green!35} &  6 \cellcolor{green!3} \\\hline

\end{tabular}
}%
\caption{\new{Values of MAE obtained in our binary experiments, for methods optimized for MAE.}}
 \label{tab:bin_mae}
\end{table}

\begin{table}[t]
 \centering
\resizebox{\textwidth}{!}{%

            \begin{tabular}{|c|c|c|c|c|c|c|c|c|c|c|c|c|} \cline{2-13}           
            \multicolumn{1}{c}{} & 
            \multicolumn{2}{|c}{Adjustment} & 
            \multicolumn{6}{|c|}{Distribution Matching} & 
            \multicolumn{4}{c|}{Maximum Likelihood} \\
            \hline               
            \textbf{UCI-binary} & ACC & PACC & HDy & DM-T & DM-HD & KDEy-HD & DM-CS & KDEy-CS & DIR & EMQ & EMQ-BCTS & KDEy-ML \\\hline
\hline 
 \textit{Average} &  0.34051$^{\phantom{\ddag}}$ \cellcolor{red!14} &  0.26198$^{\phantom{\ddag}}$ \cellcolor{green!8} &  0.40669$^{\phantom{\ddag}}$ \cellcolor{red!35} &  0.23992$^{\phantom{\ddag}}$ \cellcolor{green!15} &  0.26182$^{\phantom{\ddag}}$ \cellcolor{green!8} & \textbf{0.17587}$^{\phantom{\ddag}}$ \cellcolor{green!35} &  0.25255$^{\phantom{\ddag}}$ \cellcolor{green!11} &  0.23828$^{\phantom{\ddag}}$ \cellcolor{green!16} &  0.33095$^{\phantom{\ddag}}$ \cellcolor{red!12} &  0.19214$^{\ddag}$ \cellcolor{green!30} &  0.29014$^{\phantom{\ddag}}$ \cellcolor{green!0} &  0.18541$^{\dag}$ \cellcolor{green!32} \\\cline{2-13}

\textit{Rank} &  9.8 \cellcolor{red!35} &  8.5 \cellcolor{red!20} &  8.8 \cellcolor{red!23} &  5.2 \cellcolor{green!19} &  6.0 \cellcolor{green!9} &  4.0 \cellcolor{green!33} &  6.9 \cellcolor{red!0} &  5.7 \cellcolor{green!13} &  8.1 \cellcolor{red!15} & \textbf{3.9} \cellcolor{green!35} &  6.6 \cellcolor{green!2} &  4.5 \cellcolor{green!28} \\\hline
\hline
\textbf{LeQua} & ACC & PACC & HDy & DM-T & DM-HD & KDEy-HD & DM-CS & KDEy-CS & DIR & EMQ & EMQ-BCTS & KDEy-ML \\\hline
T1A &  0.18025$^{\phantom{\ddag}}$ \cellcolor{red!35} &  0.16476$^{\phantom{\ddag}}$ \cellcolor{red!19} &  0.15654$^{\phantom{\ddag}}$ \cellcolor{red!11} &  0.12019$^{\phantom{\ddag}}$ \cellcolor{green!23} &  0.11711$^{\phantom{\ddag}}$ \cellcolor{green!26} &  0.11766$^{\phantom{\ddag}}$ \cellcolor{green!26} &  0.14913$^{\phantom{\ddag}}$ \cellcolor{red!4} &  0.14167$^{\phantom{\ddag}}$ \cellcolor{green!2} &  0.13777$^{\phantom{\ddag}}$ \cellcolor{green!6} & \textbf{0.10878}$^{\phantom{\ddag}}$ \cellcolor{green!35} &  0.11305$^{\phantom{\ddag}}$ \cellcolor{green!30} &  0.11641$^{\phantom{\ddag}}$ \cellcolor{green!27} \\\cline{2-13}

\textit{Rank} &  12 \cellcolor{red!35} &  11 \cellcolor{red!28} &  10 \cellcolor{red!22} &  6 \cellcolor{green!3} &  4 \cellcolor{green!15} &  5 \cellcolor{green!9} &  9 \cellcolor{red!15} &  8 \cellcolor{red!9} &  7 \cellcolor{red!3} & \textbf{1} \cellcolor{green!35} &  2 \cellcolor{green!28} &  3 \cellcolor{green!22} \\\hline

\end{tabular}
}%
 \caption{\new{Values of MRAE obtained in our binary experiments, for methods optimized for MRAE.}}
 \label{tab:bin_mrae}
\end{table}

\new{
In contrast to previous experiments reported in Section~\ref{sec:results}, these results do not hint at a clear winning method.
While KDEy-HD seems to stand out in terms of MAE in the ``UCI-binary'' group, the average ranking does not deviate much from the second-best (EMQ) nor from the third-best (KDEy-ML). \nnew{The best MAE result for the Lequa-T1A is attained by EMQ-BCTS, but the difference in average performance with respect to KDEy-HD, EMQ, DM-T, and DM-HD is not statistically significant.} In terms of MRAE we observe similar trends in the ``UCI-binary'' group, i.e., that KDEy-HD obtains the best MAE results but that \nnew{other methods (EMQ, KDEy-ML) perform on par, from a statistically significant point of view.}
For LeQua-T1A, though, EMQ seems to be the best performer in terms of MRAE. 

A second, probably more interesting observation that emerges from Tables~\ref{tab:bin_mae} and \ref{tab:bin_mrae} is that the pairwise comparisons do no longer show a clear trend, i.e., that there are no clear winners within the pairs of methods (DM-HD, KDEy-HD), (DM-CS, KDEy-CS), (EMQ, KDEy-ML). All in all, these outcomes were to be expected, since the advantage of bringing to bear the inter-class correlations (the main motivation that has driven this research) might be witnessed, if at all, in problems with more than 2 classes. The scope of this set of experiments was instead to show that KDEy is a general method that is not limited to the multiclass case, i.e., that KDEy can safely be applied in binary problems too. 
The results we have obtained testify that this is indeed the case.}




\section{Conclusions}
\label{sec:conclusions}

Many disciplines exist for which the interest lies in knowing the distribution of the classes in unlabelled data samples, and in which we are not interested in individual label predictions. A myriad of quantification methods have been proposed so far, among which, distribution matching methods represent a fairly important family of approaches. While the distribution matching methods are natively binary, some extensions have been proposed to the multiclass case in the literature. In this article, we argue that such extensions are suboptimal, since they fail to capture the possible inter-class interactions that might exist in the data. 

We have investigated possible ways for bringing these class-class correlations into the model; to this aim, we propose to switch the representation mechanism from independent class-wise histograms to multivariate GMMs obtained through KDE. We have presented different instances of our KDE-based solution, depending on whether the solution is framed as a distribution matching setting, or whether it is framed under the maximum likelihood framework. 
%
In all cases, our proposed methods performed very well, beating the previously proposed histogram-based models, and  also setting a new state-of-the-art result in the multiclass task T1B of the LeQua competition. The method has demonstrated competitive performance also in binary problems, hence proving itself a versatile approach for quantification problems.

The methods we present introduce a new hyperparameter: the bandwidth of the kernel. 
The experiments we have carried out indicate that our method behaves stably with respect to the hyperparameter (small variations produce small effects in performance). This seems to suggest more sophisticated alternatives might be explored that aim at finding the optimal value avoiding a brute-force optimization. In future work, we plan to pursue this idea. 

\section*{Acknowledgments}
The work of the first author has been supported by the SoBigData.it (grant IR0000013), FAIR (grant PE00000013), and QuaDaSh (grant P2022TB5JF) projects funded by the Italian MUR (Ministry of University and Research) under the European Commission ``Next Generation EU'' program.
The work by the second, and third authors has been funded by MINECO (the Spanish Ministerio de Econom\'ia y Competitividad) and FEDER (Fondo Europeo de Desarrollo Regional), grant PID2019-110742RB-I00 (MINECO/FEDER). 

Additionally, we are thankful to Mirko Bunse for insightful discussions about current frameworks for multiclass distribution matching methods held during the preliminary stages of this work, which greatly contributed to enriching our understanding of the topic.

\appendix
\clearpage
\newpage
\section{Appendix}
\label{app:multiclassfull}

\new{In Section~\ref{sec:results} we omitted the results for the methods ACC, HDy-OvA, and DIR, since these methods performed significantly worse than the rest of the methods, thus blurring their relative merits and hindering the visual comparison among them. In Tables~\ref{tab:multi_mae_full} and \ref{tab:multi_mrae_full} we report the full set of results, including those for ACC, HDy-OvA, and DIR.}

\begin{table}[h]
 \centering
\resizebox{\textwidth}{!}{%

            \begin{tabular}{|c|c|c|c|c|c|c|c|c|c|c|c|c|} \cline{2-13}           
            \multicolumn{1}{c}{} & 
            \multicolumn{2}{|c}{Adjustment} & 
            \multicolumn{6}{|c|}{Distribution Matching} & 
            \multicolumn{4}{c|}{Maximum Likelihood} \\
            \hline               
            \textbf{Tweets} & ACC & PACC & HDy-OvA & DM-T & DM-HD & KDEy-HD & DM-CS & KDEy-CS & DIR & EMQ & EMQ-BCTS & KDEy-ML \\\hline
gasp & .05751$^{\phantom{\ddag}}$ \cellcolor{green!16} & .04270$^{\phantom{\ddag}}$ \cellcolor{green!30} & .07744$^{\phantom{\ddag}}$ \cellcolor{red!1} & .04029$^{\phantom{\ddag}}$ \cellcolor{green!32} & .03976$^{\phantom{\ddag}}$ \cellcolor{green!32} & .03824$^{\phantom{\ddag}}$ \cellcolor{green!34} & .04639$^{\phantom{\ddag}}$ \cellcolor{green!26} & .04198$^{\phantom{\ddag}}$ \cellcolor{green!30} & .05397$^{\phantom{\ddag}}$ \cellcolor{green!19} & .04474$^{\phantom{\ddag}}$ \cellcolor{green!28} & .11451$^{\phantom{\ddag}}$ \cellcolor{red!35} & \textbf{.03721}$^{\phantom{\ddag}}$ \cellcolor{green!35} \\\cline{2-13}
hcr & .10375$^{\phantom{\ddag}}$ \cellcolor{green!3} & .07136$^{\phantom{\ddag}}$ \cellcolor{green!32} & .14581$^{\phantom{\ddag}}$ \cellcolor{red!35} & .08687$^{\phantom{\ddag}}$ \cellcolor{green!18} & .08313$^{\phantom{\ddag}}$ \cellcolor{green!21} & .07979$^{\phantom{\ddag}}$ \cellcolor{green!24} & .07600$^{\phantom{\ddag}}$ \cellcolor{green!28} & .07984$^{\phantom{\ddag}}$ \cellcolor{green!24} & .07163$^{\ddag}$ \cellcolor{green!32} & .07871$^{\phantom{\ddag}}$ \cellcolor{green!25} & .12531$^{\phantom{\ddag}}$ \cellcolor{red!16} & \textbf{.06878}$^{\phantom{\ddag}}$ \cellcolor{green!35} \\\cline{2-13}
omd & .07695$^{\phantom{\ddag}}$ \cellcolor{green!12} & .06163$^{\ddag}$ \cellcolor{green!33} & .10363$^{\phantom{\ddag}}$ \cellcolor{red!23} & .07749$^{\phantom{\ddag}}$ \cellcolor{green!12} & .06163$^{\ddag}$ \cellcolor{green!33} & .08041$^{\phantom{\ddag}}$ \cellcolor{green!8} & .06797$^{\phantom{\ddag}}$ \cellcolor{green!25} & .06646$^{\phantom{\ddag}}$ \cellcolor{green!27} & .06264$^{\ddag}$ \cellcolor{green!32} & \textbf{.06085}$^{\phantom{\ddag}}$ \cellcolor{green!35} & .11185$^{\phantom{\ddag}}$ \cellcolor{red!35} & .06099$^{\ddag}$ \cellcolor{green!34} \\\cline{2-13}
Sanders & .07331$^{\phantom{\ddag}}$ \cellcolor{green!11} & .05104$^{\phantom{\ddag}}$ \cellcolor{green!27} & .14099$^{\phantom{\ddag}}$ \cellcolor{red!35} & .04935$^{\phantom{\ddag}}$ \cellcolor{green!28} & .04955$^{\phantom{\ddag}}$ \cellcolor{green!28} & \textbf{.03945}$^{\phantom{\ddag}}$ \cellcolor{green!35} & .04417$^{\phantom{\ddag}}$ \cellcolor{green!31} & .04598$^{\phantom{\ddag}}$ \cellcolor{green!30} & .08510$^{\phantom{\ddag}}$ \cellcolor{green!3} & .04476$^{\phantom{\ddag}}$ \cellcolor{green!31} & .12739$^{\phantom{\ddag}}$ \cellcolor{red!25} & .03989$^{\ddag}$ \cellcolor{green!34} \\\cline{2-13}
SemEval13 & .08186$^{\phantom{\ddag}}$ \cellcolor{green!9} & .07274$^{\phantom{\ddag}}$ \cellcolor{green!20} & .12226$^{\phantom{\ddag}}$ \cellcolor{red!35} & .07065$^{\phantom{\ddag}}$ \cellcolor{green!22} & .07081$^{\phantom{\ddag}}$ \cellcolor{green!22} & .07161$^{\phantom{\ddag}}$ \cellcolor{green!21} & .07641$^{\phantom{\ddag}}$ \cellcolor{green!16} & .06804$^{\phantom{\ddag}}$ \cellcolor{green!25} & .10478$^{\phantom{\ddag}}$ \cellcolor{red!15} & .09491$^{\phantom{\ddag}}$ \cellcolor{red!4} & \textbf{.05935}$^{\phantom{\ddag}}$ \cellcolor{green!35} & .06792$^{\phantom{\ddag}}$ \cellcolor{green!25} \\\cline{2-13}
SemEval14 & .06604$^{\phantom{\ddag}}$ \cellcolor{green!6} & .05813$^{\phantom{\ddag}}$ \cellcolor{green!18} & .09392$^{\phantom{\ddag}}$ \cellcolor{red!35} & .05486$^{\phantom{\ddag}}$ \cellcolor{green!23} & .05792$^{\phantom{\ddag}}$ \cellcolor{green!19} & .05657$^{\phantom{\ddag}}$ \cellcolor{green!21} & .06078$^{\phantom{\ddag}}$ \cellcolor{green!14} & .06225$^{\phantom{\ddag}}$ \cellcolor{green!12} & .08250$^{\phantom{\ddag}}$ \cellcolor{red!17} & .07518$^{\phantom{\ddag}}$ \cellcolor{red!6} & \textbf{.04726}$^{\phantom{\ddag}}$ \cellcolor{green!35} & .06012$^{\phantom{\ddag}}$ \cellcolor{green!15} \\\cline{2-13}
SemEval15 & .10465$^{\phantom{\ddag}}$ \cellcolor{red!15} & .09170$^{\phantom{\ddag}}$ \cellcolor{green!18} & .11219$^{\phantom{\ddag}}$ \cellcolor{red!35} & .08988$^{\phantom{\ddag}}$ \cellcolor{green!22} & .09112$^{\phantom{\ddag}}$ \cellcolor{green!19} & .08803$^{\phantom{\ddag}}$ \cellcolor{green!27} & .09141$^{\phantom{\ddag}}$ \cellcolor{green!18} & .08616$^{\ddag}$ \cellcolor{green!32} & .10998$^{\phantom{\ddag}}$ \cellcolor{red!29} & .09847$^{\phantom{\ddag}}$ \cellcolor{green!0} & .08585$^{\ddag}$ \cellcolor{green!33} & \textbf{.08518}$^{\phantom{\ddag}}$ \cellcolor{green!35} \\\cline{2-13}
SemEval16 & .13801$^{\phantom{\ddag}}$ \cellcolor{red!4} & .11263$^{\phantom{\ddag}}$ \cellcolor{green!22} & .16739$^{\phantom{\ddag}}$ \cellcolor{red!35} & .13654$^{\phantom{\ddag}}$ \cellcolor{red!2} & .13610$^{\phantom{\ddag}}$ \cellcolor{red!1} & .12691$^{\phantom{\ddag}}$ \cellcolor{green!7} & .14938$^{\phantom{\ddag}}$ \cellcolor{red!16} & .13643$^{\phantom{\ddag}}$ \cellcolor{red!2} & .12109$^{\phantom{\ddag}}$ \cellcolor{green!13} & \textbf{.10104}$^{\phantom{\ddag}}$ \cellcolor{green!35} & .10331$^{\ddag}$ \cellcolor{green!32} & .12048$^{\phantom{\ddag}}$ \cellcolor{green!14} \\\cline{2-13}
sst & .06982$^{\phantom{\ddag}}$ \cellcolor{green!17} & .05867$^{\phantom{\ddag}}$ \cellcolor{green!29} & .07657$^{\phantom{\ddag}}$ \cellcolor{green!9} & .05626$^{\phantom{\ddag}}$ \cellcolor{green!32} & .05650$^{\phantom{\ddag}}$ \cellcolor{green!31} & .05533$^{\phantom{\ddag}}$ \cellcolor{green!33} & .05997$^{\phantom{\ddag}}$ \cellcolor{green!28} & .05945$^{\phantom{\ddag}}$ \cellcolor{green!28} & .06415$^{\phantom{\ddag}}$ \cellcolor{green!23} & .05565$^{\dag}$ \cellcolor{green!32} & .11651$^{\phantom{\ddag}}$ \cellcolor{red!35} & \textbf{.05376}$^{\phantom{\ddag}}$ \cellcolor{green!35} \\\cline{2-13}
wa & .05206$^{\phantom{\ddag}}$ \cellcolor{red!35} & .04540$^{\phantom{\ddag}}$ \cellcolor{red!4} & .04657$^{\phantom{\ddag}}$ \cellcolor{red!10} & .03754$^{\phantom{\ddag}}$ \cellcolor{green!30} & .03771$^{\phantom{\ddag}}$ \cellcolor{green!30} & .03734$^{\phantom{\ddag}}$ \cellcolor{green!31} & .04248$^{\phantom{\ddag}}$ \cellcolor{green!8} & .03877$^{\phantom{\ddag}}$ \cellcolor{green!25} & .04324$^{\phantom{\ddag}}$ \cellcolor{green!4} & .03906$^{\phantom{\ddag}}$ \cellcolor{green!23} & .04683$^{\phantom{\ddag}}$ \cellcolor{red!11} & \textbf{.03661}$^{\phantom{\ddag}}$ \cellcolor{green!35} \\\cline{2-13}
wb & .04822$^{\phantom{\ddag}}$ \cellcolor{red!6} & .04612$^{\phantom{\ddag}}$ \cellcolor{red!1} & .05996$^{\phantom{\ddag}}$ \cellcolor{red!35} & .03550$^{\phantom{\ddag}}$ \cellcolor{green!23} & .03547$^{\phantom{\ddag}}$ \cellcolor{green!23} & .03638$^{\phantom{\ddag}}$ \cellcolor{green!21} & .03875$^{\phantom{\ddag}}$ \cellcolor{green!15} & .03963$^{\phantom{\ddag}}$ \cellcolor{green!13} & .04733$^{\phantom{\ddag}}$ \cellcolor{red!4} & .03375$^{\phantom{\ddag}}$ \cellcolor{green!27} & \textbf{.03068}$^{\phantom{\ddag}}$ \cellcolor{green!35} & .03507$^{\phantom{\ddag}}$ \cellcolor{green!24} \\\cline{2-13}
\hline 
 \textit{Average} & .07929$^{\phantom{\ddag}}$ \cellcolor{green!4} & .06474$^{\phantom{\ddag}}$ \cellcolor{green!28} & .10425$^{\phantom{\ddag}}$ \cellcolor{red!35} & .06684$^{\phantom{\ddag}}$ \cellcolor{green!24} & .06543$^{\phantom{\ddag}}$ \cellcolor{green!27} & .06455$^{\phantom{\ddag}}$ \cellcolor{green!28} & .06852$^{\phantom{\ddag}}$ \cellcolor{green!22} & .06591$^{\phantom{\ddag}}$ \cellcolor{green!26} & .07694$^{\phantom{\ddag}}$ \cellcolor{green!8} & .06610$^{\phantom{\ddag}}$ \cellcolor{green!26} & .08808$^{\phantom{\ddag}}$ \cellcolor{red!9} & \textbf{.06055}$^{\phantom{\ddag}}$ \cellcolor{green!35} \\\cline{2-13}
\textit{Rank} &  9.8 \cellcolor{red!22} &  6.1 \cellcolor{green!5} &  11.5 \cellcolor{red!35} &  5.5 \cellcolor{green!10} &  5.1 \cellcolor{green!13} &  4.4 \cellcolor{green!18} &  7.0 \cellcolor{red!1} &  5.9 \cellcolor{green!7} &  8.4 \cellcolor{red!11} &  5.3 \cellcolor{green!11} &  6.9 \cellcolor{red!0} & \textbf{2.2} \cellcolor{green!35} \\\hline
\hline
\textbf{UCI-multi} & ACC & PACC & HDy-OvA & DM-T & DM-HD & KDEy-HD & DM-CS & KDEy-CS & DIR & EMQ & EMQ-BCTS & KDEy-ML \\\hline
dry-bean & .00547$^{\phantom{\ddag}}$ \cellcolor{green!32} & .00496$^{\phantom{\ddag}}$ \cellcolor{green!33} & .03405$^{\phantom{\ddag}}$ \cellcolor{red!26} & .00483$^{\phantom{\ddag}}$ \cellcolor{green!33} & .00494$^{\phantom{\ddag}}$ \cellcolor{green!33} & .00437$^{\phantom{\ddag}}$ \cellcolor{green!34} & .00500$^{\phantom{\ddag}}$ \cellcolor{green!33} & .00493$^{\phantom{\ddag}}$ \cellcolor{green!33} & .03838$^{\phantom{\ddag}}$ \cellcolor{red!35} & .00556$^{\phantom{\ddag}}$ \cellcolor{green!32} & .00686$^{\phantom{\ddag}}$ \cellcolor{green!29} & \textbf{.00427}$^{\phantom{\ddag}}$ \cellcolor{green!35} \\\cline{2-13}
wine-quality & .17611$^{\phantom{\ddag}}$ \cellcolor{red!15} & .16971$^{\phantom{\ddag}}$ \cellcolor{red!11} & .13497$^{\phantom{\ddag}}$ \cellcolor{green!8} & .14494$^{\phantom{\ddag}}$ \cellcolor{green!2} & .10197$^{\phantom{\ddag}}$ \cellcolor{green!27} & .11481$^{\phantom{\ddag}}$ \cellcolor{green!20} & \textbf{.08939}$^{\phantom{\ddag}}$ \cellcolor{green!35} & .15445$^{\phantom{\ddag}}$ \cellcolor{red!2} & .20935$^{\phantom{\ddag}}$ \cellcolor{red!35} & .10982$^{\phantom{\ddag}}$ \cellcolor{green!23} & .14947$^{\phantom{\ddag}}$ \cellcolor{red!0} & .09836$^{\phantom{\ddag}}$ \cellcolor{green!29} \\\cline{2-13}
academic-success & .03584$^{\phantom{\ddag}}$ \cellcolor{green!20} & .02864$^{\phantom{\ddag}}$ \cellcolor{green!27} & .05448$^{\phantom{\ddag}}$ \cellcolor{green!0} & .02295$^{\phantom{\ddag}}$ \cellcolor{green!33} & \textbf{.02177}$^{\phantom{\ddag}}$ \cellcolor{green!35} & .02513$^{\phantom{\ddag}}$ \cellcolor{green!31} & .02822$^{\phantom{\ddag}}$ \cellcolor{green!28} & .03545$^{\phantom{\ddag}}$ \cellcolor{green!20} & .05892$^{\phantom{\ddag}}$ \cellcolor{red!3} & .04225$^{\phantom{\ddag}}$ \cellcolor{green!13} & .08876$^{\phantom{\ddag}}$ \cellcolor{red!35} & .02402$^{\phantom{\ddag}}$ \cellcolor{green!32} \\\cline{2-13}
digits & .00423$^{\phantom{\ddag}}$ \cellcolor{red!3} & .00372$^{\phantom{\ddag}}$ \cellcolor{green!7} & .00572$^{\phantom{\ddag}}$ \cellcolor{red!35} & .00296$^{\phantom{\ddag}}$ \cellcolor{green!23} & .00351$^{\phantom{\ddag}}$ \cellcolor{green!11} & .00264$^{\phantom{\ddag}}$ \cellcolor{green!30} & .00315$^{\phantom{\ddag}}$ \cellcolor{green!19} & .00297$^{\phantom{\ddag}}$ \cellcolor{green!23} & .00392$^{\phantom{\ddag}}$ \cellcolor{green!3} & .00249$^{\phantom{\ddag}}$ \cellcolor{green!33} & .00265$^{\phantom{\ddag}}$ \cellcolor{green!30} & \textbf{.00241}$^{\phantom{\ddag}}$ \cellcolor{green!35} \\\cline{2-13}
letter & .00618$^{\phantom{\ddag}}$ \cellcolor{green!18} & .00520$^{\phantom{\ddag}}$ \cellcolor{green!22} & .01810$^{\phantom{\ddag}}$ \cellcolor{red!35} & .00508$^{\phantom{\ddag}}$ \cellcolor{green!23} & .00543$^{\phantom{\ddag}}$ \cellcolor{green!21} & .00447$^{\phantom{\ddag}}$ \cellcolor{green!25} & .00714$^{\phantom{\ddag}}$ \cellcolor{green!13} & .00481$^{\phantom{\ddag}}$ \cellcolor{green!24} & .00934$^{\phantom{\ddag}}$ \cellcolor{green!4} & .00521$^{\phantom{\ddag}}$ \cellcolor{green!22} & .00562$^{\phantom{\ddag}}$ \cellcolor{green!20} & \textbf{.00243}$^{\phantom{\ddag}}$ \cellcolor{green!35} \\\cline{2-13}
\hline 
 \textit{Average} & .04557$^{\phantom{\ddag}}$ \cellcolor{red!0} & .04245$^{\phantom{\ddag}}$ \cellcolor{green!5} & .04946$^{\phantom{\ddag}}$ \cellcolor{red!8} & .03615$^{\phantom{\ddag}}$ \cellcolor{green!16} & .02753$^{\phantom{\ddag}}$ \cellcolor{green!32} & .03029$^{\phantom{\ddag}}$ \cellcolor{green!27} & .02658$^{\phantom{\ddag}}$ \cellcolor{green!34} & .04052$^{\phantom{\ddag}}$ \cellcolor{green!8} & .06398$^{\phantom{\ddag}}$ \cellcolor{red!35} & .03307$^{\phantom{\ddag}}$ \cellcolor{green!22} & .05067$^{\phantom{\ddag}}$ \cellcolor{red!10} & \textbf{.02630}$^{\phantom{\ddag}}$ \cellcolor{green!35} \\\cline{2-13}
\textit{Rank} &  9.4 \cellcolor{red!21} &  7.2 \cellcolor{red!5} &  10.2 \cellcolor{red!27} &  4.2 \cellcolor{green!16} &  4.8 \cellcolor{green!11} &  3.2 \cellcolor{green!23} &  6.0 \cellcolor{green!2} &  5.8 \cellcolor{green!4} &  11.2 \cellcolor{red!35} &  6.0 \cellcolor{green!2} &  8.4 \cellcolor{red!14} & \textbf{1.6} \cellcolor{green!35} \\\hline
\hline
\textbf{LeQua} & ACC & PACC & HDy-OvA & DM-T & DM-HD & KDEy-HD & DM-CS & KDEy-CS & DIR & EMQ & EMQ-BCTS & KDEy-ML \\\hline
T1B & .02081$^{\phantom{\ddag}}$ \cellcolor{red!21} & .01277$^{\phantom{\ddag}}$ \cellcolor{green!19} & .02331$^{\phantom{\ddag}}$ \cellcolor{red!35} & .01070$^{\phantom{\ddag}}$ \cellcolor{green!30} & .01112$^{\phantom{\ddag}}$ \cellcolor{green!28} & \textbf{.00990}$^{\phantom{\ddag}}$ \cellcolor{green!35} & .01336$^{\phantom{\ddag}}$ \cellcolor{green!16} & .00996$^{\phantom{\ddag}}$ \cellcolor{green!34} & .01546$^{\phantom{\ddag}}$ \cellcolor{green!5} & .01177$^{\phantom{\ddag}}$ \cellcolor{green!25} & .01208$^{\phantom{\ddag}}$ \cellcolor{green!23} & .01153$^{\phantom{\ddag}}$ \cellcolor{green!26} \\\cline{2-13}

\textit{Rank} &  11 \cellcolor{red!28} &  8 \cellcolor{red!9} &  12 \cellcolor{red!35} &  3 \cellcolor{green!22} &  4 \cellcolor{green!15} & \textbf{1} \cellcolor{green!35} &  9 \cellcolor{red!15} &  2 \cellcolor{green!28} &  10 \cellcolor{red!22} &  6 \cellcolor{green!3} &  7 \cellcolor{red!3} &  5 \cellcolor{green!9} \\\hline

\end{tabular}
}%
 \caption{Values of MAE obtained in our experiments for different multiclass quantification methods optimized for MAE on different groups of datasets.}
 \label{tab:multi_mae_full}
\end{table}
\nopagebreak 
\begin{table}[h]
 \centering
\resizebox{\textwidth}{!}{%

            \begin{tabular}{|c|c|c|c|c|c|c|c|c|c|c|c|c|} \cline{2-13}           
            \multicolumn{1}{c}{} & 
            \multicolumn{2}{|c}{Adjustment} & 
            \multicolumn{6}{|c|}{Distribution Matching} & 
            \multicolumn{4}{c|}{Maximum Likelihood} \\
            \hline               
            \textbf{Tweets} & ACC & PACC & HDy-OvA & DM-T & DM-HD & KDEy-HD & DM-CS & KDEy-CS & DIR & EMQ & EMQ-BCTS & KDEy-ML \\\hline
gasp &  0.37225$^{\phantom{\ddag}}$ \cellcolor{green!12} &  0.29326$^{\phantom{\ddag}}$ \cellcolor{green!27} &  0.44459$^{\phantom{\ddag}}$ \cellcolor{red!0} &  0.32874$^{\phantom{\ddag}}$ \cellcolor{green!21} &  0.28929$^{\phantom{\ddag}}$ \cellcolor{green!28} &  0.26124$^{\phantom{\ddag}}$ \cellcolor{green!33} &  0.35572$^{\phantom{\ddag}}$ \cellcolor{green!16} &  0.27852$^{\ddag}$ \cellcolor{green!30} &  0.42947$^{\phantom{\ddag}}$ \cellcolor{green!2} &  0.35914$^{\phantom{\ddag}}$ \cellcolor{green!15} &  0.62647$^{\phantom{\ddag}}$ \cellcolor{red!35} & \textbf{0.25496}$^{\phantom{\ddag}}$ \cellcolor{green!35} \\\cline{2-13}
hcr &  0.86801$^{\phantom{\ddag}}$ \cellcolor{red!35} &  0.79477$^{\phantom{\ddag}}$ \cellcolor{red!22} &  0.66679$^{\phantom{\ddag}}$ \cellcolor{red!1} &  0.59522$^{\phantom{\ddag}}$ \cellcolor{green!10} &  0.63863$^{\phantom{\ddag}}$ \cellcolor{green!2} &  0.60654$^{\phantom{\ddag}}$ \cellcolor{green!8} &  0.56566$^{\phantom{\ddag}}$ \cellcolor{green!15} &  0.55952$^{\phantom{\ddag}}$ \cellcolor{green!16} &  0.45479$^{\phantom{\ddag}}$ \cellcolor{green!33} & \textbf{0.44521}$^{\phantom{\ddag}}$ \cellcolor{green!35} &  0.51711$^{\phantom{\ddag}}$ \cellcolor{green!23} &  0.46854$^{\ddag}$ \cellcolor{green!31} \\\cline{2-13}
omd &  0.56967$^{\phantom{\ddag}}$ \cellcolor{red!19} &  0.42726$^{\phantom{\ddag}}$ \cellcolor{green!12} &  0.63857$^{\phantom{\ddag}}$ \cellcolor{red!35} &  0.61895$^{\phantom{\ddag}}$ \cellcolor{red!30} &  0.37808$^{\phantom{\ddag}}$ \cellcolor{green!23} &  0.45936$^{\phantom{\ddag}}$ \cellcolor{green!4} &  0.48768$^{\phantom{\ddag}}$ \cellcolor{red!1} &  0.41857$^{\phantom{\ddag}}$ \cellcolor{green!13} &  0.51853$^{\phantom{\ddag}}$ \cellcolor{red!8} &  0.39500$^{\phantom{\ddag}}$ \cellcolor{green!19} &  0.37332$^{\dag}$ \cellcolor{green!24} & \textbf{0.32427}$^{\phantom{\ddag}}$ \cellcolor{green!35} \\\cline{2-13}
Sanders &  0.40512$^{\phantom{\ddag}}$ \cellcolor{green!9} &  0.36327$^{\phantom{\ddag}}$ \cellcolor{green!15} &  0.56331$^{\phantom{\ddag}}$ \cellcolor{red!14} &  0.30559$^{\phantom{\ddag}}$ \cellcolor{green!23} &  0.26275$^{\phantom{\ddag}}$ \cellcolor{green!30} & \textbf{0.23231}$^{\phantom{\ddag}}$ \cellcolor{green!35} &  0.26648$^{\phantom{\ddag}}$ \cellcolor{green!29} &  0.26750$^{\phantom{\ddag}}$ \cellcolor{green!29} &  0.57998$^{\phantom{\ddag}}$ \cellcolor{red!17} &  0.28101$^{\phantom{\ddag}}$ \cellcolor{green!27} &  0.69777$^{\phantom{\ddag}}$ \cellcolor{red!35} &  0.26037$^{\phantom{\ddag}}$ \cellcolor{green!30} \\\cline{2-13}
SemEval13 &  0.55411$^{\phantom{\ddag}}$ \cellcolor{green!8} &  0.50860$^{\phantom{\ddag}}$ \cellcolor{green!17} &  0.79220$^{\phantom{\ddag}}$ \cellcolor{red!35} &  0.43602$^{\phantom{\ddag}}$ \cellcolor{green!30} &  0.44950$^{\phantom{\ddag}}$ \cellcolor{green!28} &  0.44898$^{\phantom{\ddag}}$ \cellcolor{green!28} &  0.49009$^{\phantom{\ddag}}$ \cellcolor{green!20} &  0.42761$^{\ddag}$ \cellcolor{green!32} &  0.70310$^{\phantom{\ddag}}$ \cellcolor{red!18} &  0.66054$^{\phantom{\ddag}}$ \cellcolor{red!10} &  0.43243$^{\phantom{\ddag}}$ \cellcolor{green!31} & \textbf{0.41223}$^{\phantom{\ddag}}$ \cellcolor{green!35} \\\cline{2-13}
SemEval14 &  0.49481$^{\phantom{\ddag}}$ \cellcolor{green!4} &  0.45413$^{\phantom{\ddag}}$ \cellcolor{green!15} &  0.64734$^{\phantom{\ddag}}$ \cellcolor{red!35} &  0.41251$^{\phantom{\ddag}}$ \cellcolor{green!25} &  0.40196$^{\phantom{\ddag}}$ \cellcolor{green!28} &  0.40232$^{\phantom{\ddag}}$ \cellcolor{green!28} &  0.47810$^{\phantom{\ddag}}$ \cellcolor{green!8} &  0.40633$^{\phantom{\ddag}}$ \cellcolor{green!27} &  0.56211$^{\phantom{\ddag}}$ \cellcolor{red!12} &  0.57333$^{\phantom{\ddag}}$ \cellcolor{red!15} & \textbf{0.37741}$^{\phantom{\ddag}}$ \cellcolor{green!35} &  0.39554$^{\phantom{\ddag}}$ \cellcolor{green!30} \\\cline{2-13}
SemEval15 &  0.69746$^{\phantom{\ddag}}$ \cellcolor{green!3} &  0.66874$^{\phantom{\ddag}}$ \cellcolor{green!11} &  0.83673$^{\phantom{\ddag}}$ \cellcolor{red!35} &  0.60081$^{\phantom{\ddag}}$ \cellcolor{green!29} &  0.60188$^{\phantom{\ddag}}$ \cellcolor{green!29} & \textbf{0.58260}$^{\phantom{\ddag}}$ \cellcolor{green!35} &  0.63606$^{\phantom{\ddag}}$ \cellcolor{green!20} &  0.60223$^{\ddag}$ \cellcolor{green!29} &  0.79960$^{\phantom{\ddag}}$ \cellcolor{red!24} &  0.75774$^{\phantom{\ddag}}$ \cellcolor{red!13} &  0.67190$^{\dag}$ \cellcolor{green!10} &  0.58579$^{\phantom{\ddag}}$ \cellcolor{green!34} \\\cline{2-13}
SemEval16 &  1.10508$^{\phantom{\ddag}}$ \cellcolor{red!0} &  0.96384$^{\ddag}$ \cellcolor{green!11} &  1.06239$^{\phantom{\ddag}}$ \cellcolor{green!2} &  0.83537$^{\phantom{\ddag}}$ \cellcolor{green!21} &  0.81140$^{\phantom{\ddag}}$ \cellcolor{green!23} & \textbf{0.67679}$^{\phantom{\ddag}}$ \cellcolor{green!35} &  0.84354$^{\phantom{\ddag}}$ \cellcolor{green!21} &  0.97161$^{\phantom{\ddag}}$ \cellcolor{green!10} &  1.51534$^{\phantom{\ddag}}$ \cellcolor{red!35} &  0.98026$^{\ddag}$ \cellcolor{green!9} &  1.00988$^{\ddag}$ \cellcolor{green!7} &  0.76681$^{\phantom{\ddag}}$ \cellcolor{green!27} \\\cline{2-13}
sst &  0.54269$^{\phantom{\ddag}}$ \cellcolor{red!3} &  0.47170$^{\phantom{\ddag}}$ \cellcolor{green!10} &  0.41352$^{\phantom{\ddag}}$ \cellcolor{green!21} &  0.37498$^{\phantom{\ddag}}$ \cellcolor{green!28} &  0.35942$^{\phantom{\ddag}}$ \cellcolor{green!31} & \textbf{0.34271}$^{\phantom{\ddag}}$ \cellcolor{green!35} &  0.42177$^{\phantom{\ddag}}$ \cellcolor{green!19} &  0.46344$^{\phantom{\ddag}}$ \cellcolor{green!11} &  0.48335$^{\phantom{\ddag}}$ \cellcolor{green!7} &  0.44465$^{\ddag}$ \cellcolor{green!15} &  0.70442$^{\phantom{\ddag}}$ \cellcolor{red!35} &  0.35142$^{\dag}$ \cellcolor{green!33} \\\cline{2-13}
wa &  0.32136$^{\phantom{\ddag}}$ \cellcolor{red!35} &  0.29951$^{\phantom{\ddag}}$ \cellcolor{red!20} &  0.29201$^{\phantom{\ddag}}$ \cellcolor{red!15} & \textbf{0.21424}$^{\phantom{\ddag}}$ \cellcolor{green!35} &  0.22128$^{\phantom{\ddag}}$ \cellcolor{green!30} &  0.21789$^{\ddag}$ \cellcolor{green!32} &  0.30173$^{\phantom{\ddag}}$ \cellcolor{red!22} &  0.23331$^{\ddag}$ \cellcolor{green!22} &  0.27256$^{\phantom{\ddag}}$ \cellcolor{red!3} &  0.22379$^{\ddag}$ \cellcolor{green!28} &  0.26533$^{\phantom{\ddag}}$ \cellcolor{green!1} &  0.23197$^{\ddag}$ \cellcolor{green!23} \\\cline{2-13}
wb &  0.33726$^{\phantom{\ddag}}$ \cellcolor{red!35} &  0.31116$^{\phantom{\ddag}}$ \cellcolor{red!20} &  0.30866$^{\phantom{\ddag}}$ \cellcolor{red!19} &  0.24166$^{\phantom{\ddag}}$ \cellcolor{green!17} &  0.23300$^{\phantom{\ddag}}$ \cellcolor{green!22} &  0.22224$^{\phantom{\ddag}}$ \cellcolor{green!28} &  0.27188$^{\phantom{\ddag}}$ \cellcolor{green!0} &  0.25751$^{\phantom{\ddag}}$ \cellcolor{green!8} &  0.24260$^{\phantom{\ddag}}$ \cellcolor{green!17} & \textbf{0.20992}$^{\phantom{\ddag}}$ \cellcolor{green!35} &  0.21274$^{\phantom{\ddag}}$ \cellcolor{green!33} &  0.22016$^{\phantom{\ddag}}$ \cellcolor{green!29} \\\cline{2-13}
\hline 
 \textit{Average} &  0.56980$^{\phantom{\ddag}}$ \cellcolor{red!23} &  0.50511$^{\phantom{\ddag}}$ \cellcolor{red!2} &  0.60601$^{\phantom{\ddag}}$ \cellcolor{red!35} &  0.45128$^{\phantom{\ddag}}$ \cellcolor{green!14} &  0.42247$^{\phantom{\ddag}}$ \cellcolor{green!24} &  0.40482$^{\phantom{\ddag}}$ \cellcolor{green!29} &  0.46534$^{\phantom{\ddag}}$ \cellcolor{green!10} &  0.44420$^{\phantom{\ddag}}$ \cellcolor{green!17} &  0.59649$^{\phantom{\ddag}}$ \cellcolor{red!31} &  0.48460$^{\phantom{\ddag}}$ \cellcolor{green!4} &  0.53534$^{\phantom{\ddag}}$ \cellcolor{red!12} & \textbf{0.38837}$^{\phantom{\ddag}}$ \cellcolor{green!35} \\\cline{2-13}
\textit{Rank} &  10.3 \cellcolor{red!35} &  8.0 \cellcolor{red!15} &  10.3 \cellcolor{red!35} &  5.4 \cellcolor{green!7} &  4.2 \cellcolor{green!17} &  3.3 \cellcolor{green!25} &  7.0 \cellcolor{red!6} &  5.4 \cellcolor{green!7} &  9.2 \cellcolor{red!25} &  6.4 \cellcolor{red!1} &  6.5 \cellcolor{red!2} & \textbf{2.2} \cellcolor{green!35} \\\hline
\hline
\textbf{UCI-multi} & ACC & PACC & HDy-OvA & DM-T & DM-HD & KDEy-HD & DM-CS & KDEy-CS & DIR & EMQ & EMQ-BCTS & KDEy-ML \\\hline
dry-bean &  0.10697$^{\phantom{\ddag}}$ \cellcolor{green!33} &  0.09932$^{\phantom{\ddag}}$ \cellcolor{green!33} &  0.67745$^{\phantom{\ddag}}$ \cellcolor{red!0} &  0.08905$^{\phantom{\ddag}}$ \cellcolor{green!34} &  0.08681$^{\phantom{\ddag}}$ \cellcolor{green!34} &  0.07687$^{\ddag}$ \cellcolor{green!34} &  0.10495$^{\phantom{\ddag}}$ \cellcolor{green!33} &  0.08853$^{\phantom{\ddag}}$ \cellcolor{green!34} &  1.25722$^{\phantom{\ddag}}$ \cellcolor{red!35} &  0.08582$^{\phantom{\ddag}}$ \cellcolor{green!34} &  0.09642$^{\phantom{\ddag}}$ \cellcolor{green!33} & \textbf{0.07321}$^{\phantom{\ddag}}$ \cellcolor{green!35} \\\cline{2-13}
wine-quality &  3.95377$^{\phantom{\ddag}}$ \cellcolor{red!22} &  4.11163$^{\phantom{\ddag}}$ \cellcolor{red!27} &  3.27246$^{\phantom{\ddag}}$ \cellcolor{red!0} &  2.34737$^{\ddag}$ \cellcolor{green!28} &  2.15553$^{\ddag}$ \cellcolor{green!34} &  2.23004$^{\dag}$ \cellcolor{green!32} &  2.99908$^{\phantom{\ddag}}$ \cellcolor{green!8} &  2.52138$^{\phantom{\ddag}}$ \cellcolor{green!23} &  4.36191$^{\phantom{\ddag}}$ \cellcolor{red!35} &  2.56550$^{\ddag}$ \cellcolor{green!21} &  3.61678$^{\phantom{\ddag}}$ \cellcolor{red!11} & \textbf{2.14771}$^{\phantom{\ddag}}$ \cellcolor{green!35} \\\cline{2-13}
academic-success &  0.45828$^{\phantom{\ddag}}$ \cellcolor{red!2} &  0.31747$^{\phantom{\ddag}}$ \cellcolor{green!15} &  0.54860$^{\phantom{\ddag}}$ \cellcolor{red!14} & \textbf{0.16107}$^{\phantom{\ddag}}$ \cellcolor{green!35} &  0.17045$^{\ddag}$ \cellcolor{green!33} &  0.21692$^{\phantom{\ddag}}$ \cellcolor{green!27} &  0.22369$^{\phantom{\ddag}}$ \cellcolor{green!27} &  0.28252$^{\phantom{\ddag}}$ \cellcolor{green!19} &  0.71326$^{\phantom{\ddag}}$ \cellcolor{red!35} &  0.27068$^{\phantom{\ddag}}$ \cellcolor{green!21} &  0.22257$^{\phantom{\ddag}}$ \cellcolor{green!27} &  0.20673$^{\phantom{\ddag}}$ \cellcolor{green!29} \\\cline{2-13}
digits &  0.11388$^{\phantom{\ddag}}$ \cellcolor{red!8} &  0.08512$^{\phantom{\ddag}}$ \cellcolor{green!11} &  0.15331$^{\phantom{\ddag}}$ \cellcolor{red!35} &  0.05744$^{\phantom{\ddag}}$ \cellcolor{green!29} &  0.08200$^{\phantom{\ddag}}$ \cellcolor{green!13} &  0.05270$^{\phantom{\ddag}}$ \cellcolor{green!33} &  0.10291$^{\phantom{\ddag}}$ \cellcolor{red!0} &  0.05563$^{\phantom{\ddag}}$ \cellcolor{green!31} &  0.12836$^{\phantom{\ddag}}$ \cellcolor{red!18} &  0.05121$^{\ddag}$ \cellcolor{green!34} &  0.05080$^{\phantom{\ddag}}$ \cellcolor{green!34} & \textbf{0.04997}$^{\phantom{\ddag}}$ \cellcolor{green!35} \\\cline{2-13}
letter &  0.37570$^{\phantom{\ddag}}$ \cellcolor{green!12} &  0.31580$^{\phantom{\ddag}}$ \cellcolor{green!18} &  0.88616$^{\phantom{\ddag}}$ \cellcolor{red!35} &  0.25673$^{\phantom{\ddag}}$ \cellcolor{green!23} &  0.27577$^{\phantom{\ddag}}$ \cellcolor{green!22} &  0.21494$^{\phantom{\ddag}}$ \cellcolor{green!27} &  0.58296$^{\phantom{\ddag}}$ \cellcolor{red!6} &  0.24267$^{\phantom{\ddag}}$ \cellcolor{green!25} &  0.71883$^{\phantom{\ddag}}$ \cellcolor{red!19} &  0.26120$^{\phantom{\ddag}}$ \cellcolor{green!23} &  0.26144$^{\phantom{\ddag}}$ \cellcolor{green!23} & \textbf{0.13769}$^{\phantom{\ddag}}$ \cellcolor{green!35} \\\cline{2-13}
\hline 
 \textit{Average} &  1.00172$^{\phantom{\ddag}}$ \cellcolor{red!1} &  0.98587$^{\phantom{\ddag}}$ \cellcolor{red!0} &  1.10760$^{\phantom{\ddag}}$ \cellcolor{red!9} &  0.58233$^{\phantom{\ddag}}$ \cellcolor{green!30} &  0.55411$^{\phantom{\ddag}}$ \cellcolor{green!32} &  0.55829$^{\phantom{\ddag}}$ \cellcolor{green!32} &  0.80272$^{\phantom{\ddag}}$ \cellcolor{green!13} &  0.63815$^{\phantom{\ddag}}$ \cellcolor{green!26} &  1.43592$^{\phantom{\ddag}}$ \cellcolor{red!35} &  0.64688$^{\phantom{\ddag}}$ \cellcolor{green!25} &  0.84960$^{\phantom{\ddag}}$ \cellcolor{green!9} & \textbf{0.52306}$^{\phantom{\ddag}}$ \cellcolor{green!35} \\\cline{2-13}
\textit{Rank} &  9.8 \cellcolor{red!22} &  8.8 \cellcolor{red!15} &  10.8 \cellcolor{red!29} &  4.2 \cellcolor{green!15} &  4.4 \cellcolor{green!14} &  3.0 \cellcolor{green!24} &  8.2 \cellcolor{red!11} &  5.2 \cellcolor{green!8} &  11.6 \cellcolor{red!35} &  4.8 \cellcolor{green!11} &  5.8 \cellcolor{green!4} & \textbf{1.4} \cellcolor{green!35} \\\hline
\hline
\textbf{LeQua} & ACC & PACC & HDy-OvA & DM-T & DM-HD & KDEy-HD & DM-CS & KDEy-CS & DIR & EMQ & EMQ-BCTS & KDEy-ML \\\hline
T1B &  1.72371$^{\phantom{\ddag}}$ \cellcolor{red!17} &  1.37638$^{\phantom{\ddag}}$ \cellcolor{green!2} &  2.04481$^{\phantom{\ddag}}$ \cellcolor{red!35} &  0.92422$^{\phantom{\ddag}}$ \cellcolor{green!27} &  0.91116$^{\phantom{\ddag}}$ \cellcolor{green!27} & \textbf{0.78339}$^{\phantom{\ddag}}$ \cellcolor{green!35} &  1.73593$^{\phantom{\ddag}}$ \cellcolor{red!17} &  0.84367$^{\phantom{\ddag}}$ \cellcolor{green!31} &  1.97045$^{\phantom{\ddag}}$ \cellcolor{red!30} &  0.87802$^{\phantom{\ddag}}$ \cellcolor{green!29} &  0.98367$^{\phantom{\ddag}}$ \cellcolor{green!23} &  0.82795$^{\phantom{\ddag}}$ \cellcolor{green!32} \\\cline{2-13}

\textit{Rank} &  9 \cellcolor{red!15} &  8 \cellcolor{red!9} &  12 \cellcolor{red!35} &  6 \cellcolor{green!3} &  5 \cellcolor{green!9} & \textbf{1} \cellcolor{green!35} &  10 \cellcolor{red!22} &  3 \cellcolor{green!22} &  11 \cellcolor{red!28} &  4 \cellcolor{green!15} &  7 \cellcolor{red!3} &  2 \cellcolor{green!28} \\\hline

\end{tabular}
}%
 \caption{Values of MRAE obtained in our experiments for different multiclass quantification methods optimized for MRAE on different groups of datasets.}
 \label{tab:multi_mrae_full}
\end{table}

\section{Appendix}
\label{app:binresults}

Tables~\ref{tab:uci_mae} and \ref{tab:uci_mrae} report the results we have obtained for the 29 UCI binary datasets (that we omitted in Section~\ref{sec:binary}) when the methods have been optimized for, and evaluated in terms of MAE and MRAE, respectively.

\begin{table}[h]
 \centering
\resizebox{\textwidth}{!}{%

            \begin{tabular}{|c|c|c|c|c|c|c|c|c|c|c|c|c|} \cline{2-13}           
            \multicolumn{1}{c}{} & 
            \multicolumn{2}{|c}{Adjustment} & 
            \multicolumn{6}{|c|}{Distribution Matching} & 
            \multicolumn{4}{c|}{Maximum Likelihood} \\
            \hline               
            \textbf{UCI-binary} & ACC & PACC & HDy & DM-T & DM-HD & KDEy-HD & DM-CS & KDEy-CS & DIR & EMQ & EMQ-BCTS & KDEy-ML \\\hline
balance.1 & .01769$^{\phantom{\ddag}}$ \cellcolor{green!24} & \textbf{.01125}$^{\phantom{\ddag}}$ \cellcolor{green!35} & .01929$^{\phantom{\ddag}}$ \cellcolor{green!21} & .01358$^{\phantom{\ddag}}$ \cellcolor{green!31} & .01924$^{\phantom{\ddag}}$ \cellcolor{green!21} & .02146$^{\phantom{\ddag}}$ \cellcolor{green!17} & .03711$^{\phantom{\ddag}}$ \cellcolor{red!8} & .05275$^{\phantom{\ddag}}$ \cellcolor{red!35} & .01379$^{\phantom{\ddag}}$ \cellcolor{green!30} & .02025$^{\phantom{\ddag}}$ \cellcolor{green!19} & .03820$^{\phantom{\ddag}}$ \cellcolor{red!10} & .01825$^{\phantom{\ddag}}$ \cellcolor{green!23} \\\cline{2-13}
balance.3 & .02383$^{\phantom{\ddag}}$ \cellcolor{red!6} & .01855$^{\phantom{\ddag}}$ \cellcolor{green!26} & .01790$^{\ddag}$ \cellcolor{green!30} & .02847$^{\phantom{\ddag}}$ \cellcolor{red!35} & .01889$^{\phantom{\ddag}}$ \cellcolor{green!24} & .02109$^{\phantom{\ddag}}$ \cellcolor{green!10} & .02396$^{\phantom{\ddag}}$ \cellcolor{red!6} & .02422$^{\phantom{\ddag}}$ \cellcolor{red!8} & \textbf{.01719}$^{\phantom{\ddag}}$ \cellcolor{green!35} & .02420$^{\phantom{\ddag}}$ \cellcolor{red!8} & .02464$^{\phantom{\ddag}}$ \cellcolor{red!11} & .02372$^{\phantom{\ddag}}$ \cellcolor{red!5} \\\cline{2-13}
breast-cancer & .02764$^{\phantom{\ddag}}$ \cellcolor{red!6} & .02360$^{\phantom{\ddag}}$ \cellcolor{green!2} & .04109$^{\phantom{\ddag}}$ \cellcolor{red!35} & .01731$^{\phantom{\ddag}}$ \cellcolor{green!15} & .02434$^{\phantom{\ddag}}$ \cellcolor{green!0} & .02028$^{\phantom{\ddag}}$ \cellcolor{green!9} & .00976$^{\phantom{\ddag}}$ \cellcolor{green!31} & \textbf{.00800}$^{\phantom{\ddag}}$ \cellcolor{green!35} & .01275$^{\phantom{\ddag}}$ \cellcolor{green!24} & .00895$^{\phantom{\ddag}}$ \cellcolor{green!32} & .00893$^{\phantom{\ddag}}$ \cellcolor{green!33} & .01311$^{\phantom{\ddag}}$ \cellcolor{green!24} \\\cline{2-13}
cmc.1 & .08956$^{\phantom{\ddag}}$ \cellcolor{green!25} & .08625$^{\dag}$ \cellcolor{green!28} & .15430$^{\phantom{\ddag}}$ \cellcolor{red!35} & .09403$^{\phantom{\ddag}}$ \cellcolor{green!21} & .09622$^{\phantom{\ddag}}$ \cellcolor{green!19} & .09223$^{\phantom{\ddag}}$ \cellcolor{green!23} & .12185$^{\phantom{\ddag}}$ \cellcolor{red!4} & .10372$^{\phantom{\ddag}}$ \cellcolor{green!12} & \textbf{.07961}$^{\phantom{\ddag}}$ \cellcolor{green!35} & .09124$^{\phantom{\ddag}}$ \cellcolor{green!24} & .13579$^{\phantom{\ddag}}$ \cellcolor{red!17} & .09736$^{\phantom{\ddag}}$ \cellcolor{green!18} \\\cline{2-13}
cmc.2 & .09874$^{\phantom{\ddag}}$ \cellcolor{green!23} & .10111$^{\phantom{\ddag}}$ \cellcolor{green!22} & .27122$^{\phantom{\ddag}}$ \cellcolor{red!35} & .17772$^{\phantom{\ddag}}$ \cellcolor{red!3} & .18373$^{\phantom{\ddag}}$ \cellcolor{red!5} & .08614$^{\phantom{\ddag}}$ \cellcolor{green!27} & .08641$^{\phantom{\ddag}}$ \cellcolor{green!27} & .07881$^{\phantom{\ddag}}$ \cellcolor{green!29} & \textbf{.06343}$^{\phantom{\ddag}}$ \cellcolor{green!35} & .07000$^{\phantom{\ddag}}$ \cellcolor{green!32} & .15614$^{\phantom{\ddag}}$ \cellcolor{green!3} & .08788$^{\phantom{\ddag}}$ \cellcolor{green!26} \\\cline{2-13}
cmc.3 & .14608$^{\phantom{\ddag}}$ \cellcolor{red!32} & .13990$^{\phantom{\ddag}}$ \cellcolor{red!23} & .10782$^{\phantom{\ddag}}$ \cellcolor{green!19} & .13005$^{\phantom{\ddag}}$ \cellcolor{red!10} & .12160$^{\phantom{\ddag}}$ \cellcolor{green!0} & .12377$^{\phantom{\ddag}}$ \cellcolor{red!2} & .14353$^{\phantom{\ddag}}$ \cellcolor{red!28} & .14828$^{\phantom{\ddag}}$ \cellcolor{red!35} & .10183$^{\ddag}$ \cellcolor{green!27} & .12717$^{\phantom{\ddag}}$ \cellcolor{red!6} & \textbf{.09612}$^{\phantom{\ddag}}$ \cellcolor{green!35} & .12330$^{\phantom{\ddag}}$ \cellcolor{red!1} \\\cline{2-13}
ctg.1 & .03067$^{\phantom{\ddag}}$ \cellcolor{red!35} & .01921$^{\phantom{\ddag}}$ \cellcolor{green!15} & .01687$^{\phantom{\ddag}}$ \cellcolor{green!25} & .01847$^{\phantom{\ddag}}$ \cellcolor{green!18} & .01826$^{\phantom{\ddag}}$ \cellcolor{green!19} & .01751$^{\phantom{\ddag}}$ \cellcolor{green!22} & .01786$^{\phantom{\ddag}}$ \cellcolor{green!21} & .02317$^{\phantom{\ddag}}$ \cellcolor{red!2} & \textbf{.01466}$^{\phantom{\ddag}}$ \cellcolor{green!35} & .01526$^{\phantom{\ddag}}$ \cellcolor{green!32} & .01492$^{\dag}$ \cellcolor{green!33} & .01607$^{\phantom{\ddag}}$ \cellcolor{green!28} \\\cline{2-13}
ctg.2 & .02474$^{\phantom{\ddag}}$ \cellcolor{green!14} & .03317$^{\phantom{\ddag}}$ \cellcolor{red!9} & .04204$^{\phantom{\ddag}}$ \cellcolor{red!35} & .01845$^{\ddag}$ \cellcolor{green!33} & .01862$^{\ddag}$ \cellcolor{green!32} & .01798$^{\ddag}$ \cellcolor{green!34} & .02012$^{\phantom{\ddag}}$ \cellcolor{green!28} & .01920$^{\phantom{\ddag}}$ \cellcolor{green!30} & .02198$^{\phantom{\ddag}}$ \cellcolor{green!22} & \textbf{.01780}$^{\phantom{\ddag}}$ \cellcolor{green!35} & .02466$^{\phantom{\ddag}}$ \cellcolor{green!15} & .01935$^{\phantom{\ddag}}$ \cellcolor{green!30} \\\cline{2-13}
ctg.3 & .02595$^{\phantom{\ddag}}$ \cellcolor{green!21} & .05073$^{\phantom{\ddag}}$ \cellcolor{red!19} & .03328$^{\phantom{\ddag}}$ \cellcolor{green!9} & .03806$^{\phantom{\ddag}}$ \cellcolor{green!1} & .02972$^{\phantom{\ddag}}$ \cellcolor{green!15} & .02911$^{\phantom{\ddag}}$ \cellcolor{green!16} & \textbf{.01820}$^{\phantom{\ddag}}$ \cellcolor{green!35} & .01864$^{\ddag}$ \cellcolor{green!34} & .04014$^{\phantom{\ddag}}$ \cellcolor{red!1} & .05973$^{\phantom{\ddag}}$ \cellcolor{red!35} & .04490$^{\phantom{\ddag}}$ \cellcolor{red!10} & .02543$^{\phantom{\ddag}}$ \cellcolor{green!22} \\\cline{2-13}
german & .07083$^{\phantom{\ddag}}$ \cellcolor{green!20} & .05913$^{\phantom{\ddag}}$ \cellcolor{green!29} & .08455$^{\phantom{\ddag}}$ \cellcolor{green!10} & .07001$^{\phantom{\ddag}}$ \cellcolor{green!21} & .07002$^{\phantom{\ddag}}$ \cellcolor{green!21} & .06507$^{\phantom{\ddag}}$ \cellcolor{green!24} & .05341$^{\phantom{\ddag}}$ \cellcolor{green!33} & \textbf{.05129}$^{\phantom{\ddag}}$ \cellcolor{green!35} & .06726$^{\phantom{\ddag}}$ \cellcolor{green!23} & .10180$^{\phantom{\ddag}}$ \cellcolor{red!1} & .14762$^{\phantom{\ddag}}$ \cellcolor{red!35} & .08384$^{\phantom{\ddag}}$ \cellcolor{green!11} \\\cline{2-13}
haberman & .15226$^{\phantom{\ddag}}$ \cellcolor{green!20} & .15689$^{\phantom{\ddag}}$ \cellcolor{green!19} & .31347$^{\phantom{\ddag}}$ \cellcolor{red!22} & .30098$^{\phantom{\ddag}}$ \cellcolor{red!19} & .36011$^{\phantom{\ddag}}$ \cellcolor{red!35} & \textbf{.09742}$^{\phantom{\ddag}}$ \cellcolor{green!35} & .12137$^{\phantom{\ddag}}$ \cellcolor{green!28} & .18291$^{\phantom{\ddag}}$ \cellcolor{green!12} & .13972$^{\phantom{\ddag}}$ \cellcolor{green!23} & .13554$^{\phantom{\ddag}}$ \cellcolor{green!24} & .14637$^{\phantom{\ddag}}$ \cellcolor{green!21} & .12167$^{\phantom{\ddag}}$ \cellcolor{green!28} \\\cline{2-13}
ionosphere & .06711$^{\phantom{\ddag}}$ \cellcolor{green!2} & .05077$^{\phantom{\ddag}}$ \cellcolor{green!18} & .06637$^{\phantom{\ddag}}$ \cellcolor{green!2} & .03729$^{\phantom{\ddag}}$ \cellcolor{green!32} & .04210$^{\phantom{\ddag}}$ \cellcolor{green!27} & \textbf{.03441}$^{\phantom{\ddag}}$ \cellcolor{green!35} & .09228$^{\phantom{\ddag}}$ \cellcolor{red!23} & .07849$^{\phantom{\ddag}}$ \cellcolor{red!9} & .04236$^{\phantom{\ddag}}$ \cellcolor{green!27} & .03466$^{\ddag}$ \cellcolor{green!34} & .10394$^{\phantom{\ddag}}$ \cellcolor{red!35} & .06550$^{\phantom{\ddag}}$ \cellcolor{green!3} \\\cline{2-13}
iris.2 & .16216$^{\phantom{\ddag}}$ \cellcolor{red!9} & .08910$^{\phantom{\ddag}}$ \cellcolor{green!13} & .08674$^{\phantom{\ddag}}$ \cellcolor{green!14} & .02508$^{\phantom{\ddag}}$ \cellcolor{green!34} & .03726$^{\phantom{\ddag}}$ \cellcolor{green!30} & .03905$^{\phantom{\ddag}}$ \cellcolor{green!29} & .03841$^{\phantom{\ddag}}$ \cellcolor{green!30} & \textbf{.02314}$^{\phantom{\ddag}}$ \cellcolor{green!35} & .06572$^{\phantom{\ddag}}$ \cellcolor{green!21} & .09805$^{\phantom{\ddag}}$ \cellcolor{green!11} & .24293$^{\phantom{\ddag}}$ \cellcolor{red!35} & .03034$^{\phantom{\ddag}}$ \cellcolor{green!32} \\\cline{2-13}
iris.3 & .06777$^{\phantom{\ddag}}$ \cellcolor{red!33} & .06269$^{\phantom{\ddag}}$ \cellcolor{red!26} & \textbf{.01884}$^{\phantom{\ddag}}$ \cellcolor{green!35} & .06778$^{\phantom{\ddag}}$ \cellcolor{red!33} & .06777$^{\phantom{\ddag}}$ \cellcolor{red!33} & .06686$^{\phantom{\ddag}}$ \cellcolor{red!32} & .06781$^{\phantom{\ddag}}$ \cellcolor{red!33} & .06483$^{\phantom{\ddag}}$ \cellcolor{red!29} & .06757$^{\phantom{\ddag}}$ \cellcolor{red!33} & .06276$^{\phantom{\ddag}}$ \cellcolor{red!26} & .05541$^{\phantom{\ddag}}$ \cellcolor{red!16} & .06868$^{\phantom{\ddag}}$ \cellcolor{red!35} \\\cline{2-13}
mammographic & .04967$^{\phantom{\ddag}}$ \cellcolor{green!2} & .04319$^{\phantom{\ddag}}$ \cellcolor{green!16} & .06640$^{\phantom{\ddag}}$ \cellcolor{red!34} & .04238$^{\phantom{\ddag}}$ \cellcolor{green!18} & .04264$^{\phantom{\ddag}}$ \cellcolor{green!17} & .04378$^{\phantom{\ddag}}$ \cellcolor{green!15} & .03881$^{\phantom{\ddag}}$ \cellcolor{green!26} & .04650$^{\phantom{\ddag}}$ \cellcolor{green!9} & .06642$^{\phantom{\ddag}}$ \cellcolor{red!35} & .04248$^{\phantom{\ddag}}$ \cellcolor{green!17} & \textbf{.03477}$^{\phantom{\ddag}}$ \cellcolor{green!35} & .04001$^{\phantom{\ddag}}$ \cellcolor{green!23} \\\cline{2-13}
pageblocks.5 & .02860$^{\phantom{\ddag}}$ \cellcolor{green!32} & .02209$^{\phantom{\ddag}}$ \cellcolor{green!33} & .09531$^{\phantom{\ddag}}$ \cellcolor{green!17} & .02234$^{\phantom{\ddag}}$ \cellcolor{green!33} & .02061$^{\phantom{\ddag}}$ \cellcolor{green!34} & .01950$^{\phantom{\ddag}}$ \cellcolor{green!34} & .02275$^{\phantom{\ddag}}$ \cellcolor{green!33} & .02052$^{\phantom{\ddag}}$ \cellcolor{green!34} & .33955$^{\phantom{\ddag}}$ \cellcolor{red!35} & \textbf{.01667}$^{\phantom{\ddag}}$ \cellcolor{green!35} & .02379$^{\phantom{\ddag}}$ \cellcolor{green!33} & .01886$^{\phantom{\ddag}}$ \cellcolor{green!34} \\\cline{2-13}
semeion & .02235$^{\phantom{\ddag}}$ \cellcolor{green!26} & .03079$^{\phantom{\ddag}}$ \cellcolor{green!20} & .02418$^{\phantom{\ddag}}$ \cellcolor{green!25} & .01703$^{\phantom{\ddag}}$ \cellcolor{green!31} & .01987$^{\phantom{\ddag}}$ \cellcolor{green!28} & .01498$^{\phantom{\ddag}}$ \cellcolor{green!32} & .01760$^{\phantom{\ddag}}$ \cellcolor{green!30} & .01792$^{\phantom{\ddag}}$ \cellcolor{green!30} & .02811$^{\phantom{\ddag}}$ \cellcolor{green!22} & \textbf{.01215}$^{\phantom{\ddag}}$ \cellcolor{green!35} & .09933$^{\phantom{\ddag}}$ \cellcolor{red!35} & .01796$^{\phantom{\ddag}}$ \cellcolor{green!30} \\\cline{2-13}
sonar & .16360$^{\phantom{\ddag}}$ \cellcolor{red!35} & .15535$^{\phantom{\ddag}}$ \cellcolor{red!29} & .13170$^{\phantom{\ddag}}$ \cellcolor{red!15} & .06595$^{\phantom{\ddag}}$ \cellcolor{green!25} & .06867$^{\phantom{\ddag}}$ \cellcolor{green!23} & \textbf{.04994}$^{\phantom{\ddag}}$ \cellcolor{green!35} & .06179$^{\phantom{\ddag}}$ \cellcolor{green!27} & .05253$^{\dag}$ \cellcolor{green!33} & .09989$^{\phantom{\ddag}}$ \cellcolor{green!4} & .06729$^{\phantom{\ddag}}$ \cellcolor{green!24} & .07215$^{\phantom{\ddag}}$ \cellcolor{green!21} & .05860$^{\phantom{\ddag}}$ \cellcolor{green!29} \\\cline{2-13}
spambase & .02304$^{\phantom{\ddag}}$ \cellcolor{green!22} & .02058$^{\phantom{\ddag}}$ \cellcolor{green!29} & .02357$^{\phantom{\ddag}}$ \cellcolor{green!21} & .01894$^{\dag}$ \cellcolor{green!33} & .01908$^{\dag}$ \cellcolor{green!33} & .01865$^{\ddag}$ \cellcolor{green!34} & .02030$^{\phantom{\ddag}}$ \cellcolor{green!29} & .01945$^{\phantom{\ddag}}$ \cellcolor{green!32} & .04484$^{\phantom{\ddag}}$ \cellcolor{red!35} & .01945$^{\phantom{\ddag}}$ \cellcolor{green!32} & .01981$^{\phantom{\ddag}}$ \cellcolor{green!31} & \textbf{.01836}$^{\phantom{\ddag}}$ \cellcolor{green!35} \\\cline{2-13}
spectf & .19959$^{\phantom{\ddag}}$ \cellcolor{red!35} & .05076$^{\phantom{\ddag}}$ \cellcolor{green!28} & .18881$^{\phantom{\ddag}}$ \cellcolor{red!30} & .04095$^{\phantom{\ddag}}$ \cellcolor{green!32} & .04095$^{\phantom{\ddag}}$ \cellcolor{green!32} & \textbf{.03623}$^{\phantom{\ddag}}$ \cellcolor{green!35} & .04413$^{\phantom{\ddag}}$ \cellcolor{green!31} & .04334$^{\phantom{\ddag}}$ \cellcolor{green!31} & .06379$^{\phantom{\ddag}}$ \cellcolor{green!23} & .12048$^{\phantom{\ddag}}$ \cellcolor{red!1} & .10734$^{\phantom{\ddag}}$ \cellcolor{green!4} & .04391$^{\phantom{\ddag}}$ \cellcolor{green!31} \\\cline{2-13}
tictactoe & .00954$^{\phantom{\ddag}}$ \cellcolor{green!18} & .00905$^{\phantom{\ddag}}$ \cellcolor{green!22} & .00977$^{\phantom{\ddag}}$ \cellcolor{green!16} & .00833$^{\phantom{\ddag}}$ \cellcolor{green!27} & .00778$^{\phantom{\ddag}}$ \cellcolor{green!32} & .00827$^{\phantom{\ddag}}$ \cellcolor{green!28} & \textbf{.00740}$^{\phantom{\ddag}}$ \cellcolor{green!35} & .00760$^{\ddag}$ \cellcolor{green!33} & .01645$^{\phantom{\ddag}}$ \cellcolor{red!35} & .00746$^{\ddag}$ \cellcolor{green!34} & .01342$^{\phantom{\ddag}}$ \cellcolor{red!11} & .01093$^{\phantom{\ddag}}$ \cellcolor{green!7} \\\cline{2-13}
transfusion & .10365$^{\phantom{\ddag}}$ \cellcolor{green!16} & .08166$^{\phantom{\ddag}}$ \cellcolor{green!27} & .20860$^{\phantom{\ddag}}$ \cellcolor{red!35} & .07376$^{\phantom{\ddag}}$ \cellcolor{green!30} & .07381$^{\phantom{\ddag}}$ \cellcolor{green!30} & .06597$^{\ddag}$ \cellcolor{green!34} & .07609$^{\phantom{\ddag}}$ \cellcolor{green!29} & .07328$^{\phantom{\ddag}}$ \cellcolor{green!31} & .09593$^{\phantom{\ddag}}$ \cellcolor{green!20} & .06775$^{\phantom{\ddag}}$ \cellcolor{green!33} & \textbf{.06556}$^{\phantom{\ddag}}$ \cellcolor{green!35} & .07195$^{\phantom{\ddag}}$ \cellcolor{green!31} \\\cline{2-13}
wdbc & .01997$^{\phantom{\ddag}}$ \cellcolor{green!9} & .01690$^{\phantom{\ddag}}$ \cellcolor{green!17} & .03576$^{\phantom{\ddag}}$ \cellcolor{red!35} & \textbf{.01076}$^{\phantom{\ddag}}$ \cellcolor{green!35} & .01206$^{\phantom{\ddag}}$ \cellcolor{green!31} & .01364$^{\phantom{\ddag}}$ \cellcolor{green!26} & .01617$^{\phantom{\ddag}}$ \cellcolor{green!19} & .01884$^{\phantom{\ddag}}$ \cellcolor{green!12} & .02080$^{\phantom{\ddag}}$ \cellcolor{green!6} & .01298$^{\phantom{\ddag}}$ \cellcolor{green!28} & .01141$^{\dag}$ \cellcolor{green!33} & .01904$^{\phantom{\ddag}}$ \cellcolor{green!11} \\\cline{2-13}
wine.1 & .03010$^{\phantom{\ddag}}$ \cellcolor{green!19} & .03781$^{\phantom{\ddag}}$ \cellcolor{green!13} & .10453$^{\phantom{\ddag}}$ \cellcolor{red!35} & .02190$^{\phantom{\ddag}}$ \cellcolor{green!25} & \textbf{.00851}$^{\phantom{\ddag}}$ \cellcolor{green!35} & .01334$^{\phantom{\ddag}}$ \cellcolor{green!31} & .02221$^{\phantom{\ddag}}$ \cellcolor{green!25} & .02124$^{\phantom{\ddag}}$ \cellcolor{green!25} & .02214$^{\phantom{\ddag}}$ \cellcolor{green!25} & .01299$^{\phantom{\ddag}}$ \cellcolor{green!31} & .02015$^{\phantom{\ddag}}$ \cellcolor{green!26} & .00962$^{\phantom{\ddag}}$ \cellcolor{green!34} \\\cline{2-13}
wine.2 & .02731$^{\phantom{\ddag}}$ \cellcolor{red!35} & .02417$^{\phantom{\ddag}}$ \cellcolor{red!23} & .01976$^{\phantom{\ddag}}$ \cellcolor{red!7} & .01076$^{\phantom{\ddag}}$ \cellcolor{green!25} & .01092$^{\phantom{\ddag}}$ \cellcolor{green!25} & .01142$^{\phantom{\ddag}}$ \cellcolor{green!23} & .01203$^{\phantom{\ddag}}$ \cellcolor{green!21} & .01148$^{\phantom{\ddag}}$ \cellcolor{green!23} & .01281$^{\phantom{\ddag}}$ \cellcolor{green!18} & .00936$^{\phantom{\ddag}}$ \cellcolor{green!31} & .01559$^{\phantom{\ddag}}$ \cellcolor{green!8} & \textbf{.00831}$^{\phantom{\ddag}}$ \cellcolor{green!35} \\\cline{2-13}
wine.3 & .05240$^{\phantom{\ddag}}$ \cellcolor{red!2} & .02997$^{\phantom{\ddag}}$ \cellcolor{green!17} & .03851$^{\phantom{\ddag}}$ \cellcolor{green!9} & .01439$^{\phantom{\ddag}}$ \cellcolor{green!31} & .01760$^{\phantom{\ddag}}$ \cellcolor{green!28} & .01276$^{\phantom{\ddag}}$ \cellcolor{green!32} & .02422$^{\phantom{\ddag}}$ \cellcolor{green!22} & .02337$^{\phantom{\ddag}}$ \cellcolor{green!23} & .03922$^{\phantom{\ddag}}$ \cellcolor{green!9} & .01733$^{\phantom{\ddag}}$ \cellcolor{green!28} & .08824$^{\phantom{\ddag}}$ \cellcolor{red!35} & \textbf{.01040}$^{\phantom{\ddag}}$ \cellcolor{green!35} \\\cline{2-13}
wine-q-red & .05967$^{\phantom{\ddag}}$ \cellcolor{red!35} & .04968$^{\dag}$ \cellcolor{green!21} & .05258$^{\phantom{\ddag}}$ \cellcolor{green!5} & .04943$^{\phantom{\ddag}}$ \cellcolor{green!23} & .04957$^{\phantom{\ddag}}$ \cellcolor{green!22} & .05926$^{\phantom{\ddag}}$ \cellcolor{red!32} & .05595$^{\phantom{\ddag}}$ \cellcolor{red!13} & .05427$^{\phantom{\ddag}}$ \cellcolor{red!4} & .04874$^{\ddag}$ \cellcolor{green!27} & .04785$^{\ddag}$ \cellcolor{green!32} & \textbf{.04737}$^{\phantom{\ddag}}$ \cellcolor{green!35} & .04781$^{\ddag}$ \cellcolor{green!32} \\\cline{2-13}
wine-q-white & .09855$^{\phantom{\ddag}}$ \cellcolor{red!35} & .06748$^{\phantom{\ddag}}$ \cellcolor{green!28} & .07640$^{\phantom{\ddag}}$ \cellcolor{green!9} & .07032$^{\phantom{\ddag}}$ \cellcolor{green!22} & .06685$^{\phantom{\ddag}}$ \cellcolor{green!29} & .06945$^{\phantom{\ddag}}$ \cellcolor{green!24} & .07032$^{\phantom{\ddag}}$ \cellcolor{green!22} & .07399$^{\phantom{\ddag}}$ \cellcolor{green!14} & .06487$^{\dag}$ \cellcolor{green!33} & \textbf{.06409}$^{\phantom{\ddag}}$ \cellcolor{green!35} & .06476$^{\ddag}$ \cellcolor{green!33} & .06559$^{\phantom{\ddag}}$ \cellcolor{green!31} \\\cline{2-13}
yeast & .12623$^{\phantom{\ddag}}$ \cellcolor{green!3} & .10517$^{\phantom{\ddag}}$ \cellcolor{green!13} & .06788$^{\dag}$ \cellcolor{green!32} & .07149$^{\phantom{\ddag}}$ \cellcolor{green!30} & .07077$^{\phantom{\ddag}}$ \cellcolor{green!30} & \textbf{.06255}$^{\phantom{\ddag}}$ \cellcolor{green!35} & .08089$^{\phantom{\ddag}}$ \cellcolor{green!25} & .07534$^{\phantom{\ddag}}$ \cellcolor{green!28} & .08417$^{\phantom{\ddag}}$ \cellcolor{green!24} & .09952$^{\phantom{\ddag}}$ \cellcolor{green!16} & .20408$^{\phantom{\ddag}}$ \cellcolor{red!35} & .07758$^{\phantom{\ddag}}$ \cellcolor{green!27} \\\cline{2-13}
\hline 
 \textit{Average} & .06963$^{\phantom{\ddag}}$ \cellcolor{red!11} & .05679$^{\phantom{\ddag}}$ \cellcolor{green!10} & .08336$^{\phantom{\ddag}}$ \cellcolor{red!35} & .05434$^{\phantom{\ddag}}$ \cellcolor{green!14} & .05647$^{\phantom{\ddag}}$ \cellcolor{green!11} & \textbf{.04249}$^{\phantom{\ddag}}$ \cellcolor{green!35} & .04906$^{\phantom{\ddag}}$ \cellcolor{green!23} & .04956$^{\phantom{\ddag}}$ \cellcolor{green!22} & .06192$^{\phantom{\ddag}}$ \cellcolor{green!1} & .05122$^{\phantom{\ddag}}$ \cellcolor{green!20} & .07339$^{\phantom{\ddag}}$ \cellcolor{red!17} & .04529$^{\phantom{\ddag}}$ \cellcolor{green!30} \\\cline{2-13}
\textit{Rank} &  9.4 \cellcolor{red!35} &  7.6 \cellcolor{red!9} &  8.9 \cellcolor{red!28} &  5.7 \cellcolor{green!17} &  5.7 \cellcolor{green!17} & \textbf{4.4} \cellcolor{green!35} &  6.5 \cellcolor{green!5} &  6.2 \cellcolor{green!9} &  6.8 \cellcolor{green!1} &  4.8 \cellcolor{green!29} &  7.3 \cellcolor{red!5} &  4.9 \cellcolor{green!27} \\\hline

\end{tabular}
}%
\caption{UCI binary datasets, methods optimized for MAE, and evaluated in terms of MAE}
 \label{tab:uci_mae}
\end{table}

\begin{table}[h]
 \centering
\resizebox{\textwidth}{!}{%

            \begin{tabular}{|c|c|c|c|c|c|c|c|c|c|c|c|c|} \cline{2-13}           
            \multicolumn{1}{c}{} & 
            \multicolumn{2}{|c}{Adjustment} & 
            \multicolumn{6}{|c|}{Distribution Matching} & 
            \multicolumn{4}{c|}{Maximum Likelihood} \\
            \hline               
            \textbf{UCI-binary} & ACC & PACC & HDy & DM-T & DM-HD & KDEy-HD & DM-CS & KDEy-CS & DIR & EMQ & EMQ-BCTS & KDEy-ML \\\hline
balance.1 &  0.07112$^{\dag}$ \cellcolor{green!21} &  0.05394$^{\phantom{\ddag}}$ \cellcolor{green!32} &  0.06789$^{\phantom{\ddag}}$ \cellcolor{green!23} & \textbf{0.05070}$^{\phantom{\ddag}}$ \cellcolor{green!35} &  0.07789$^{\phantom{\ddag}}$ \cellcolor{green!17} &  0.07458$^{\phantom{\ddag}}$ \cellcolor{green!19} &  0.15757$^{\phantom{\ddag}}$ \cellcolor{red!35} &  0.14958$^{\phantom{\ddag}}$ \cellcolor{red!29} &  0.06028$^{\dag}$ \cellcolor{green!28} &  0.06760$^{\phantom{\ddag}}$ \cellcolor{green!23} &  0.15011$^{\phantom{\ddag}}$ \cellcolor{red!30} &  0.09535$^{\phantom{\ddag}}$ \cellcolor{green!5} \\\cline{2-13}
balance.3 &  0.11974$^{\phantom{\ddag}}$ \cellcolor{red!13} &  0.11497$^{\phantom{\ddag}}$ \cellcolor{red!9} &  0.12617$^{\phantom{\ddag}}$ \cellcolor{red!19} &  0.09061$^{\phantom{\ddag}}$ \cellcolor{green!11} &  0.11132$^{\phantom{\ddag}}$ \cellcolor{red!6} &  0.10595$^{\phantom{\ddag}}$ \cellcolor{red!1} &  0.14388$^{\phantom{\ddag}}$ \cellcolor{red!35} &  0.12839$^{\phantom{\ddag}}$ \cellcolor{red!21} & \textbf{0.06365}$^{\phantom{\ddag}}$ \cellcolor{green!35} &  0.10724$^{\phantom{\ddag}}$ \cellcolor{red!3} &  0.11009$^{\phantom{\ddag}}$ \cellcolor{red!5} &  0.11134$^{\phantom{\ddag}}$ \cellcolor{red!6} \\\cline{2-13}
breast-cancer &  0.12421$^{\phantom{\ddag}}$ \cellcolor{red!27} &  0.11482$^{\phantom{\ddag}}$ \cellcolor{red!21} &  0.13413$^{\phantom{\ddag}}$ \cellcolor{red!35} &  0.07536$^{\phantom{\ddag}}$ \cellcolor{green!7} &  0.09187$^{\phantom{\ddag}}$ \cellcolor{red!4} &  0.07237$^{\phantom{\ddag}}$ \cellcolor{green!9} &  0.04959$^{\phantom{\ddag}}$ \cellcolor{green!25} &  0.04249$^{\phantom{\ddag}}$ \cellcolor{green!30} &  0.06936$^{\phantom{\ddag}}$ \cellcolor{green!11} & \textbf{0.03660}$^{\phantom{\ddag}}$ \cellcolor{green!35} &  0.04530$^{\phantom{\ddag}}$ \cellcolor{green!28} &  0.03825$^{\phantom{\ddag}}$ \cellcolor{green!33} \\\cline{2-13}
cmc.1 &  0.64740$^{\phantom{\ddag}}$ \cellcolor{red!35} &  0.40790$^{\phantom{\ddag}}$ \cellcolor{green!17} &  0.54136$^{\dag}$ \cellcolor{red!11} &  0.33894$^{\phantom{\ddag}}$ \cellcolor{green!32} &  0.34912$^{\phantom{\ddag}}$ \cellcolor{green!30} & \textbf{0.32662}$^{\phantom{\ddag}}$ \cellcolor{green!35} &  0.47610$^{\phantom{\ddag}}$ \cellcolor{green!2} &  0.37953$^{\phantom{\ddag}}$ \cellcolor{green!23} &  0.47213$^{\dag}$ \cellcolor{green!3} &  0.34765$^{\phantom{\ddag}}$ \cellcolor{green!30} &  0.42833$^{\phantom{\ddag}}$ \cellcolor{green!12} &  0.33036$^{\phantom{\ddag}}$ \cellcolor{green!34} \\\cline{2-13}
cmc.2 &  0.49309$^{\phantom{\ddag}}$ \cellcolor{green!22} &  0.44461$^{\phantom{\ddag}}$ \cellcolor{green!25} &  1.45497$^{\phantom{\ddag}}$ \cellcolor{red!35} &  0.79129$^{\phantom{\ddag}}$ \cellcolor{green!4} &  0.78553$^{\phantom{\ddag}}$ \cellcolor{green!4} &  0.35677$^{\phantom{\ddag}}$ \cellcolor{green!30} &  0.69000$^{\phantom{\ddag}}$ \cellcolor{green!10} &  0.58605$^{\phantom{\ddag}}$ \cellcolor{green!16} & \textbf{0.28042}$^{\phantom{\ddag}}$ \cellcolor{green!35} &  0.30444$^{\phantom{\ddag}}$ \cellcolor{green!33} &  0.64009$^{\phantom{\ddag}}$ \cellcolor{green!13} &  0.31391$^{\phantom{\ddag}}$ \cellcolor{green!33} \\\cline{2-13}
cmc.3 &  0.54046$^{\phantom{\ddag}}$ \cellcolor{red!3} &  0.67741$^{\phantom{\ddag}}$ \cellcolor{red!35} &  0.62307$^{\phantom{\ddag}}$ \cellcolor{red!22} &  0.43845$^{\phantom{\ddag}}$ \cellcolor{green!19} &  0.42377$^{\phantom{\ddag}}$ \cellcolor{green!23} &  0.41053$^{\phantom{\ddag}}$ \cellcolor{green!26} &  0.52581$^{\phantom{\ddag}}$ \cellcolor{red!0} &  0.48838$^{\phantom{\ddag}}$ \cellcolor{green!8} &  0.61394$^{\phantom{\ddag}}$ \cellcolor{red!20} &  0.41600$^{\phantom{\ddag}}$ \cellcolor{green!25} & \textbf{0.37328}$^{\phantom{\ddag}}$ \cellcolor{green!35} &  0.39327$^{\phantom{\ddag}}$ \cellcolor{green!30} \\\cline{2-13}
ctg.1 &  0.07898$^{\phantom{\ddag}}$ \cellcolor{red!8} &  0.09070$^{\phantom{\ddag}}$ \cellcolor{red!35} &  0.07908$^{\phantom{\ddag}}$ \cellcolor{red!8} &  0.06919$^{\phantom{\ddag}}$ \cellcolor{green!13} &  0.06860$^{\phantom{\ddag}}$ \cellcolor{green!14} &  0.06594$^{\phantom{\ddag}}$ \cellcolor{green!20} &  0.08054$^{\phantom{\ddag}}$ \cellcolor{red!12} &  0.07822$^{\phantom{\ddag}}$ \cellcolor{red!6} &  0.08243$^{\phantom{\ddag}}$ \cellcolor{red!16} &  0.06000$^{\phantom{\ddag}}$ \cellcolor{green!34} & \textbf{0.05974}$^{\phantom{\ddag}}$ \cellcolor{green!35} &  0.06267$^{\phantom{\ddag}}$ \cellcolor{green!28} \\\cline{2-13}
ctg.2 &  0.12080$^{\phantom{\ddag}}$ \cellcolor{green!8} &  0.17255$^{\phantom{\ddag}}$ \cellcolor{red!31} &  0.17651$^{\phantom{\ddag}}$ \cellcolor{red!35} &  0.09506$^{\ddag}$ \cellcolor{green!28} &  0.09565$^{\ddag}$ \cellcolor{green!28} &  0.08988$^{\ddag}$ \cellcolor{green!32} &  0.09545$^{\phantom{\ddag}}$ \cellcolor{green!28} &  0.09703$^{\dag}$ \cellcolor{green!27} &  0.12283$^{\phantom{\ddag}}$ \cellcolor{green!6} & \textbf{0.08684}$^{\phantom{\ddag}}$ \cellcolor{green!35} &  0.09951$^{\phantom{\ddag}}$ \cellcolor{green!25} &  0.09802$^{\ddag}$ \cellcolor{green!26} \\\cline{2-13}
ctg.3 &  0.13629$^{\phantom{\ddag}}$ \cellcolor{green!5} &  0.23311$^{\phantom{\ddag}}$ \cellcolor{red!35} &  0.11940$^{\phantom{\ddag}}$ \cellcolor{green!12} &  0.12351$^{\phantom{\ddag}}$ \cellcolor{green!11} &  0.18102$^{\phantom{\ddag}}$ \cellcolor{red!13} &  0.12668$^{\phantom{\ddag}}$ \cellcolor{green!9} &  0.07125$^{\phantom{\ddag}}$ \cellcolor{green!33} & \textbf{0.06693}$^{\phantom{\ddag}}$ \cellcolor{green!35} &  0.17313$^{\phantom{\ddag}}$ \cellcolor{red!9} &  0.07292$^{\ddag}$ \cellcolor{green!32} &  0.19585$^{\phantom{\ddag}}$ \cellcolor{red!19} &  0.09961$^{\phantom{\ddag}}$ \cellcolor{green!21} \\\cline{2-13}
german &  0.40364$^{\phantom{\ddag}}$ \cellcolor{green!11} &  0.30472$^{\phantom{\ddag}}$ \cellcolor{green!23} &  0.79423$^{\phantom{\ddag}}$ \cellcolor{red!35} &  0.22694$^{\phantom{\ddag}}$ \cellcolor{green!32} &  0.46656$^{\phantom{\ddag}}$ \cellcolor{green!4} &  0.23917$^{\phantom{\ddag}}$ \cellcolor{green!31} &  0.21290$^{\phantom{\ddag}}$ \cellcolor{green!34} & \textbf{0.20972}$^{\phantom{\ddag}}$ \cellcolor{green!35} &  0.23909$^{\phantom{\ddag}}$ \cellcolor{green!31} &  0.33776$^{\phantom{\ddag}}$ \cellcolor{green!19} &  0.59948$^{\phantom{\ddag}}$ \cellcolor{red!11} &  0.30216$^{\phantom{\ddag}}$ \cellcolor{green!23} \\\cline{2-13}
haberman &  1.09284$^{\phantom{\ddag}}$ \cellcolor{green!1} &  0.50400$^{\phantom{\ddag}}$ \cellcolor{green!29} &  1.49048$^{\phantom{\ddag}}$ \cellcolor{red!17} &  1.59175$^{\phantom{\ddag}}$ \cellcolor{red!21} &  1.87374$^{\phantom{\ddag}}$ \cellcolor{red!35} & \textbf{0.38303}$^{\phantom{\ddag}}$ \cellcolor{green!35} &  1.17885$^{\phantom{\ddag}}$ \cellcolor{red!2} &  1.18703$^{\phantom{\ddag}}$ \cellcolor{red!2} &  0.49954$^{\phantom{\ddag}}$ \cellcolor{green!29} &  0.40947$^{\phantom{\ddag}}$ \cellcolor{green!33} &  0.43695$^{\phantom{\ddag}}$ \cellcolor{green!32} &  0.41646$^{\phantom{\ddag}}$ \cellcolor{green!33} \\\cline{2-13}
ionosphere &  0.34158$^{\phantom{\ddag}}$ \cellcolor{red!4} &  0.24006$^{\phantom{\ddag}}$ \cellcolor{green!15} &  0.27348$^{\phantom{\ddag}}$ \cellcolor{green!8} & \textbf{0.13774}$^{\phantom{\ddag}}$ \cellcolor{green!35} &  0.14993$^{\phantom{\ddag}}$ \cellcolor{green!32} &  0.14287$^{\phantom{\ddag}}$ \cellcolor{green!34} &  0.49831$^{\phantom{\ddag}}$ \cellcolor{red!35} &  0.41450$^{\phantom{\ddag}}$ \cellcolor{red!18} &  0.34257$^{\phantom{\ddag}}$ \cellcolor{red!4} &  0.18468$^{\ddag}$ \cellcolor{green!25} &  0.31626$^{\phantom{\ddag}}$ \cellcolor{green!0} &  0.36913$^{\phantom{\ddag}}$ \cellcolor{red!9} \\\cline{2-13}
iris.2 &  0.48852$^{\phantom{\ddag}}$ \cellcolor{green!8} &  0.32526$^{\phantom{\ddag}}$ \cellcolor{green!20} &  0.29141$^{\phantom{\ddag}}$ \cellcolor{green!22} &  0.17372$^{\phantom{\ddag}}$ \cellcolor{green!30} &  0.13826$^{\phantom{\ddag}}$ \cellcolor{green!33} &  0.12597$^{\phantom{\ddag}}$ \cellcolor{green!33} &  0.18014$^{\phantom{\ddag}}$ \cellcolor{green!30} &  0.12498$^{\phantom{\ddag}}$ \cellcolor{green!33} &  0.62834$^{\phantom{\ddag}}$ \cellcolor{red!0} &  0.47648$^{\phantom{\ddag}}$ \cellcolor{green!9} &  1.12056$^{\phantom{\ddag}}$ \cellcolor{red!35} & \textbf{0.11035}$^{\phantom{\ddag}}$ \cellcolor{green!35} \\\cline{2-13}
iris.3 &  0.37933$^{\phantom{\ddag}}$ \cellcolor{red!31} &  0.33284$^{\phantom{\ddag}}$ \cellcolor{red!19} & \textbf{0.11242}$^{\phantom{\ddag}}$ \cellcolor{green!35} &  0.37974$^{\phantom{\ddag}}$ \cellcolor{red!31} &  0.37933$^{\phantom{\ddag}}$ \cellcolor{red!31} &  0.36952$^{\phantom{\ddag}}$ \cellcolor{red!28} &  0.37954$^{\phantom{\ddag}}$ \cellcolor{red!31} &  0.36225$^{\phantom{\ddag}}$ \cellcolor{red!27} &  0.38025$^{\phantom{\ddag}}$ \cellcolor{red!31} &  0.33049$^{\phantom{\ddag}}$ \cellcolor{red!19} &  0.26591$^{\phantom{\ddag}}$ \cellcolor{red!3} &  0.39412$^{\phantom{\ddag}}$ \cellcolor{red!35} \\\cline{2-13}
mammographic &  0.32067$^{\phantom{\ddag}}$ \cellcolor{green!2} &  0.29023$^{\phantom{\ddag}}$ \cellcolor{green!8} &  0.47184$^{\phantom{\ddag}}$ \cellcolor{red!28} &  0.23843$^{\phantom{\ddag}}$ \cellcolor{green!19} &  0.23732$^{\phantom{\ddag}}$ \cellcolor{green!19} &  0.25063$^{\phantom{\ddag}}$ \cellcolor{green!16} &  0.25182$^{\phantom{\ddag}}$ \cellcolor{green!16} &  0.26240$^{\phantom{\ddag}}$ \cellcolor{green!14} &  0.50550$^{\phantom{\ddag}}$ \cellcolor{red!35} &  0.17420$^{\phantom{\ddag}}$ \cellcolor{green!32} & \textbf{0.15954}$^{\phantom{\ddag}}$ \cellcolor{green!35} &  0.17949$^{\phantom{\ddag}}$ \cellcolor{green!30} \\\cline{2-13}
pageblocks.5 &  0.13987$^{\phantom{\ddag}}$ \cellcolor{green!32} &  0.11171$^{\ddag}$ \cellcolor{green!34} &  0.31369$^{\phantom{\ddag}}$ \cellcolor{green!26} &  0.09612$^{\ddag}$ \cellcolor{green!34} &  0.08858$^{\phantom{\ddag}}$ \cellcolor{green!34} &  0.08695$^{\phantom{\ddag}}$ \cellcolor{green!34} &  0.08941$^{\ddag}$ \cellcolor{green!34} &  0.08708$^{\ddag}$ \cellcolor{green!34} &  1.88826$^{\phantom{\ddag}}$ \cellcolor{red!35} &  0.09617$^{\phantom{\ddag}}$ \cellcolor{green!34} & \textbf{0.08619}$^{\phantom{\ddag}}$ \cellcolor{green!35} &  0.09144$^{\ddag}$ \cellcolor{green!34} \\\cline{2-13}
semeion &  0.27897$^{\phantom{\ddag}}$ \cellcolor{red!27} &  0.17288$^{\phantom{\ddag}}$ \cellcolor{green!1} &  0.08747$^{\phantom{\ddag}}$ \cellcolor{green!23} &  0.16026$^{\phantom{\ddag}}$ \cellcolor{green!4} &  0.18683$^{\phantom{\ddag}}$ \cellcolor{red!2} &  0.06313$^{\phantom{\ddag}}$ \cellcolor{green!30} &  0.07645$^{\phantom{\ddag}}$ \cellcolor{green!26} &  0.06755$^{\phantom{\ddag}}$ \cellcolor{green!29} &  0.07445$^{\phantom{\ddag}}$ \cellcolor{green!27} & \textbf{0.04577}$^{\phantom{\ddag}}$ \cellcolor{green!35} &  0.30756$^{\phantom{\ddag}}$ \cellcolor{red!35} &  0.06940$^{\phantom{\ddag}}$ \cellcolor{green!28} \\\cline{2-13}
sonar &  0.81838$^{\phantom{\ddag}}$ \cellcolor{red!35} &  0.59209$^{\phantom{\ddag}}$ \cellcolor{red!10} &  0.74573$^{\phantom{\ddag}}$ \cellcolor{red!27} &  0.22124$^{\phantom{\ddag}}$ \cellcolor{green!29} &  0.22782$^{\phantom{\ddag}}$ \cellcolor{green!29} & \textbf{0.17289}$^{\phantom{\ddag}}$ \cellcolor{green!35} &  0.22830$^{\phantom{\ddag}}$ \cellcolor{green!28} &  0.21346$^{\phantom{\ddag}}$ \cellcolor{green!30} &  0.52661$^{\phantom{\ddag}}$ \cellcolor{red!3} &  0.34445$^{\phantom{\ddag}}$ \cellcolor{green!16} &  0.27959$^{\phantom{\ddag}}$ \cellcolor{green!23} &  0.19494$^{\phantom{\ddag}}$ \cellcolor{green!32} \\\cline{2-13}
spambase &  0.11150$^{\phantom{\ddag}}$ \cellcolor{green!26} &  0.09588$^{\phantom{\ddag}}$ \cellcolor{green!31} &  0.09794$^{\phantom{\ddag}}$ \cellcolor{green!30} & \textbf{0.08206}$^{\phantom{\ddag}}$ \cellcolor{green!35} &  0.08218$^{\ddag}$ \cellcolor{green!34} &  0.08247$^{\ddag}$ \cellcolor{green!34} &  0.08953$^{\phantom{\ddag}}$ \cellcolor{green!32} &  0.08565$^{\phantom{\ddag}}$ \cellcolor{green!33} &  0.33164$^{\phantom{\ddag}}$ \cellcolor{red!35} &  0.09334$^{\dag}$ \cellcolor{green!31} &  0.09055$^{\dag}$ \cellcolor{green!32} &  0.08302$^{\ddag}$ \cellcolor{green!34} \\\cline{2-13}
spectf &  0.57311$^{\phantom{\ddag}}$ \cellcolor{green!1} &  0.33522$^{\phantom{\ddag}}$ \cellcolor{green!20} &  1.04759$^{\phantom{\ddag}}$ \cellcolor{red!35} &  0.18082$^{\phantom{\ddag}}$ \cellcolor{green!32} &  0.19621$^{\phantom{\ddag}}$ \cellcolor{green!31} & \textbf{0.14596}$^{\phantom{\ddag}}$ \cellcolor{green!35} &  0.18122$^{\phantom{\ddag}}$ \cellcolor{green!32} &  0.17453$^{\phantom{\ddag}}$ \cellcolor{green!32} &  0.32534$^{\phantom{\ddag}}$ \cellcolor{green!21} &  0.15283$^{\phantom{\ddag}}$ \cellcolor{green!34} &  0.47857$^{\phantom{\ddag}}$ \cellcolor{green!9} &  0.16363$^{\phantom{\ddag}}$ \cellcolor{green!33} \\\cline{2-13}
tictactoe &  0.04126$^{\ddag}$ \cellcolor{green!25} &  0.03911$^{\dag}$ \cellcolor{green!27} & \textbf{0.03092}$^{\phantom{\ddag}}$ \cellcolor{green!35} &  0.03596$^{\phantom{\ddag}}$ \cellcolor{green!30} &  0.04190$^{\ddag}$ \cellcolor{green!24} &  0.03262$^{\phantom{\ddag}}$ \cellcolor{green!33} &  0.03262$^{\phantom{\ddag}}$ \cellcolor{green!33} &  0.03138$^{\phantom{\ddag}}$ \cellcolor{green!34} &  0.10565$^{\phantom{\ddag}}$ \cellcolor{red!35} &  0.03092$^{\phantom{\ddag}}$ \cellcolor{green!34} &  0.06338$^{\phantom{\ddag}}$ \cellcolor{green!4} &  0.03597$^{\phantom{\ddag}}$ \cellcolor{green!30} \\\cline{2-13}
transfusion &  0.54969$^{\phantom{\ddag}}$ \cellcolor{green!13} &  0.40587$^{\phantom{\ddag}}$ \cellcolor{green!24} &  1.19529$^{\phantom{\ddag}}$ \cellcolor{red!35} &  0.28794$^{\ddag}$ \cellcolor{green!32} &  0.28768$^{\ddag}$ \cellcolor{green!33} &  0.29478$^{\phantom{\ddag}}$ \cellcolor{green!32} &  0.32646$^{\phantom{\ddag}}$ \cellcolor{green!30} &  0.38365$^{\ddag}$ \cellcolor{green!25} &  0.41422$^{\phantom{\ddag}}$ \cellcolor{green!23} & \textbf{0.26102}$^{\phantom{\ddag}}$ \cellcolor{green!35} &  0.27035$^{\phantom{\ddag}}$ \cellcolor{green!34} &  0.33877$^{\phantom{\ddag}}$ \cellcolor{green!29} \\\cline{2-13}
wdbc &  0.09007$^{\phantom{\ddag}}$ \cellcolor{green!6} &  0.11053$^{\phantom{\ddag}}$ \cellcolor{red!6} &  0.07228$^{\phantom{\ddag}}$ \cellcolor{green!17} &  0.05068$^{\ddag}$ \cellcolor{green!31} &  0.05029$^{\ddag}$ \cellcolor{green!31} &  0.08002$^{\phantom{\ddag}}$ \cellcolor{green!12} &  0.06259$^{\phantom{\ddag}}$ \cellcolor{green!23} &  0.06356$^{\phantom{\ddag}}$ \cellcolor{green!23} &  0.15600$^{\phantom{\ddag}}$ \cellcolor{red!35} &  0.06334$^{\phantom{\ddag}}$ \cellcolor{green!23} & \textbf{0.04454}$^{\phantom{\ddag}}$ \cellcolor{green!35} &  0.08086$^{\phantom{\ddag}}$ \cellcolor{green!12} \\\cline{2-13}
wine.1 &  0.18729$^{\phantom{\ddag}}$ \cellcolor{red!2} &  0.16758$^{\phantom{\ddag}}$ \cellcolor{green!2} &  0.30635$^{\phantom{\ddag}}$ \cellcolor{red!35} &  0.10045$^{\phantom{\ddag}}$ \cellcolor{green!20} &  0.11713$^{\phantom{\ddag}}$ \cellcolor{green!16} &  0.06193$^{\phantom{\ddag}}$ \cellcolor{green!31} &  0.09921$^{\phantom{\ddag}}$ \cellcolor{green!21} &  0.10123$^{\phantom{\ddag}}$ \cellcolor{green!20} &  0.09346$^{\phantom{\ddag}}$ \cellcolor{green!22} &  0.05039$^{\phantom{\ddag}}$ \cellcolor{green!34} &  0.07956$^{\phantom{\ddag}}$ \cellcolor{green!26} & \textbf{0.04826}$^{\phantom{\ddag}}$ \cellcolor{green!35} \\\cline{2-13}
wine.2 &  0.11551$^{\phantom{\ddag}}$ \cellcolor{red!35} &  0.10548$^{\phantom{\ddag}}$ \cellcolor{red!26} &  0.07450$^{\phantom{\ddag}}$ \cellcolor{red!0} &  0.04633$^{\phantom{\ddag}}$ \cellcolor{green!23} &  0.05494$^{\phantom{\ddag}}$ \cellcolor{green!15} &  0.04024$^{\phantom{\ddag}}$ \cellcolor{green!28} &  0.05554$^{\phantom{\ddag}}$ \cellcolor{green!15} &  0.03692$^{\phantom{\ddag}}$ \cellcolor{green!30} &  0.06236$^{\phantom{\ddag}}$ \cellcolor{green!9} &  0.03263$^{\dag}$ \cellcolor{green!34} &  0.07850$^{\phantom{\ddag}}$ \cellcolor{red!3} & \textbf{0.03204}$^{\phantom{\ddag}}$ \cellcolor{green!35} \\\cline{2-13}
wine.3 &  0.21232$^{\phantom{\ddag}}$ \cellcolor{red!16} &  0.12182$^{\phantom{\ddag}}$ \cellcolor{green!12} &  0.12997$^{\phantom{\ddag}}$ \cellcolor{green!9} &  0.06949$^{\phantom{\ddag}}$ \cellcolor{green!29} &  0.06729$^{\phantom{\ddag}}$ \cellcolor{green!29} &  0.12294$^{\phantom{\ddag}}$ \cellcolor{green!12} &  0.07350$^{\phantom{\ddag}}$ \cellcolor{green!28} &  0.08203$^{\phantom{\ddag}}$ \cellcolor{green!25} &  0.14965$^{\phantom{\ddag}}$ \cellcolor{green!3} &  0.05549$^{\phantom{\ddag}}$ \cellcolor{green!33} &  0.27041$^{\phantom{\ddag}}$ \cellcolor{red!35} & \textbf{0.05166}$^{\phantom{\ddag}}$ \cellcolor{green!35} \\\cline{2-13}
wine-q-red &  0.26981$^{\phantom{\ddag}}$ \cellcolor{red!35} &  0.22687$^{\phantom{\ddag}}$ \cellcolor{green!9} &  0.25928$^{\ddag}$ \cellcolor{red!24} &  0.21423$^{\phantom{\ddag}}$ \cellcolor{green!22} &  0.21412$^{\phantom{\ddag}}$ \cellcolor{green!22} &  0.21717$^{\phantom{\ddag}}$ \cellcolor{green!19} &  0.20453$^{\ddag}$ \cellcolor{green!32} &  0.20959$^{\ddag}$ \cellcolor{green!27} &  0.26873$^{\dag}$ \cellcolor{red!33} &  0.22927$^{\ddag}$ \cellcolor{green!7} &  0.21602$^{\phantom{\ddag}}$ \cellcolor{green!20} & \textbf{0.20249}$^{\phantom{\ddag}}$ \cellcolor{green!35} \\\cline{2-13}
wine-q-white &  0.51072$^{\phantom{\ddag}}$ \cellcolor{red!35} &  0.33435$^{\phantom{\ddag}}$ \cellcolor{green!14} &  0.28526$^{\phantom{\ddag}}$ \cellcolor{green!28} &  0.31307$^{\phantom{\ddag}}$ \cellcolor{green!20} &  0.26376$^{\phantom{\ddag}}$ \cellcolor{green!34} &  0.29319$^{\phantom{\ddag}}$ \cellcolor{green!25} &  0.38848$^{\phantom{\ddag}}$ \cellcolor{red!0} &  0.38418$^{\phantom{\ddag}}$ \cellcolor{green!0} &  0.26674$^{\ddag}$ \cellcolor{green!33} & \textbf{0.26069}$^{\phantom{\ddag}}$ \cellcolor{green!35} &  0.26502$^{\ddag}$ \cellcolor{green!33} &  0.27093$^{\phantom{\ddag}}$ \cellcolor{green!32} \\\cline{2-13}
yeast &  0.61773$^{\phantom{\ddag}}$ \cellcolor{red!4} &  0.47094$^{\phantom{\ddag}}$ \cellcolor{green!11} &  0.39139$^{\phantom{\ddag}}$ \cellcolor{green!20} &  0.27772$^{\phantom{\ddag}}$ \cellcolor{green!33} &  0.28420$^{\phantom{\ddag}}$ \cellcolor{green!32} & \textbf{0.26532}$^{\phantom{\ddag}}$ \cellcolor{green!35} &  0.42430$^{\phantom{\ddag}}$ \cellcolor{green!16} &  0.41177$^{\phantom{\ddag}}$ \cellcolor{green!18} &  0.40084$^{\phantom{\ddag}}$ \cellcolor{green!19} &  0.44344$^{\phantom{\ddag}}$ \cellcolor{green!14} &  0.88268$^{\phantom{\ddag}}$ \cellcolor{red!35} &  0.39885$^{\dag}$ \cellcolor{green!19} \\\cline{2-13}
\hline 
 \textit{Average} &  0.34051$^{\phantom{\ddag}}$ \cellcolor{red!14} &  0.26198$^{\phantom{\ddag}}$ \cellcolor{green!8} &  0.40669$^{\phantom{\ddag}}$ \cellcolor{red!35} &  0.23992$^{\phantom{\ddag}}$ \cellcolor{green!15} &  0.26182$^{\phantom{\ddag}}$ \cellcolor{green!8} & \textbf{0.17587}$^{\phantom{\ddag}}$ \cellcolor{green!35} &  0.25255$^{\phantom{\ddag}}$ \cellcolor{green!11} &  0.23828$^{\phantom{\ddag}}$ \cellcolor{green!16} &  0.33095$^{\phantom{\ddag}}$ \cellcolor{red!12} &  0.19214$^{\ddag}$ \cellcolor{green!30} &  0.29014$^{\phantom{\ddag}}$ \cellcolor{green!0} &  0.18541$^{\dag}$ \cellcolor{green!32} \\\cline{2-13}
\textit{Rank} &  9.8 \cellcolor{red!35} &  8.5 \cellcolor{red!20} &  8.8 \cellcolor{red!23} &  5.2 \cellcolor{green!19} &  6.0 \cellcolor{green!9} &  4.0 \cellcolor{green!33} &  6.9 \cellcolor{red!0} &  5.7 \cellcolor{green!13} &  8.1 \cellcolor{red!15} & \textbf{3.9} \cellcolor{green!35} &  6.6 \cellcolor{green!2} &  4.5 \cellcolor{green!28} \\\hline

\end{tabular}
}%
 \caption{UCI binary datasets, methods optimized for MRAE, and evaluated in terms of MRAE}
 \label{tab:uci_mrae}
\end{table}

\ifcauchyerror
    \clearpage
    \newpage

    \section*{Appendix C}
    \label{app:formula}

In this section, we turn to comment on the difference between \eqref{eq:Dcs} and the corresponding equation of \citet[Equation 3]{kampa2011closed}. We believe there is an error in the original formula (here replicated for the reader's convenience):
\begin{align*}
\begin{split}
    \mathcal{D}_{\mathrm{CS}}(p||q) 
        =& -\log\left( \sum_{m=1}^N \sum_{k=1}^M \pi_m \tau_k z_{mk} \right) \\
        & +\frac{1}{2}\log\left( \sum_{m=1}^N \frac{\pi_m^2 |\Lambda_m|^{1/2}}{(2\pi)^{D/2}} + 2\sum_{m=1}^N \sum_{m'<m} \pi_m \pi_{m'}z_{mm'} \right) \\
        & +\frac{1}{2}\log\left( \sum_{k=1}^M \frac{\tau_k^2 |\Omega_k|^{1/2}}{(2\pi)^{D/2}} + 2\sum_{k=1}^M \sum_{k'<k} \tau_k \tau_{k'}z_{kk'} \right) 
\end{split}
\end{align*}
The discrepancy shows up in the second and third terms. Since the derivation is analogous for both cases, we will discuss only on the former. Note that this term is actually a derivation of:
\begin{align*}
\begin{split}    
        \frac{1}{2}\log\left( \sum_{m=1}^N \frac{\pi_m^2 |\Lambda_m|^{1/2}}{(2\pi)^{D/2}} + 2\sum_{m=1}^N \sum_{m'<m} \pi_m \pi_{m'}z_{mm'} \right) = \frac{1}{2}\log\left( \sum_{m=1}^N \sum_{m'=1}^N \pi_m \pi_{m'}z_{mm'} \right) \\         
\end{split}
\end{align*}
\noindent in which the cases $m=m'$ have been factored out, and in which only the terms corresponding to $m>m'$ are computed (the remaining cases for $m<m'$ are symmetric, and this explains why the second factor is multiplied by 2). Let focus our attention to the cases for $m=m'$, that follow from the correspondence:
\begin{align}
\begin{split}
\label{eq:diag1}
        \sum_{m=1}^N \frac{\pi_m^2 |\Lambda_m|^{1/2}}{(2\pi)^{\frac{D}{2}}} 
        	&= \sum_{m=1}^N  \pi_m^2 z_{mm} \\        	
\end{split}
\end{align}
which in turn follows from the fact that
\begin{align}
\begin{split}
\label{eq:diag2}
        \frac{|\Lambda_m|^{1/2}}{(2\pi)^{\frac{D}{2}}} 
        	&=  z_{mm}         	
\end{split}
\end{align}
However, by taking a closer look at \eqref{eq:zmm} and at the well-known definition of the Normal distribution given by:
\begin{align}
\label{eq:normal}
    \mathcal{N}(\bm{x}|\mu_i, \Lambda_i^{-1}) = \frac{|\Lambda_i|^{ \frac{1}{2} }}{(2\pi)^{\frac{D}{2}}} \exp\left( -\frac{1}{2}(\bm{x}-\mu_i)^\top \Lambda_i (\bm{x}-\mu_i)\right)
\end{align}
\noindent we have the following derivation:
\begin{align*}
\label{eq:diag3}
	z_{mm} 	&= \mathcal{N}(\mu_m | \mu_{m}, (\Lambda_m^{-1}+\Lambda_{m}^{-1})) \\
        	&= \frac{|(2\Lambda_m^{-1})^{-1}|^{\frac{1}{2}}}{(2\pi)^{\frac{D}{2}}}\exp{\left( -\frac{1}{2}(\mu_m-\mu_m)^\top (2\Lambda_m^{-1})^{-1} (\mu_m-\mu_i)\right)} \\
			&= \frac{|\frac{1}{2}(\Lambda_m^{-1})^{-1}|^{\frac{1}{2}}}{(2\pi)^{\frac{D}{2}}}\exp{(0)} \\
			&= \frac{|\frac{1}{2}\Lambda_m|^{\frac{1}{2}}}{(2\pi)^{\frac{D}{2}}} \cdot 1\\
			\label{eq:different}
			&= \frac{\left(\frac{1}{2}^{D}|\Lambda_m|\right)^{\frac{1}{2}}}{(2\pi)^{\frac{D}{2}}} \\
        	&= \frac{\frac{1}{2}^{\frac{D}{2}}|\Lambda_m|^{\frac{1}{2}}}{(2\pi)^{\frac{D}{2}}} \\
                &= \frac{|\Lambda_m|^{\frac{1}{2}}}{(4\pi)^{\frac{D}{2}}} 
\end{align*}
\noindent which differs from the term $\frac{|\Lambda_m|^{\frac{1}{2}}}{(2\pi)^{\frac{D}{2}}}$ shown in \citet[Equation 3]{kampa2011closed}. Analogous derivations apply for the case $k=k'$.

\fi

\clearpage
\newpage

\bibliographystyle{plainnat}
\bibliography{references}

\end{document}